\documentclass[11pt]{article}
\usepackage[margin=1in]{geometry}

\usepackage{xcolor}

\usepackage{graphicx}
\usepackage{tikz}
\usetikzlibrary{positioning}
\usepackage{pgfplots}
\pgfplotsset{compat=1.18}

\definecolor{hoeffding_col}{RGB}{31,119,180}
\definecolor{bernstein_col}{RGB}{255,127,14}
\definecolor{empbern_col}{RGB}{44,160,44}

\usepackage{amsmath, amssymb, amsthm}

\usepackage{booktabs}
\usepackage{float}
\usepackage{subcaption}
\usepackage{placeins}          

\usepackage{enumitem}

\usepackage{algorithm}
\usepackage{algpseudocode}

\usepackage[round]{natbib}

\usepackage{authblk}
\usepackage[utf8]{inputenc}

\usepackage{hyperref}
\hypersetup{colorlinks=true, linkcolor=black, citecolor=blue, urlcolor=black}

\usepackage{float}
\usepackage{setspace}
\numberwithin{equation}{section}

\theoremstyle{plain}
\newtheorem{theorem}{Theorem}[section]
\newtheorem{lemma}[theorem]{Lemma}
\newtheorem{prop}[theorem]{Proposition}
\newtheorem{corollary}[theorem]{Corollary}

\newtheorem{example}[theorem]{Example}

\theoremstyle{remark}
\newtheorem{remark}[theorem]{Remark}

\usepackage{placeins}   
\usepackage{booktabs}   

\title{Group-Conditional Conformal Risk Control:\\
       Exact Guarantees, Limits, and Overlapping Extensions}

\author{Tareq Aldirawi}
\author{Wenge Guo\thanks{Author e-mail addresses: ta429@njit.edu; wenge.guo@njit.edu}}

\affil{Department of Mathematical Sciences \\ New Jersey Institute of Technology, Newark, NJ, USA}

\begin{document}
\maketitle
\begin{abstract}
We study group-conditional conformal risk control (CRC) for
disjoint and overlapping groups under bounded monotone losses.
For disjoint groups, we establish exact finite-sample conditional
risk control, two-sided bounds on the group-restricted risk, and fundamental
limitations arising from local support and calibration-rank
granularity.

For overlapping groups, we study two exactly valid approaches:
group-restricted calibration followed by union deployment, and
direct calibration of the deployed prediction-set family.
We characterize the union's limiting reconciliation gap,
identify conditions for its disappearance, and establish sharp
lower bounds for structured losses in specified product-form
models. We also demonstrate a strict size advantage for direct
calibration in an ordinal model.

Experiments show that direct calibration can reduce prediction-set
size, but its efficiency depends on group structure, loss type,
and shape fitting. Results on historical diabetic-retinopathy
scores further illustrate these tradeoffs. Overall, local support,
reconciliation slack, and threshold heterogeneity determine
efficiency, with neither approach uniformly dominating.
\end{abstract}
\section{Introduction}
\label{sec:introduction}

A prediction set can meet a marginal risk target while performing poorly on rare or difficult groups, particularly under structured, outcome-dependent losses. Group-conditional CRC addresses this disparity but raises three questions: when is local calibration informative, how conservative is union deployment across overlapping groups, and when can direct calibration of the deployed loss improve efficiency? We distinguish exact validity from the tightness of the achieved conditional risk.

\paragraph{Related work.}
Conformal risk control extends miscoverage guarantees to bounded monotone losses~\citep{angelopoulos2024conformal}; RCPS and learn-then-test instead provide high-probability bounds or more general testing-based risk control~\citep{bates2021distribution,angelopoulos2021learn}. Non-monotone and multidimensional variants require additional tools~\citep{angelopoulos2026nonmonotonic,aldirawi2026nonmonotonic}. Here, guarantees concern groupwise risk averaged over calibration and test draws, not high-probability risk conditional on the observed calibration set.

Mondrian and class-conditional conformal methods calibrate within partitions, including label-dependent cells~\citep{vovk2012conditional,lofstrom2015bias,sadinle2019least}. Small cells motivate class clustering~\citep{ding2023class}; our support barrier instead concerns exact validity over a fixed exchangeable loss-curve class, not arbitrary-point conditional coverage impossibility~\citep{lei2014distribution,barber2021limits} or procedures that pool data.

Overlapping coverage is studied through multivalid calibration~\citep{jung2023batch}, function classes~\citep{gibbs2025conditional}, Kandinsky's label-dependent groups~\citep{bairaktari2025kandinsky}, and approximate or sample-conditional methods~\citep{duchi2026sample}. Related risk methods include adaptive CRC~\citep{blot2024automatically}, multidimensional calibration~\citep{angelopoulos2026nonmonotonic}, and overlapping Mondrian regression intervals~\citep{bostrom2026overlapping}. Our contribution is not overlapping conditioning itself, but exact groupwise risk control for structured losses and the efficiency limits of group-conditional and direct deployed-loss calibration.

\paragraph{Contributions.}
For disjoint feature- or label-dependent groups, we establish exact finite-sample CRC and a two-sided bound on the group-restricted risk. A local-support barrier, a stronger full-triviality result for symmetric interval rules, and rank-granularity minimax limits for symmetric rules quantify the limits of informative deterministic calibration.

For overlapping groups, separately calibrated groups retain exact control
 after union deployment. We characterize finite-sample shortfall and the
 limiting reconciliation gap, give an inclusion-type alignment result, and
 prove sharp product-form lower bounds for structured losses even at small
 equal targets. Deployment-aware CRC instead calibrates the full deployed
 loss. We prove its exact validity and a strict size separation in a specified
 ordinal model. Synthetic and clinical studies show both gains from reclaimed
 reconciliation slack and losses from a restrictive global threshold.

 Section~\ref{sec:mondrian-crc} gives exact CRC and a precision bound for
 disjoint groups; Section~\ref{sec:lower-bounds}, the support barrier and
 rank-grid limits; Section~\ref{sec:split-multigroup-estimator}, overlapping
 group unions and their reconciliation gap; Section~\ref{sec:deployment-aware},
 deployment-aware calibration. Section~\ref{sec:numerical-illustrations} gives
 experiments and Section~\ref{sec:conclusion} concludes.
\section{Mondrian conformal risk control}
\label{sec:mondrian-crc}

We establish exact group-conditional CRC for disjoint groups using
 group-restricted losses, including for label-dependent partitions.

Given calibration data and groupwise targets \(\alpha_k\in(0,B_k)\), we construct a
prediction set \(\widehat{\mathcal C}(X_{n+1})\subseteq\mathcal Y\) satisfying
\begin{equation}
    \mathbb E\!\left[
        \ell\!\left(\widehat{\mathcal C}(X_{n+1}),Y_{n+1}\right)
        \,\middle|\, g(X_{n+1},Y_{n+1})=k
    \right]
    \le \alpha_k
    \label{eq:mondrian-crc-target}
\end{equation}
for every positive-probability group. Group-restricted losses decouple
 calibration even when candidate-label-dependent thresholds interact in the
 deployed set.

\subsection{Setup and coordinatewise construction}
\label{subsec:mondrian-setup}

Fix an independently fitted score $s:\mathcal X\times\mathcal Y\to\mathbb R$,
with larger values indicating less plausible labels. The calibration pairs
$Z_1,\ldots,Z_n$ and test pair $Z_{n+1}$ are exchangeable, where
$Z_i=(X_i,Y_i)$. For a measurable group identifier
$g:\mathcal X\times\mathcal Y\to[K]$, write
$G_i=g(Z_i)$, $\mathcal I_k=\{i\le n:G_i=k\}$,
$\mathcal I_k^+=\{i\le n+1:G_i=k\}$, and
$n_k=|\mathcal I_k|$.

Let $\Lambda\subseteq\overline{\mathbb R}$ be closed and nonempty, with
endpoints $\lambda_{\min}$ and $\lambda_{\max}$. At $\lambda_{\max}$ every
candidate in the group is admitted; right-continuity ensures feasible
thresholds are attained. For $\lambda\in\Lambda^K$,
\begin{equation}
    \mathcal C_\lambda(x):=\{y\in\mathcal Y:s(x,y)\le\lambda_{g(x,y)}\}.
    \label{eq:mondrian-set}
\end{equation}
Assume $\ell:2^{\mathcal Y}\times\mathcal Y\to[0,B]$ is monotone:
$C\subseteq D$ implies $\ell(D,y)\le\ell(C,y)$.
The monotonicity is essential to the construction
\citep{aldirawi2026nonmonotonic,angelopoulos2026nonmonotonic}.

For group $k$ define the group-restricted set
$\mathcal C_{k,\mu}^{0}(x)=\{y:g(x,y)=k,\ s(x,y)\le\mu\}$
and its loss
\begin{equation}
        L_i^{(k)}(\mu):=\ell\!\left(\mathcal C_{k,\mu}^{0}(X_i),Y_i\right),\qquad i\in[n+1],
    \label{eq:one-coordinate-loss}
\end{equation}
These measurable loss paths are non-increasing and, for finite labels,
step functions. The deployed loss is
$L_i^{\mathrm{dep}}(\lambda)=\ell(\mathcal C_\lambda(X_i),Y_i)$.
Since $\mathcal C_{k,\lambda_k}^{0}(x)\subseteq\mathcal C_\lambda(x)$,
on $\{G_{n+1}=k\}$ we have
\begin{equation}
    L_{n+1}^{\mathrm{dep}}(\lambda)\le L_{n+1}^{(k)}(\lambda_k),
    \label{eq:slice-domination}
\end{equation}
Thus group-restricted validity transfers to the deployed risk, without
requiring the loss to be additive across groups.

Choose $0<\alpha_k<B_k\le B$, where the deterministic constant $B_k$ bounds
every group-restricted loss in group $k$, including at test time. The empirical
maximum loss over the calibration sample does not suffice as $B_k$; a genuine
almost-sure bound, such as $B$ or the conditional essential supremum of
$\ell(\varnothing,Y)$, is required.
For $n_k\ge1$, let
$\widehat R_{n,k}^{0}(\mu)=n_k^{-1}\sum_{i\in\mathcal I_k}L_i^{(k)}(\mu)$
and set
\begin{equation}
    \widehat\lambda_k
    :=
    \inf\left\{
        \mu\in\Lambda:
        \frac{n_k}{n_k+1}\widehat R_{n,k}^{0}(\mu)+\frac{B_k}{n_k+1}\le\alpha_k
    \right\},
    \label{eq:mondrian-threshold}
\end{equation}
with \(\widehat\lambda_k=\lambda_{\max}\) if \(n_k=0\) or the feasible set is empty;
the deployed set is \(\mathcal C_{\widehat\lambda}\) as in \eqref{eq:mondrian-set}. The
\(K\) searches are decoupled, each depending on the calibration sample only through
the group-\(k\) observations. Results apply groupwise with $(B,\alpha,n)$ replaced by
 $(B_k,\alpha_k,n_k)$. Equivalently, the corrected target is
 $\alpha_k-(B_k-\alpha_k)/n_k$; if $n_k+1<B_k/\alpha_k$,
 the maximal threshold is forced. Section~\ref{sec:lower-bounds}
 analyzes this support barrier; Appendix~\ref{app:computation}
 describes computation.

\begin{algorithm}[t]
\caption{Coordinatewise Mondrian CRC}
\label{alg:coordinatewise-mondrian-crc}
\begin{algorithmic}[1]
\Require calibration sample \((X_i,Y_i)_{i=1}^{n}\); fixed score \(s\); identifier \(g\);
threshold set \(\Lambda\) with maximum \(\lambda_{\max}\); monotone loss \(\ell\);
per-group bounds \(B_k\in(0,B]\) and targets \(\alpha_k\in(0,B_k)\).
\Statex \emph{Calibration.}
\For{\(k=1,\ldots,K\)}
  \State \(\mathcal I_k\gets\{i\le n:g(X_i,Y_i)=k\}\),\quad \(n_k\gets|\mathcal I_k|\).
  \If{\(n_k=0\)}
    \State \(\widehat\lambda_k\gets\lambda_{\max}\).
  \Else
    \State With \(\widehat R_{n,k}^{0}(\mu)=n_k^{-1}\sum_{i\in\mathcal I_k}L_i^{(k)}(\mu)\)
    non-increasing in \(\mu\), take \(\widehat\lambda_k\) as the smallest \(\mu\in\Lambda\)
    with \(\tfrac{n_k}{n_k+1}\widehat R_{n,k}^{0}(\mu)+\tfrac{B_k}{n_k+1}\le\alpha_k\),
    as in \eqref{eq:mondrian-threshold}; set \(\widehat\lambda_k\gets\lambda_{\max}\) if no
    such \(\mu\) exists.
  \EndIf
\EndFor
\Statex \emph{Deployment.}
\State \Return the rule
\(\mathcal C_{\widehat\lambda}(x)=\{y\in\mathcal Y:s(x,y)\le\widehat\lambda_{g(x,y)}\}\),
each candidate label \(y\) gated by the threshold \(\widehat\lambda_{g(x,y)}\) of its own
group.
\end{algorithmic}
\end{algorithm}

\subsection{Finite-sample validity}
\label{subsec:mondrian-validity}

\begin{lemma}
\label{lem:within-group-exchangeability}
Conditional on $(G_1,\ldots,G_{n+1})$, exchangeability of the
observations implies exchangeability within every nonempty group $k$.
\end{lemma}

\begin{theorem}
\label{thm:mondrian-crc}
Let \(\widehat\lambda\) be defined by \eqref{eq:mondrian-threshold}. Assume
\((Z_i)_{i=1}^{n+1}\) is exchangeable, \(\ell\) is monotone, the maps
\(\mu\mapsto L_i^{(k)}(\mu)\) are right-continuous, \(B_k\) dominates the loss as
above, and, conditional on \(\{G_{n+1}=k\}\),
\begin{equation}
    L_i^{(k)}(\lambda_{\max})\le\alpha_k
    \qquad\text{for every } i\in\mathcal I_k\cup\{n+1\}
    \label{eq:endpoint-condition}
\end{equation}
almost surely. Then, for every \(k\) with \(\mathbb P(G_{n+1}=k)>0\),
\[
    \mathbb E\!\left[
        \ell\!\left(\mathcal C_{\widehat\lambda}(X_{n+1}),Y_{n+1}\right)
        \,\middle|\, G_{n+1}=k
    \right]\le\alpha_k.
\]
\end{theorem}

Appendix~\ref{app:proof-validity} contains the proof. The endpoint
 condition must hold for the restricted group loss, not merely the deployed set.

\subsection{Finite-sample tightness}
\label{subsec:mondrian-sandwich}

The validity bound admits a two-sided refinement for the one-coordinate risk. Write
\(\mathcal E_k:=\{\!\{L_i^{(k)}:i\in\mathcal I_k^+\}\!\}\) for the unordered multiset
of augmented one-coordinate loss functions within group \(k\).

\begin{theorem}
\label{thm:mondrian-crc-sandwich}
Under the assumptions of Theorem~\ref{thm:mondrian-crc}, fix \(k\) with
\(\mathbb P(G_{n+1}=k)>0\). Suppose, conditional on \(\{G_{n+1}=k\}\), that
\begin{equation}
    \frac{1}{n_k+1}\sum_{i\in\mathcal I_k^+}L_i^{(k)}(\lambda_{\min})
    \ge\alpha_k-\frac{B_k}{n_k+1}
    \label{eq:lower-endpoint-condition}
\end{equation}
almost surely. Let \(M_k\ge1\) be a \(\sigma(\mathcal E_k)\)-measurable
upper bound, almost surely, on the number of augmented loss paths that have a
strict downward jump at any common threshold \(\mu>\lambda_{\min}\):
\[
    \sup_{\mu\in\Lambda,\ \mu>\lambda_{\min}}
    \left|\left\{i\in\mathcal I_k^+:
    L_i^{(k)}(\mu^-)>L_i^{(k)}(\mu)\right\}\right|\le M_k,
\]
where \(L_i^{(k)}(\mu^-):=\inf_{\nu\in\Lambda,\,\nu<\mu}L_i^{(k)}(\nu)\).
The minimum threshold is handled by \eqref{eq:lower-endpoint-condition}.
Define the largest total simultaneous decrease by
\begin{equation}
    \mathfrak J_k:=\sup_{\mu\in\Lambda,\ \mu>\lambda_{\min}}
       \sum_{i\in\mathcal I_k^+}\bigl(L_i^{(k)}(\mu^-)-L_i^{(k)}(\mu)\bigr).
    \label{eq:total-jump-mass}
\end{equation}
The supremum is taken to be zero if the index set is empty. Then
\[
    \alpha_k-\mathbb E\!\left[\frac{B_k+\mathfrak J_k}{n_k+1}\,\middle|\,G_{n+1}=k\right]
    \le
    \mathbb E\!\left[L_{n+1}^{(k)}(\widehat\lambda_k)\,\middle|\,G_{n+1}=k\right]
    \le\alpha_k.
\]
Since \(\mathfrak J_k\le M_kB_k\), the original bound with
\((M_k+1)B_k\) in the numerator follows.
\end{theorem}

The proof is in Appendix~\ref{app:proof-sandwich}. For $g\equiv1$ and
 $M_k=1$, this recovers the marginal CRC sandwich
 \citep{angelopoulos2024conformal}. Endpoint and jump conditions appear in
 Appendix~\ref{app:sandwich-conditions}; exact rank tightness in
 Appendices~\ref{app:exact-two-valued}--\ref{app:rank-granularity}.
 Figure~\ref{fig:sim-sandwich} illustrates the bound.

\begin{remark}
\label{rem:slice-vs-deployed}
The sandwich bounds concern the one-coordinate risk. By
\eqref{eq:slice-domination}, the upper bound also holds for the deployed risk under any bounded monotone loss. The lower bound transfers to deployed risk when the restricted and deployed losses agree on the test group (as for inclusion-type losses), but need not transfer for general structured losses.
\end{remark}          
\section{The local-support barrier}
\label{sec:lower-bounds}

For $n_k\ge1$, the corrected group-$k$ CRC criterion has effective target
$\alpha_k-(B_k-\alpha_k)/n_k$. If $n_k+1<B_k/\alpha_k$, or if
$n_k=0$, that criterion has no feasible threshold and returns the maximal
threshold $\lambda_{\max}$. We distinguish this algorithmic fallback from
limits that apply to every distribution-free local rule, and from the
separate granularity imposed by discrete calibration ranks.

For this section, let $\Lambda$ be a closed subset
of the extended real line $\overline{\mathbb R}$, with distinct endpoints
$\lambda_{\min}<\lambda_{\max}$; $\lambda_{\max}=+\infty$ is permitted
and denotes the maximal-threshold default. Finite real endpoints are assumed
only when expressly stated, in particular in
Proposition~\ref{prop:symmetric-full-support-barrier} and the finite-endpoint
prediction-set witnesses. Thus the abstract loss-curve results below also
accommodate the $+\infty$ default of Section~\ref{sec:deployment-aware}.

Fix a local calibration count $m\ge0$ and write $N=m+1$. For
$1\le M\le N$, let $\mathcal P(m,M,B)$ comprise the exchangeable laws of
$N$ measurable, right-continuous, non-increasing loss functions
$L_i:\Lambda\to[0,B]$ satisfying $L_i(\lambda_{\max})=0$ and the
restriction that at most $M$ functions have a strict downward jump at any
common threshold $\mu>\lambda_{\min}$. A deterministic local rule
$A:(L_1,\ldots,L_m)\mapsto\Lambda$ is valid at level
$\alpha\in(0,B)$ over $\mathcal P$ if
\[
 \mathbb E_P\bigl[L_N\{A(L_1,\ldots,L_m)\}\bigr]\le\alpha
 \qquad\text{for every }P\in\mathcal P.
\]
On a finite ordered grid, a strict downward jump at $\mu>\lambda_{\min}$
means a loss decrease from the immediately preceding grid point to $\mu$.
At left-accumulation points (including $+\infty$ when it is
approached by finite thresholds), use the usual left limit along $\Lambda$. This
convention also applies in Appendix~\ref{app:rank-granularity}.
Symmetry means invariance to permutations of the $m$ calibration curves;
it is not assumed unless explicitly stated.

\paragraph{A support barrier for arbitrary deterministic rules.}
When $N<B/\alpha$, even a single unseen height-$B$ loss can force the
default action on uninformative calibration input. This conclusion needs
neither permutation invariance nor a particular CRC algorithm.

\begin{prop}
\label{thm:small-group-nontriviality}
Suppose $N<B/\alpha$. Every deterministic local rule valid over
$\mathcal P(m,1,B)$, whether symmetric or not, satisfies:
\begin{enumerate}[label=(\roman*),leftmargin=*,itemsep=2pt]
 \item $A(0,\ldots,0)=\lambda_{\max}$, with $A()$ used when $m=0$.
 \item For every $\varepsilon\in(0,1]$, there is an exchangeable law
 $P_\varepsilon\in\mathcal P(m,1,B)$ under which exactly one augmented
 curve is nonzero with probability $\varepsilon$ and all curves are zero
 otherwise, such that
 \begin{equation}
 \mathbb P_{P_\varepsilon}\{A=\lambda_{\max}\}
 \ge 1-\frac{m}{N}\varepsilon.
 \label{eq:barrier-mixture}
 \end{equation}
 In particular, taking $\varepsilon=N\alpha/B<1$ gives a witness
 $P^\ddagger$ for which
 \begin{equation}
 \mathbb P_{P^\ddagger}\{A=\lambda_{\max}\}
 \ge 1-\frac{m\alpha}{B},
 \label{eq:barrier-original-probability}
 \end{equation}
 whereas $\varepsilon=1$ gives a witness with one nonzero augmented
 curve almost surely.
\end{enumerate}
\end{prop}

\begin{proof}
If $A(0,\ldots,0)=t<\lambda_{\max}$, choose a threshold
$\mu\in\Lambda$ strictly above $t$ and let
$H_\mu(\lambda)=B\mathbf 1\{\lambda<\mu\}$. Uniformly permute an
augmented collection containing $H_\mu$ and $m$ zero curves. This is an
exchangeable member of $\mathcal P(m,1,B)$. When $H_\mu$ occupies the
test position, an event of probability $1/N$, the rule sees $m$ zero
curves and incurs loss $B$. Its risk is therefore at least $B/N>\alpha$,
a contradiction. This proves (i).

For (ii), use the all-zero collection with probability $1-\varepsilon$
and the uniformly permuted one-jump collection with probability
$\varepsilon$. The calibration curves are all zero with probability
$1-\varepsilon m/N$, and (i) forces the maximal threshold on that event.
\end{proof}

\begin{remark}\label{rem:barrier-scope}
The bound is worst-case over exchangeable local loss-curve laws. It is not
an assertion that every data distribution incurs this fallback probability,
nor does it rule out methods using additional observations, pooling, or
side information outside the stated local model. In particular, the lower
bound $1-m\alpha/B$ is tied to the displayed mixture witness.
\end{remark}

\paragraph{Full triviality for symmetric rules on intervals.}
The preceding result concerns the zero calibration input and a specific
witness law. For permutation-invariant rules on a real interval, a stronger
statement holds on every admissible input.

\begin{prop}
\label{prop:symmetric-full-support-barrier}
Suppose $\Lambda=[\lambda_{\min},\lambda_{\max}]$ is a
nondegenerate interval with finite real endpoints, $N<B/\alpha$,
and $1\le M\le N$. Every deterministic permutation-invariant local
rule valid over $\mathcal P(m,M,B)$ returns $\lambda_{\max}$ on
\emph{every} admissible $m$-tuple of loss functions whose simultaneous
strict-jump multiplicity is at most $M$.
\end{prop}

\begin{proof}
Suppose the rule returns $t<\lambda_{\max}$ on an admissible tuple.
Each monotone curve has at most countably many strict jumps, so there is
$\mu\in(t,\lambda_{\max})$ at which none of the $m$ input curves jumps.
Append $H_\mu(\lambda)=B\mathbf1\{\lambda<\mu\}$ and uniformly
permute the resulting $N$ curves. The augmented collection respects the
multiplicity bound $M$, and its random permutation is exchangeable.
With probability $1/N$, $H_\mu$ is the test curve. By symmetry, the
remaining curves then induce the original output $t$, so the test loss is
$B$. The expected risk is at least $B/N>\alpha$, a contradiction.
\end{proof}

The interval hypothesis guarantees an unused jump location above every
nonmaximal action. That step need not work on a finite threshold grid, where
all available jump sites may already be occupied. The conclusion also does
not extend as stated to randomized or order-dependent procedures.

\begin{remark}\label{rem:barrier-scale}
At $B=1$, the scale $B/\alpha$ coincides with the familiar split-CRC
feasibility scale~\citep{angelopoulos2024conformal}. The full-triviality
claim has the specific scope of the preceding proposition: deterministic,
symmetric, local rules; interval thresholds; and validity uniformly over
the specified exchangeable class.
\end{remark}

\paragraph{A distinct rank-granularity limit.}
Above the support scale, a deterministic rule may act nontrivially but still
be unable to match $\alpha$ exactly. To make this precise, consider the
\emph{common-height one-jump class}: each augmented curve has the form
$L_i(\lambda)=B\mathbf1\{\lambda<\tau_i\}$, with $N$ almost surely
distinct admissible jump thresholds. Assume $\Lambda$ contains at least
$N$ such locations. Write
\[
 \Delta_m(\alpha,B)
 :=\alpha-\frac{B}{m+1}
 \left\lfloor\frac{(m+1)\alpha}{B}\right\rfloor
 \in\left[0,\frac{B}{m+1}\right).
\]

The rank-grid mechanism is an orbit argument, not a requirement that a
general rule output an observed threshold. Fix any unordered augmented
collection of $N$ distinct one-jump curves and uniformly permute it.
For a deterministic symmetric rule, each possible test position contributes
either zero or $B$; consequently the \emph{risk on this exchangeable orbit}
is an integer multiple of $B/N$. Validity for every such orbit forces that
risk to be at most
$B\lfloor N\alpha/B\rfloor/N=\alpha-\Delta_m(\alpha,B)$.
Mixtures of orbit laws need not themselves have quantized risks, but inherit
this upper bound.

Conversely, put $r=\lfloor N\alpha/B\rfloor$. For $r=0$, return
$\lambda_{\max}$; otherwise return the $(N-r)$th smallest calibration
jump threshold. Exchangeability and distinctness give risk $Br/N$ for
every law in the one-jump class. Therefore, \emph{within that class},
\[
 \inf_{A}\ \sup_{P}
 \bigl\{\alpha-\mathbb E_P[L_N\{A(L_1,\ldots,L_m)\}]\bigr\}
 =\Delta_m(\alpha,B),
\]
where the infimum is over deterministic symmetric rules valid for all laws
in the one-jump class and the supremum is over those laws. In particular,
the usual rank-based Mondrian rule has shortfall exactly
$\Delta_m(\alpha,B)$ on this witness class. This is a restricted minimax
claim, not an assertion of exact rank quantization for arbitrary bounded
monotone losses.

A related lower bound uses tied blocks for any $1\le M\le N$.
Take $\lfloor N/M\rfloor$ distinct jump locations, each occupied
by $M$ identical height-$B$ curves, with the remaining $N\bmod M$
curves identically zero. On an exchangeable orbit, symmetry makes each
active block contribute either $0$ or $BM/N$ to expected risk. Thus a
deterministic symmetric rule valid over $\mathcal P(m,M,B)$ has a witness
with shortfall at least
\[
 \Delta_{m,M}(\alpha,B)
 :=\alpha-\frac{BM}{m+1}
 \left\lfloor\frac{(m+1)\alpha}{BM}\right\rfloor.
\]
This includes $M\nmid N$ and can vanish at rank-grid coincidences. The
witness complements, but need not attain, the general constants in the
sandwich bound of Section~\ref{sec:mondrian-crc}. Formal statements
and proofs appear in Appendices~\ref{app:exact-two-valued} and
\ref{app:rank-granularity}.

\begin{remark}\label{rem:level-prescription}
The statements fix the local count $m$ and a prespecified target $\alpha$.
For group $k$, substitute $(B,\alpha,m)=(B_k,\alpha_k,n_k)$.
Selecting targets after observing local counts requires a separate
conditional-on-count validity argument; it is not covered by the preceding
unconditional formulations.
\end{remark}

With $L$ binary labels, conditioning on every label vector can create
$2^L$ sparse cells. Overlapping groups, such as
$G_j=\{(x,y):y_j=1\}$, retain larger local calibration samples but raise
a different question: how to reconcile multiple group-restricted prediction
sets without losing conditional risk control. Section~\ref{sec:split-multigroup-estimator}
addresses this question.      
\section{Overlapping-group CRC via group-restricted unions}
\label{sec:split-multigroup-estimator}

Disjoint refinement of $m$ overlapping groups can create $2^m$ sparse
membership cells subject to Section~\ref{sec:lower-bounds}'s support barrier.
Instead, we calibrate a set within each original group and deploy their union.
This yields exact one-sided groupwise risk control for bounded monotone losses,
but may be conservative because the union contains candidates from other groups.
Section~\ref{sec:deployment-aware} studies direct calibration of the deployed set.
Overlapping joint-quantile calibration has a precedent in
Kandinsky~\citep{bairaktari2025kandinsky}, while \citet{barber2021limits}
characterize the limits of distribution-free conditional coverage; our focus is
the validity--tightness distinction for general bounded monotone losses.

\subsection{Pointwise correction and exact validity}
\label{subsec:overlap-validity}

Fix groups $G_1,\ldots,G_m\subseteq\mathcal X\times\mathcal Y$ before
calibration, with $p_j=\mathbb P(Z\in G_j)>0$, a score $s$ independent of
correction/test observations, and a measurable monotone loss $0\le\ell(C,y)\le B$. Fit a measurable shape $\widehat b$ on an independent
$\mathcal D_{\mathrm{fit}}$; for example,
$b_\beta(x,y)=\beta_0+\sum_j\beta_j\mathbf1\{(x,y)\in G_j\}$.
For $t\in\Lambda$, define
\[
 \mathcal C^{\widehat b}_{j,t}(x)
 :=\{y:(x,y)\in G_j,\ s(x,y)-\widehat b(x,y)\le t\}.
\]
Here $\Lambda\subseteq\overline{\mathbb R}$ is closed with largest element
$\lambda_{\max}$; $+\infty$ may be used to admit all eligible candidates.
Let $n_j=\#\{i\in\mathcal D_{\mathrm{cal}}:Z_i\in G_j\}$ and
$0<\alpha_j<B_j\le B$. Each $B_j$ is fixed before correction (possibly
using independent fitting data) and bounds the restricted loss at every
threshold conditional on $Z\in G_j$ and the fit; a correction-sample
maximum is not sufficient. Set
\begin{equation}
\label{eq:overlap-pw-threshold}
 \widehat t_j:=\inf\left\{t\in\Lambda:
 \frac{\sum_{i\in\mathcal D_{\mathrm{cal}}:Z_i\in G_j}
  \ell(\mathcal C^{\widehat b}_{j,t}(X_i),Y_i)+B_j}{n_j+1}
 \le\alpha_j\right\},
\end{equation}
using $\widehat t_j=\lambda_{\max}$ if $n_j=0$ or the feasible set is empty.
Deploy
\begin{equation}
\label{eq:overlap-pw-set}
 \widehat{\mathcal C}_{\mathrm{PW}}(x)
 :=\bigcup_{j=1}^m\mathcal C^{\widehat b}_{j,\widehat t_j}(x)
 =\left\{y:(x,y)\in\textstyle\bigcup_jG_j,\;
 s(x,y)-\widehat b(x,y)\le T_{\max}(x,y)\right\},
\end{equation}
where $T_{\max}(x,y)=\max_{j:(x,y)\in G_j}\widehat t_j$ for eligible
candidates. This is candidate-dependent unless the local cutoffs coincide.

\begin{algorithm}[t]
\caption{Overlapping group-restricted union CRC}
\label{alg:overlap-union-crc}
\begin{algorithmic}[1]
\Require Correction fold $\mathcal D_{\mathrm{cal}}$, fixed groups $G_j$,
score $s$, independent shape $\widehat b$, bounds $B_j$, targets $\alpha_j$,
domain $\Lambda$.
\For{$j=1,\ldots,m$}
  \State $I_j\gets\{i\in\mathcal D_{\mathrm{cal}}:Z_i\in G_j\}$.
  \State $\widehat t_j\gets\Call{CRC-Cutoff}{I_j,\mathcal C^{\widehat b}_{j,\cdot},B_j,\alpha_j,\Lambda}$
  \Comment{Algorithm~\ref{alg:crc-event-sweep}; empty group defaults}
\EndFor
\State \Return $\widehat{\mathcal C}_{\mathrm{PW}}(x)=\bigcup_j
\mathcal C^{\widehat b}_{j,\widehat t_j}(x)$.
\end{algorithmic}
\end{algorithm}

For disjoint groups and $\widehat b\equiv0$, this is split Mondrian CRC.
The union inherits validity for any independently fitted shape, subject to
the restricted loss-path and endpoint assumptions.

\begin{theorem}
\label{thm:joint-fit-crc}
Suppose $\mathcal D_{\mathrm{fit}}$ is independent of the correction fold and
test observation, which are exchangeable conditional on the fit. For each $j$,
assume that the group-restricted loss paths are measurable,
non-increasing and right-continuous on the closed threshold domain
$\Lambda\subseteq\overline{\mathbb R}$ with
$\lambda_{\min}<\lambda_{\max}$, are bounded by $B_j$ on $G_j$,
and satisfy the endpoint condition of Theorem~\ref{thm:mondrian-crc},
namely that the loss at $\lambda_{\max}$ is at most $\alpha_j$
almost surely on $G_j$, conditional on the fit. Then
\begin{equation}
\label{eq:overlap-conditional-validity}
 \mathbb E\!\left[
 \ell(\widehat{\mathcal C}_{\mathrm{PW}}(X_{n+1}),Y_{n+1})
 \middle|Z_{n+1}\in G_j,\mathcal D_{\mathrm{fit}}\right]
 \le\alpha_j\qquad\text{a.s.}
\end{equation}
No consistency assumption on $\widehat b$ is required.
\end{theorem}

\begin{proof}
Condition on the fit. By within-group exchangeability
(Lemma~\ref{lem:within-group-exchangeability}),
Theorem~\ref{thm:mondrian-crc} applied to the $G_j$-restricted losses gives
$\mathbb E[\ell(\mathcal C^{\widehat b}_{j,\widehat t_j}(X_{n+1}),Y_{n+1})
\mid Z_{n+1}\in G_j,\mathcal D_{\mathrm{fit}}]\le\alpha_j$.
The inclusion $\mathcal C^{\widehat b}_{j,\widehat t_j}(x)
\subseteq\widehat{\mathcal C}_{\mathrm{PW}}(x)$ and loss monotonicity
establish the claim.
\end{proof}

\begin{remark}\label{rem:groupwise-not-pooled}
Overlapping groupwise bounds do not generally imply a marginal risk bound.
Including the full space as an additional calibration group yields one,
provided its bound and endpoint hypotheses hold.
\end{remark}

\subsection{Finite-sample conservativeness}
\label{subsec:efficiency}

Assume finite $\mathcal Y$. Conditional on $\mathcal D_{\mathrm{fit}}$,
write $\mathbb E_j^{\mathrm{fit}}$ for expectation over the correction fold
and an independent test observation with $Z\in G_j$. Define the
\emph{conditional expected} restricted and deployed risks and sizes by
\[
\begin{aligned}
 R_j^{\mathrm{slice}}&:=\mathbb E_j^{\mathrm{fit}}
   [\ell(\mathcal C^{\widehat b}_{j,\widehat t_j}(X),Y)],
 &R_j^{\mathrm{dep}}&:=\mathbb E_j^{\mathrm{fit}}
   [\ell(\widehat{\mathcal C}_{\mathrm{PW}}(X),Y)],\\
 S_j^{\mathrm{slice}}&:=\mathbb E_j^{\mathrm{fit}}
   [|\mathcal C^{\widehat b}_{j,\widehat t_j}(X)|],
 &S_j^{\mathrm{dep}}&:=\mathbb E_j^{\mathrm{fit}}
   [|\widehat{\mathcal C}_{\mathrm{PW}}(X)|].
\end{aligned}
\]
These are not empirical correction risks. Put
$A_j=\alpha_j-R_j^{\mathrm{slice}}\ge0$ and
$\Delta S_j=S_j^{\mathrm{dep}}-S_j^{\mathrm{slice}}\ge0$.
Under the single-group sandwich assumptions
(Theorem~\ref{thm:mondrian-crc-sandwich}),
\[
 A_j\le\mathbb E\!\left[
 \frac{B_j+\mathfrak J_j}{n_j+1}
 \middle|\mathcal D_{\mathrm{fit}},Z_{n+1}\in G_j\right],
\]
with the theorem's empty-cell convention. It vanishes when the right side
does, for example with growing group support and bounded jump multiplicity.

\begin{lemma}
\label{lem:risk-size}
For finite $C\subseteq D$ and monotone $\ell\in[0,B]$,
$0\le\ell(C,y)-\ell(D,y)\le B\mathbf1\{C\ne D\}\le B(|D|-|C|)$. Hence, for arbitrary
overlapping groups,
\begin{equation}
\label{eq:risk-size}
 0\le\alpha_j-R_j^{\mathrm{dep}}\le A_j+B\,\Delta S_j,
\end{equation}
so $R_j^{\mathrm{dep}}\to\alpha_j$ whenever $A_j\to0$ and $\Delta S_j\to0$.
\end{lemma}

\begin{proof}
Use $\mathcal C^{\widehat b}_{j,\widehat t_j}(X)
\subseteq\widehat{\mathcal C}_{\mathrm{PW}}(X)$, take
$\mathbb E_j^{\mathrm{fit}}$ of the pointwise inequality, and add $A_j$.
\end{proof}

The bound is sufficient, not sharp, and is most useful when size inflation
is small. Figure~\ref{fig:sim-overlap-gap} and the DDR tables illustrate
risk slack and its empirical counterpart. For two feature-defined groups
$G_j=\Gamma_j\times\mathcal Y$ with independent membership, the gap is
confined to their intersection. Writing $p_2=\mathbb P(G_2)$,
$\widehat t_\vee=\max(\widehat t_1,\widehat t_2)$, and
$\mathcal C_t^{\widehat b}(x)=\{y:s(x,y)-\widehat b(x,y)\le t\}$ gives
\[
0\le R_1^{\mathrm{slice}}-R_1^{\mathrm{dep}}
=p_2\,\mathbb E\!\left[
\ell(\mathcal C_{\widehat t_1}^{\widehat b}(X),Y)
-\ell(\mathcal C_{\widehat t_\vee}^{\widehat b}(X),Y)
\mid Z\in G_1\cap G_2\right]\le Bp_2,
\]
where the expectation averages over correction data and the test
observation conditional on $G_1\cap G_2$ and the fit. Similarly, $\Delta S_1\le p_2|\mathcal Y|$.
These bounds control the frequency of overlap, not score behavior.

\subsection{The limiting reconciliation gap}
\label{subsec:reconciliation-gap}

Suppose all local cutoffs converge in probability to the same interior
$K$ and $\|\widehat b-b^\star\|_\infty\to_{\mathbb P}0$ for a
deterministic measurable $b^\star$. Assume finite $\mathcal Y$, tight
population restricted risks
$\mathbb E[\ell(\mathcal C^{b^\star}_{j,K}(X),Y)\mid G_j]=\alpha_j$,
and all-candidate boundary stability,
\begin{equation}
\label{eq:boundary-stability}
 \mathbb P\!\left(\exists y\in\mathcal Y:\;
 s(X,y)-b^\star(X,y)=K\mid G_j\right)=0
 \qquad(j=1,\ldots,m).
\end{equation}
Write $\mathcal C^{b^\star}_{j,t}(x)
=\{y:(x,y)\in G_j,\ s(x,y)-b^\star(x,y)\le t\}$ and
$\mathcal C^{b^\star}_{\cup,K}(x)=\bigcup_j\mathcal C^{b^\star}_{j,K}(x)$.
The \emph{reconciliation gap} is
\begin{equation}
\label{eq:residual-gap}
 \Delta_j^\star:=\mathbb E\!\left[
 \ell(\mathcal C^{b^\star}_{j,K}(X),Y)
 -\ell(\mathcal C^{b^\star}_{\cup,K}(X),Y)\mid G_j\right]\ge0.
\end{equation}

\begin{theorem}
\label{thm:residual-gap}
Under Theorem~\ref{thm:joint-fit-crc}, finite $\mathcal Y$, and the alignment,
boundary-stability, and restricted-risk tightness conditions above, for every bounded
monotone loss and every $j$,
\begin{equation}
\label{eq:residual-limit}
 R_j^{\mathrm{dep}}\xrightarrow{\mathbb P}\alpha_j-\Delta_j^\star,
\end{equation}
so $R_j^{\mathrm{dep}}\to\alpha_j$ if and only if $\Delta_j^\star=0$.
\end{theorem}

\begin{proof}
Uniform shape convergence, cutoff alignment, and
\eqref{eq:boundary-stability} imply convergence in probability of all
finitely many candidate-admission indicators. Therefore the bounded
loss of the deployed set converges in $L^1$, conditional on $G_j$, to
$\ell(\mathcal C^{b^\star}_{\cup,K}(X),Y)$. Taking conditional
expectations over correction data yields
$R_j^{\mathrm{dep}}\to_{\mathbb P}
\mathbb E[\ell(\mathcal C^{b^\star}_{\cup,K}(X),Y)\mid G_j]
=\alpha_j-\Delta_j^\star$.
\end{proof}

\begin{corollary}
\label{cor:gap-zero}
$\Delta_j^\star=0$, hence $R_j^{\mathrm{dep}}\to\alpha_j$, if either the loss is
inclusion-type $\ell(C,y)=c(y)\mathbf1\{y\notin C\}$, or the groups are
feature-defined, $G_j=\Gamma_j\times\mathcal Y$. In the feature-defined case also
$\Delta S_j\to0$, so the bound \eqref{eq:risk-size} closes.
\end{corollary}

\begin{proof}
For inclusion-type loss, membership of the true label at common cutoff
$K$ is identical in its eligible group and in the union. For feature-defined
groups, $\mathcal C^{b^\star}_{j,K}(x)=
\mathcal C^{b^\star}_{\cup,K}(x)$ on $G_j$; boundary stability and
alignment then also imply $\Delta S_j\to0$.
\end{proof}

For inclusion-type losses one can construct an aligning shape via weighted
hinge regression. With $\Phi_j(z)=\mathbf1\{z\in G_j\}$, use a
\emph{fixed} baseline $q_0$ and
$q_\beta(z)=q_0(z)+\sum_{j=1}^m\beta_j\Phi_j(z)$,
$\beta\in\mathbb R^m$. A freely optimized unpenalized intercept would make
the objective below degenerate; an intercept may instead be fixed in $q_0$
or represented by a penalized full-population coordinate. Define
\begin{equation}
\label{eq:pinball-shape-objective}
 Q(\beta)=\mathbb E\!\left[
 c(Y)\{s(X,Y)-q_\beta(X,Y)\}_{+}
 +\sum_{j=1}^{m}\alpha_j\beta_j\Phi_j(X,Y)
 \right].
\end{equation}
Algorithm~\ref{alg:weighted-hinge-fit} uses its empirical analogue with
the same parameterization, taking $q_0\equiv0$; a nonzero fixed baseline is
equivalent to replacing the score $s$ by $s-q_0$.

\begin{corollary}
\label{prop:pinball-alignment}
Let $\ell(C,y)=c(y)\mathbf1\{y\notin C\}$ with
$0\le c(y)\le B$. Fit $\widehat b=q_{\widehat\beta}$ on the independent
fold using~\eqref{eq:pinball-shape-objective}
(Algorithm~\ref{alg:weighted-hinge-fit}), with the fixed-baseline
parameterization just defined. Assume $Q$ is finite in a neighborhood of a
finite, unconstrained minimizer $\beta^\star\in\mathbb R^m$, and put
$b^\star=q_{\beta^\star}$.
\begin{enumerate}[label=\textup{(\alph*)},leftmargin=*]
\item \emph{Exact finite-sample control.} Suppose, as in
Theorem~\ref{thm:joint-fit-crc}, that $\mathcal D_{\mathrm{fit}}$ is independent of
the exchangeable correction and test data and that the shaped restricted losses are
measurable, non-increasing, right-continuous, bounded by $B_j$ on $G_j$, and
satisfy the group-restricted endpoint condition. Then for every $j$ and every sample size,
$\mathbb E[\ell(\widehat{\mathcal C}_{\mathrm{PW}}(X_{n+1}),Y_{n+1})\mid
Z_{n+1}\in G_j,\mathcal D_{\mathrm{fit}}]\le\alpha_j$ almost surely, with no
further condition on the fitted shape.
\item \emph{Asymptotic tightness.} Assume in addition that
(i)~$\|q_{\widehat\beta}-q_{\beta^\star}\|_\infty\xrightarrow{\mathbb P}0$;
(ii)~for each $j$, the \emph{shaped true-label residual}
$s(X,Y)-b^\star(X,Y)$ has no atom at zero conditional on $Z\in G_j$,
and the population restricted risk
\[
 r_j(t;b^\star)=\mathbb E\!\left[
 c(Y)\mathbf1\{s(X,Y)-b^\star(X,Y)>t\}
 \middle|Z\in G_j\right]
\]
crosses its target transversally at $0$, namely
$r_j(-\varepsilon;b^\star)>\alpha_j>r_j(\varepsilon;b^\star)$
for every sufficiently small $\varepsilon>0$;
(iii)~the groupwise empirical correction risks at $\widehat b$ converge in
probability to $r_j(\cdot;b^\star)$ uniformly on a neighborhood of $0$, a
groupwise law of large numbers that holds under conditionally i.i.d.\ correction
data but not under exchangeability alone; and
(iv)~$0\in\operatorname{int}\Lambda$, $n_j\xrightarrow{\mathbb P}\infty$, and the
maximal threshold admits every group-eligible true label almost surely. Then
$\widehat t_j\xrightarrow{\mathbb P}0$ and
$R_j^{\mathrm{dep}}\xrightarrow{\mathbb P}\alpha_j$ for every $j$.
\end{enumerate}
\end{corollary}

The proof is in Appendix~\ref{app:pinball-alignment}. Its first-order
conditions, using atomlessness of the \emph{shaped true-label residual},
yield $r_j(0;b^\star)=\alpha_j$; the other hypotheses give
$\widehat t_j\to_{\mathbb P}0$ and deployed tightness without requiring
all-candidate boundary stability. The objective reduces, up to a constant,
to Kandinsky's population pinball formulation for $c\equiv1$ and equal
targets~\citep{bairaktari2025kandinsky}. The result does not prove shape
consistency or groupwise uniform laws of large numbers. In addition,
patternwise feasibility does not guarantee feasibility within the additive
shape class (Appendix~\ref{app:overlap-feasibility}); the construction does
not cover arbitrary structured losses.

For structured losses, alignment need not remove the reconciliation gap:

\begin{example}[Positive gap at equal thresholds]
\label{ex:residual-gap}
Let $\mathcal Y=\{0,1,2\}$ with ordinal distance-to-set loss
 $\ell(C,y)=\min_{c\in C}|c-y|$ ($\ell(\varnothing,y)=2$).
 Take label-defined groups $G_1=\{(x,y):y\in\{0,2\}\}$ and
$G_2=\{(x,y):y\in\{1,2\}\}$, uniform $Y$, and independent
 $\mathrm{Unif}[0,1]$ scores for each candidate. With $b^\star=0$
 and common cutoff $K=1/2$, the candidate admission events are independent,
 with restricted risks $(1,3/4)$. The union can admit label $1$ when the
 $G_1$ restricted set excludes true label $0$, reducing ordinal loss from $2$ to $1$.
 Thus $R_1^{\mathrm{dep}}=3/4$ and $\Delta_1^\star=1/4>0$ despite
 aligned cutoffs.
\end{example}

The next result shows that this gap cannot always be eliminated by
changing the aligned shape, even when all restricted risks are tight.
Its scope is candidate-only shapes with independent admission events.

\begin{theorem}
\label{thm:product-reconciliation-obstruction}
Suppose $Y$ is uniform on the stated finite label space,
independently of the candidate scores
$\{s(X,v)=U_v:v\in\mathcal Y\}$, where the $U_v$ are
independent $\mathrm{Unif}[0,1]$ variables. Consider a fixed
candidate-dependent shape $b(v)$ and a common limiting threshold $K$,
so candidate $v$ is admitted with probability
$a_v=\mathbb P\{U_v\le K+b(v)\}$. In particular, admission events
are independent across candidates. Restrict attention to shapes
whose population group-restricted risks equal their prescribed targets.
\begin{enumerate}[label=\textup{(\alph*)},leftmargin=*]
\item \emph{A family of structured losses.}
Let $\mathcal Y=\{0,1,2\}$,
$G_1=\{(x,y):y\in\{0,2\}\}$,
$G_2=\{(x,y):y\in\{1,2\}\}$, and, for $r\ge1$, let
\[
 \ell_r(C,y)=
 \begin{cases}
 \min_{v\in C}|v-y|^r,&C\ne\varnothing,\\
 B,&C=\varnothing,
 \end{cases}
 \qquad B=2^r.
\]
At the targets $\alpha_1=B/2$ and $\alpha_2=(B+1)/4$,
\begin{equation}
\label{eq:product-gap-power}
 \inf_{\substack{b:\;R_1^{\mathrm{slice}}=\alpha_1\\
                        R_2^{\mathrm{slice}}=\alpha_2}}
 \max_{j=1,2}\Delta_j^\star
 =\frac{(B-1)^3}{8(2B-1)^2}>0.
\end{equation}
\item \emph{Three overlapping groups.}
Let $\mathcal Y=\{0,1,2,3\}$, with ordinary
absolute-distance-to-set loss (empty-set loss $B=3$), and
$G_1=\{0,3\}$, $G_2=\{1,3\}$, and $G_3=\{2,3\}$,
where the displayed subsets designate the corresponding
label-defined events. At
$(\alpha_1,\alpha_2,\alpha_3)=(3/2,5/4,1)$,
\begin{equation}
\label{eq:product-gap-three}
 \inf_{\substack{b:\;R_j^{\mathrm{slice}}=\alpha_j\\j=1,2,3}}
 \max_{j=1,2,3}\Delta_j^\star=\frac9{80}>0.
\end{equation}
\end{enumerate}
Both infima are attained by candidate-wise cutoffs representable by
intercept-augmented group-indicator shapes $b_\beta$.
\end{theorem}

The positive infima apply to the specified uniform-label,
independent-candidate models with tight restricted risks, not to all
shaped or jointly calibrated predictors. Coupled admissions, covariate-dependent
shapes, or different label prevalences can change the conclusion
(Appendix~\ref{app:proof-product-obstruction}). The following equal-target
witness also applies at arbitrarily small positive levels.

\begin{corollary}
\label{cor:small-alpha-obstruction}
In the three-label model of
Theorem~\ref{thm:product-reconciliation-obstruction}(a) with $B=2^r$, take
equal targets $\alpha_1=\alpha_2=\alpha$ for any $0<\alpha<\tfrac12$. Over the
same product-form candidate-only class with tight restricted risks,
\begin{equation}
\label{eq:small-alpha-gap}
 \inf_{\substack{R_1^{\mathrm{slice}}=\alpha\\ R_2^{\mathrm{slice}}=\alpha}}
 \max_{j=1,2}\Delta_j^\star=\frac{B-1}{B}\,\alpha(1-2\alpha)>0,
\end{equation}
attained at $(a_0,a_1,a_2)=(1-2\alpha/B,\,1-2\alpha,\,1)$. The gap is
strictly positive for every $0<\alpha<\tfrac12$ and thus persists at
arbitrarily small positive normalized targets $\alpha/B$. Moreover, the sharp gap as a proportion of the target tends to
$(B-1)/B$ as $\alpha\downarrow0$.
At $\alpha=0.10$ the gap equals $0.040$, $0.060$, and $0.070$ for
$B=2,4,8$, respectively.
\end{corollary}

Thus the absolute gap tends to zero with $\alpha$, but its proportion of
the target need not. Exact one-sided validity of the union and its potential
conservativeness are distinct: the latter may persist in the restricted
product-form models, without implying a universal efficiency ranking.
Kandinsky's joint-quantile calibration is a related inclusion-type comparator,
not an equivalent guarantee~\citep{bairaktari2025kandinsky}.
Section~\ref{sec:deployment-aware} instead calibrates the whole deployed loss.

\section{Deployment-aware joint calibration}
\label{sec:deployment-aware}
\label{subsec:deployment-aware-crc}

The reconciliation gap of Section~\ref{sec:split-multigroup-estimator} shows that
the group-restricted risk can exceed the loss of the deployed union. This section
develops the second exact-validity route: calibrate the full deployed loss
directly, retain exact validity, and test whether removing the group-restricted constraints
reduces expected set size. The gain is not universal, because a single global
cutoff can also be less efficient than group-specific thresholds; the two
sections together set up the benefit-and-cost comparison carried out in the
experiments.

\paragraph{The joint nested family.}
Let $\mathcal Y$ be finite and let
\[
 \mathcal U(x)=\{y\in\mathcal Y:(x,y)\in\textstyle\bigcup_{j=1}^mG_j\}
\]
be the prespecified candidate universe. This is the same universe available
to the pointwise union~\eqref{eq:overlap-pw-set}; the construction also permits
a larger independently fixed universe, provided the endpoint requirement
below holds. Let $s$ be a score fixed independently of all fitting and
correction data. For a measurable, real-valued offset $h$ learned from an
independent fitting fold, set
\begin{equation}
\label{eq:joint-direct-family}
 D_t^h(x)=\{y\in\mathcal U(x):s(x,y)-h(x,y)\le t\},
 \qquad t\in\overline{\mathbb R}.
\end{equation}
The family is nested, with $D_{+\infty}^h=\mathcal U$.
 For $0<\alpha_j<B_j\le B$, assume
 $\ell(D_t^h(X),Y)\le B_j$ on $G_j$ for all $t$ and
\begin{equation}
\label{eq:joint-direct-endpoint}
 \ell(\mathcal U(X),Y)\le\alpha_j\quad\text{a.s. on }G_j.
\end{equation}
For miscoverage and distance-to-set losses, this holds when
 $\mathcal U(X)$ contains the true label on $G_j$.

With a correction fold $\mathcal D_{\mathrm{cal}}$, put
$\mathcal I_j=\{i\in\mathcal D_{\mathrm{cal}}:Z_i\in G_j\}$ and
$n_j=|\mathcal I_j|$. Define the \emph{direct} correction threshold
\begin{equation}
\label{eq:joint-direct-thresholds}
 \widehat t_j^{\mathrm{dir}}
 =\inf\left\{t\in\overline{\mathbb R}:
 \frac{\sum_{i\in\mathcal I_j}\ell(D_t^h(X_i),Y_i)+B_j}
 {n_j+1}\le\alpha_j\right\},
\end{equation}
with value $+\infty$ if the feasible set is empty (including $n_j=0$). Assume the empty set attains the maximal group loss, $\ell(\varnothing,Y)=B_j$
almost surely on $G_j$, as holds for miscoverage and distance-to-set losses.
Then, since $\alpha_j<B_j$, the criterion fails as $t\to-\infty$, so each
$\widehat t_j^{\mathrm{dir}}$ is finite or $+\infty$ and is never $-\infty$.
Deploy a \emph{single} prediction set
\begin{equation}
\label{eq:joint-direct-deployment}
 \widehat t_{\mathrm{dir}}=\max_{1\le j\le m}\widehat t_j^{\mathrm{dir}},
 \qquad
 \widehat{\mathcal C}_{\mathrm{dir}}(x)=D_{\widehat t_{\mathrm{dir}}}^h(x).
\end{equation}
Each criterion evaluates loss on the entire deployed family, not a
 restricted group set. Hence $\widehat t_{\mathrm{dir}}$ is finite when every
 group is feasible and equals $+\infty$ (with $D_{+\infty}^h=\mathcal U$) as soon
 as one group defaults, so $\widehat{\mathcal C}_{\mathrm{dir}}$ is well defined
 in either case.

Because the correction now scores the entire deployed set rather than a group-restricted
set, validity no longer follows from the union argument of
Section~\ref{sec:split-multigroup-estimator} and must be re-established. The next
theorem shows it still holds exactly for every group, again without any
requirement on the fitted offset.

\begin{theorem}
\label{thm:joint-direct-validity}
Suppose the fitting fold is independent of the exchangeable correction and
test observations. Assume the fixed group family, bounds, nestedness,
measurability, right-continuity of $t\mapsto\ell(D_t^h(X_i),Y_i)$, and
endpoint condition~\eqref{eq:joint-direct-endpoint} above, conditional on
the fitting fold. Then, for every measurable fitted offset $h$ and every
$j$ with $\mathbb P(Z\in G_j)>0$,
\begin{equation}
\label{eq:joint-direct-validity}
 \mathbb E\!\left[
  \ell(\widehat{\mathcal C}_{\mathrm{dir}}(X_{n+1}),Y_{n+1})
  \mid Z_{n+1}\in G_j,\mathcal D_{\mathrm{fit}}
 \right]\le\alpha_j\qquad\text{a.s.}
\end{equation}
No consistency assumption on the learned offset is required.
\end{theorem}

\begin{proof}
Condition on the fitting fold and fix $j$. The direct loss paths
$L_i^{\mathrm{dir}}(t)=\ell(D_t^h(X_i),Y_i)$ are bounded, non-increasing and
right-continuous. The within-$G_j$ leave-one-in argument of
Theorem~\ref{thm:mondrian-crc}, using the endpoint condition, gives
$\mathbb E[L_{n+1}^{\mathrm{dir}}(\widehat t_j^{\mathrm{dir}})
\mid Z_{n+1}\in G_j,\mathcal D_{\mathrm{fit}}]\le\alpha_j$.
By~\eqref{eq:joint-direct-deployment},
$D_{\widehat t_j^{\mathrm{dir}}}^h(x)\subseteq
D_{\widehat t_{\mathrm{dir}}}^h(x)$ for every $x$. Monotonicity transfers
this bound to the deployed loss. This holds for every $j$ without intersection refinement.
\end{proof}

The guarantee holds because the deployed threshold is the \emph{maximum} of the
group thresholds, so each group's set is a superset of its own validated set and
monotonicity preserves the bound. We now specify how the offset is fitted.

\paragraph{A concrete shape-fitting rule.}
Let $\mathcal H$ be a prespecified finite offset library, for example
 a coefficient grid in
\begin{equation}
\label{eq:joint-direct-shape-class}
 h_\theta(x,y)=\theta_0+\sum_{j=1}^m\theta_j\mathbf1\{(x,y)\in G_j\}
 +\sum_{q=1}^d\gamma_q\psi_q(x,y),
\end{equation}
with features $\psi_q$ fixed before fitting. The intercept sets the nominal
 cutoff at $t=0$; shifting an offset by a constant leaves its CRC-corrected
 deployed set unchanged. Appendix~\ref{app:joint-direct-oracle}
 explains this normalization.
For $n_{\mathrm{fit},j}=\#\{i\in\mathcal D_{\mathrm{fit}}:Z_i\in G_j\}$,
define
\[
 \widehat S_{\mathrm{fit}}(h)
 =\frac1{n_{\mathrm{fit}}}\sum_{i\in\mathcal D_{\mathrm{fit}}}
   |D_0^h(X_i)|,
 \qquad
 \widehat r_{\mathrm{fit},j}(h)
 =\frac1{n_{\mathrm{fit},j}}
 \sum_{i\in\mathcal D_{\mathrm{fit}}:Z_i\in G_j}
   \ell(D_0^h(X_i),Y_i).
\]
Given nonnegative fitting margins $\eta_j$, choose
\begin{equation}
\label{eq:joint-direct-fitting}
 \widehat h\in\operatorname*{arg\,min}_{h\in\mathcal H}
 \widehat S_{\mathrm{fit}}(h)
 \quad\text{subject to}\quad
 \widehat r_{\mathrm{fit},j}(h)\le\alpha_j-\eta_j
 \quad(j=1,\ldots,m),
\end{equation}
Use a prespecified tie-break on $\mathcal H$. If any fitting group is
 empty, no offset is feasible, or the fit optimization fails, return $\mathcal U$; otherwise perform the
 independent CRC correction. The finite library is enumerable; fitting must
 not use correction data. Conservative margins and a restricted size guarantee
 appear in Appendix~\ref{app:joint-direct-oracle}. They are unnecessary for
 validity and can preclude feasible fits when support is small.

\begin{algorithm}[t]
\caption{Deployment-aware joint CRC: fit, correct, and deploy}
\label{alg:joint-direct-crc}
\begin{algorithmic}[1]
\Require Fixed groups $G_j$, common candidate universe $\mathcal U$,
score $s$, bounds $B_j$, targets $\alpha_j$; correction fold
$\mathcal D_{\mathrm{cal}}$; either a prespecified offset $h$ or
an independent fitting fold $\mathcal D_{\mathrm{fit}}$, finite library $\mathcal H$ and margins $\eta_j$.
\If{fitting an offset from $\mathcal H$}
  \State $h\gets\Call{Finite-Library-Fit}{\mathcal D_{\mathrm{fit}},\mathcal H,\mathcal U,\alpha,\eta}$
        \Comment{Algorithm~\ref{alg:direct-library-fit}}
  \If{$h=\mathrm{FAIL}$}\Comment{empty fitting group, infeasible, or optimization failure}
     \State \Return $\widehat{\mathcal C}(x)=\mathcal U(x)$.\Comment{fallback: deploy the universe}
  \EndIf
\EndIf
\State Define $D_t^h(x)\gets\{v\in\mathcal U(x):s(x,v)-h(x,v)\le t\}$.
\For{$j=1,\ldots,m$}
  \State $I_j\gets\{i\in\mathcal D_{\mathrm{cal}}:(X_i,Y_i)\in G_j\}$.
  \State $\widehat t_j^{\mathrm{dir}}\gets
  \Call{CRC-Cutoff}{I_j,D_{\cdot}^h,B_j,\alpha_j,\overline{\mathbb R}}$
  \Comment{Loss on the \emph{whole} set $D_t^h$}
\EndFor
\State $\widehat t_{\mathrm{dir}}\gets\max_{j=1,\ldots,m}\widehat t_j^{\mathrm{dir}}$.
\State \Return $\widehat{\mathcal C}_{\mathrm{dir}}(x)
 =D_{\widehat t_{\mathrm{dir}}}^h(x)$.
\Statex \emph{If any group returns $+\infty$, the output is $\mathcal U(x)$.}
\end{algorithmic}
\end{algorithm}

\paragraph{A strict efficiency separation.}
The purpose of calibrating the deployed loss directly is efficiency, so we now
exhibit a model in which the direct set is provably smaller than \emph{every}
group-restricted union meeting the same group targets, even when the union is allowed
unequal group cutoffs.

\begin{prop}
\label{prop:joint-direct-size-separation}
Consider the three-label, two-group population model and power-distance
loss of Theorem~\ref{thm:product-reconciliation-obstruction}(a), with
$B=2^r\ge2$ and equal targets $0<\alpha<1/2$. Restrict the comparison to
fixed candidate-wise cutoffs with independent score coordinates
$U_0,U_1,U_2\sim\mathrm{Unif}[0,1]$.
Among all unions of group-restricted sets whose \emph{population restricted}
risks for $G_1=\{0,2\}$ and $G_2=\{1,2\}$ are each at most $\alpha$,
the minimum expected cardinality is
\begin{equation}
\label{eq:joint-direct-slice-opt}
 S_{\mathrm{slice}}^\star=3-2\alpha-\frac{2\alpha}{B}.
\end{equation}
In contrast, a single product-form prediction set with admission
probabilities
\begin{equation}
\label{eq:joint-direct-attainer}
 (a_0,a_1,a_2)
 =\left(1-\frac{2\alpha}{1+2\alpha(B-1)},\ 1-2\alpha,\ 1\right)
\end{equation}
has \emph{deployed} conditional risks $R_1^{\mathrm{dep}}=
R_2^{\mathrm{dep}}=\alpha$ and expected size
\begin{equation}
\label{eq:joint-direct-size}
 S_{\mathrm{direct}}
 =3-2\alpha-\frac{2\alpha}{1+2\alpha(B-1)}.
\end{equation}
Consequently,
\begin{equation}
\label{eq:joint-direct-strict-gain}
 S_{\mathrm{slice}}^\star-S_{\mathrm{direct}}
 =\frac{2\alpha(B-1)(1-2\alpha)}
 {B\{1+2\alpha(B-1)\}}>0.
\end{equation}
With a candidate-only offset $h(v)=a_v$ for the admission vector
in~\eqref{eq:joint-direct-attainer}, fixed independently of correction
data, the direct construction is the asymptotic limit of
\eqref{eq:joint-direct-thresholds}--\eqref{eq:joint-direct-deployment}:
the corrected thresholds converge to zero. This assertion concerns
fixed-shape correction; it does not assert that the finite-library
fitting rule~\eqref{eq:joint-direct-fitting} learns this offset. The comparison does not assert
optimality of the direct construction among all jointly valid prediction
sets or dominance over all feature-dependent union procedures.
\end{prop}

The separation exploits the slack left by group-restricted calibration, a set-size
difference that is distinct in units from the reconciliation risk gap of
Section~\ref{sec:split-multigroup-estimator}:
Appendix~\ref{app:joint-direct-size-proof} proves the comparison for
 unequal group thresholds. At $B=4$, $\alpha=0.1$, group-restricted calibration
 requires size at least $2.750$ (deployed risks $(0.040,0.100)$), whereas
 the constructed direct set has size $2.675$ with risks $(0.100,0.100)$.

\paragraph{Interpretation and limitations.}
The separation proves that group-restricted constraints can be unnecessarily strong,
 not that direct CRC always wins. A group with insufficient correction support
 sets $\widehat t_{\mathrm{dir}}=+\infty$, forcing the entire universe;
 the union localizes that fallback. The new fitting objective accommodates
 structured losses but may be nonconvex. All empirical comparisons therefore
 use the same candidate universe, group targets, and total data budget.
\section{Numerical illustrations}
\label{sec:numerical-illustrations}

We assess groupwise validity, local support, reconciliation, and direct
 calibration using risk and expected set size. Appendix~\ref{app:reproducibility}
 gives full designs and replication details.

\subsection{Heterogeneous groups}
\label{subsec:sim-group-conditional}

\begin{figure}[t]
\centering
\includegraphics[width=\textwidth]{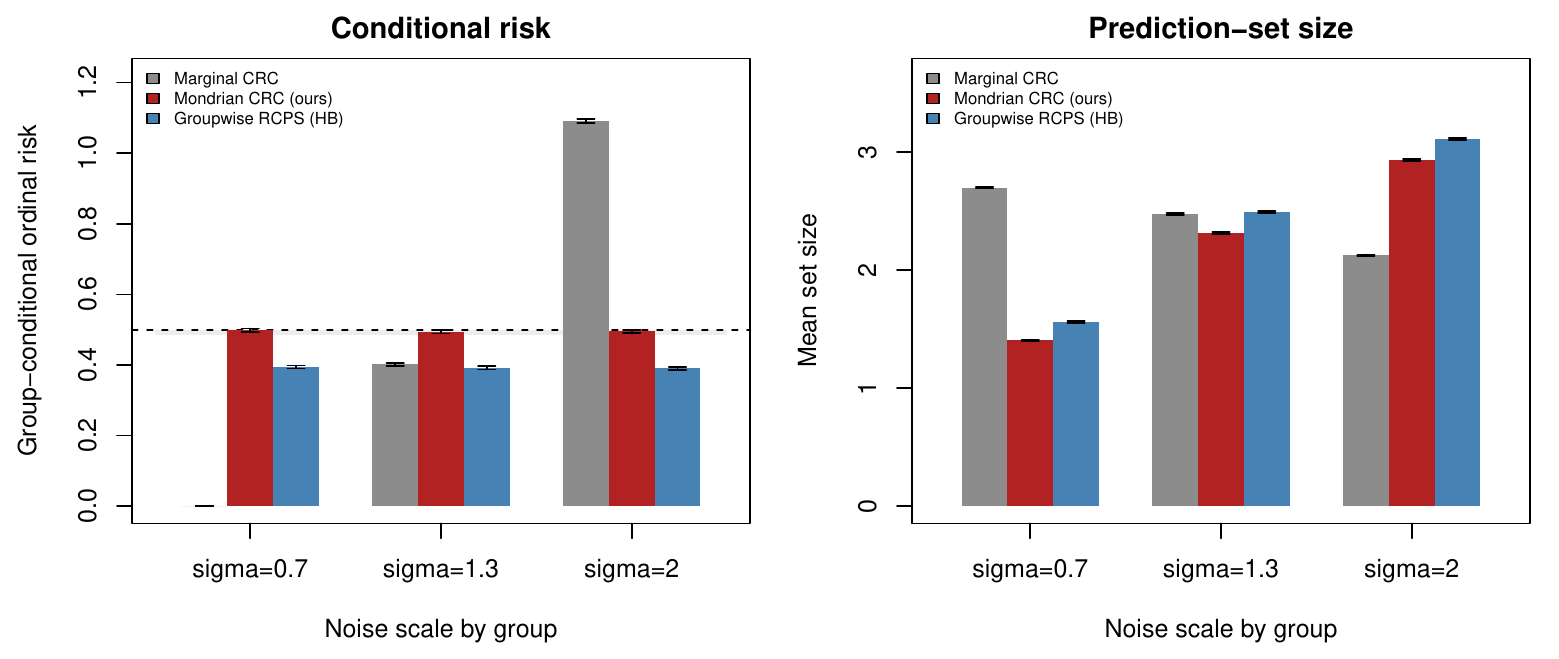}
\caption{Heterogeneous ordinal groups: conditional risk and mean set size.
The dashed line is $\alpha=0.5$; error bars are $1.96$ Monte Carlo standard
errors over $500$ replications. The shaded precision band for Mondrian CRC
is theoretical, not a confidence interval.}
\label{fig:sim-group-conditional}
\end{figure}

We compare marginal CRC, Mondrian CRC, and groupwise RCPS
 ~\citep{angelopoulos2024conformal,bates2021distribution}
 on three equally likely ordinal groups ($B=4$, $\alpha=0.5$).
 Across 500 replications, marginal CRC has pooled risk $0.497$ but risk
 $1.091$ in the hardest group. Mondrian CRC controls group risk near $0.5$
 with smaller sets than RCPS ($2.22$ versus $2.39$;
 Figure~\ref{fig:sim-group-conditional}). RCPS targets a stronger
 high-probability guarantee.

\subsection{Overlapping groups}
\label{subsec:sim-overlapping}

\begin{figure}[t]
\centering
\includegraphics[width=\textwidth]{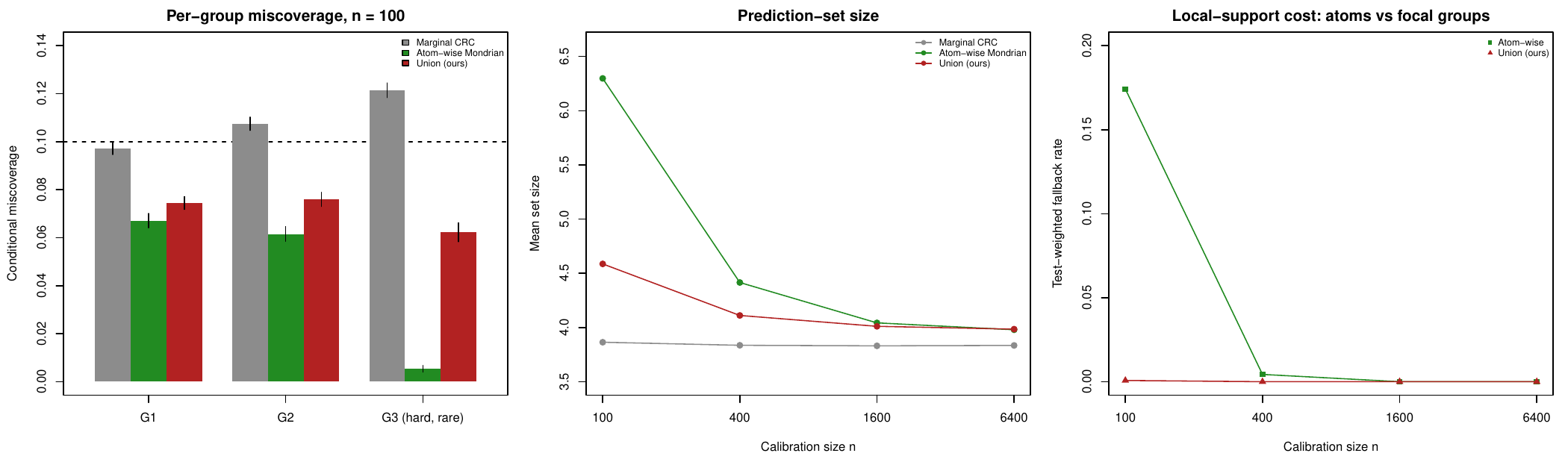}
\caption{Overlapping-label experiment (marginal CRC, eight-cell refinement, and
pointwise union): focal-group miscoverage at $n=100$
(left), set size across calibration sizes (center), and test-weighted
default frequency (right). The dashed line marks $\alpha=0.1$; left-panel
error bars are $1.96$ Monte Carlo standard errors over $500$ replications.
Kandinsky is shown separately in Figure~\ref{fig:sim-fitted-shape}.}
\label{fig:sim-overlap-coverage}
\end{figure}

For three binary labels, calibrate the three positive-label groups and
 the all-zero group under miscoverage ($\alpha=0.1$). Compare marginal CRC,
 eight-cell Mondrian refinement, and the pointwise union across $500$
 replications. At $n=100$, the rare focal group's miscoverage is $0.121$ for
 marginal CRC, $0.062$ for the union, and $0.005$ for refinement; the last has
 mean set size $6.30$ versus $4.59$ for the union, reflecting sparse-cell
 defaults. At $n=6400$, the union's pooled miscoverage is $0.096$ and $0.102$
 under independent and correlated labels, respectively
 (Figure~\ref{fig:sim-overlap-coverage}). The joint-quantile Kandinsky
 comparator~\citep{bairaktari2025kandinsky} is evaluated separately in the
 fitted-shape study below, not in these $500$ replications.

\paragraph{Fitted-hinge alignment as a supporting comparison.}
On the same model, fitted-hinge alignment
 (Corollary~\ref{prop:pinball-alignment}) reduces the restricted-to-deployed
 gap but enlarges sets relative to the unshaped union (size about $6.0$
 at $n=100$). This separate study uses $200$ replications per configuration
 ($120$ at $n=6400$). Independent CRC correction retains validity, whereas the
 uncorrected Kandinsky fit exceeds the nominal target at small $n$, with pooled
 miscoverage $0.133$ and $0.140$ under independent and correlated labels at
 $n=6400$. Figure~\ref{fig:sim-fitted-shape} and
 Appendix~\ref{app:reproducibility-fitted} give the full comparison.

Under Hamming-distance-to-set loss ($B=3$, $\alpha=0.5$),
 label-defined groups have aligned restricted risk $0.500$ but deployed union risk
 $0.362$, a residual gap of $0.138$. The gap vanishes for feature-defined
 groups (Figure~\ref{fig:sim-overlap-gap};
 Appendix~\ref{app:sim-overlap-reconciliation}).

\subsection{Deployment-aware calibration under structured loss}
\label{subsec:sim-deployment-aware}

Direct calibration can reclaim restricted-to-deployed slack but deploys a single
 common threshold, set by the most demanding group. The resulting coupling is
 harmless when groups make comparable threshold demands and costly when they do
 not. We assess this tradeoff in two structured-loss models and on DDR.

\begin{figure}[t]
\centering
\includegraphics[width=\textwidth]{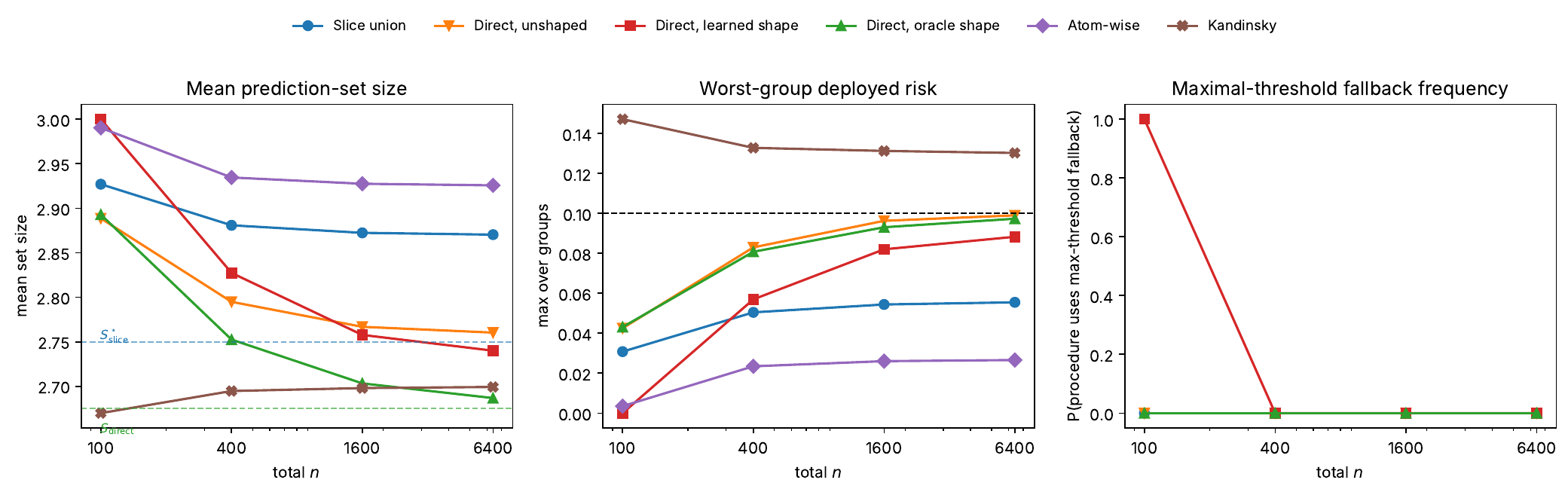}
\caption{Deployment-aware calibration under three-label ordinal loss
($\alpha=0.10$, groups $G_1=\{0,2\}$, $G_2=\{1,2\}$, $50/50$ split). Left: mean
prediction-set size against total $n$, with dashed lines at the group-restricted
population minimum $S^\star_{\mathrm{slice}}=2.750$ (the optimum over
candidate-wise product-form group-restricted constructions, not the large-$n$ limit of the
empirical union) and the constructed direct-reference size
$S_{\mathrm{direct}}=2.675$. Middle: worst-group deployed risk, dashed line at
$\alpha=0.10$; Kandinsky exceeds it at all $n$. Right: maximal-threshold fallback
frequency (a procedure-level event, not a per-instance full-set indicator),
nonzero only at $n=100$ for the learned variant, whose split correction fold falls
below the barrier. Averages over $600$ replications per configuration.}
\label{fig:sim-deployment}
\end{figure}

\paragraph{Three-label ordinal model.} We use Proposition~\ref{prop:joint-direct-size-separation}'s model
 ($B=4$, $\alpha=0.10$), with population sizes $2.750$ for optimal
 group-restricted calibration and $2.675$ for the constructed direct set.
 Across $n\in\{100,400,1600,6400\}$ and $600$ replications, direct
 calibration yields smaller sets at $n\ge400$. At $n=6400$ (50/50 split),
 sizes are $2.870$ (group-restricted union), $2.740$ (learned direct), and $2.687$
 (oracle direct), with worst-group risks $0.055$, $0.088$, and $0.097$.
 The oracle approaches its population value, remaining $0.012$ above it.
 Kandinsky's worst-group risk is $0.130$, so its smaller set is not
 comparable at achieved risk. At $n=100$ the full-budget group-restricted union and the
 unshaped and oracle direct methods do not default; only the learned direct
 method, which splits its budget, has per-group correction counts below
 $B/\alpha=40$ and defaults to the full universe, unlike the per-group union
 fallback. The gain persists across fit splits above the barrier
 (Figure~\ref{fig:sim-deployment}; Appendix~\ref{app:reproducibility-deployment}).

\begin{figure}[t]
\centering
\includegraphics[width=\textwidth]{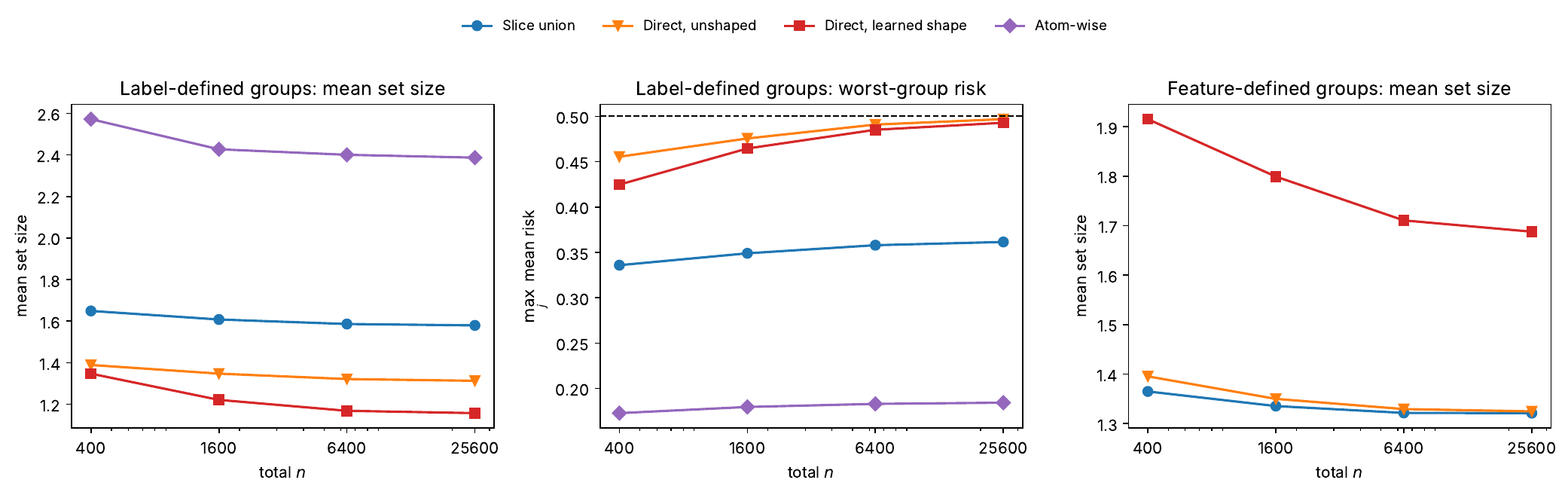}
\caption{Deployment-aware calibration under Hamming-to-set loss ($B=3$,
$\alpha=0.5$, $50/50$ split, $100$ replications). Left: mean set size under
label-defined groups $G_j=\{y_j=1\}$; the direct procedure reclaims the
reconciliation slack and is smaller than the group-restricted union. Center: worst-group
deployed risk (maximum over groups of the group-mean risk); the direct variants
use the budget up to $\alpha$ while the union remains near $0.36$. Right: under
feature-defined groups, with deployment eligibility matched across methods, the
unshaped direct procedure nearly matches the union; only the learned shape is larger.}
\label{fig:sim-hamming}
\end{figure}

\paragraph{Hamming loss on eight label vectors.} On $\mathcal Y=\{0,1\}^3$, use Hamming-to-set loss ($B=3$,
 $\alpha=0.5$) with the model of Appendix~\ref{app:sim-overlap-reconciliation}.
 For label-defined groups at $n=25{,}600$, the group-restricted union has size
 $1.579$ and worst-group risk $0.362$; direct unshaped and learned
 variants have sizes $1.313$ and $1.157$ at worst-group risks $0.497$ and
 $0.493$. The size reductions are $16.8\%$ and $26.7\%$, respectively.
 Under feature-defined groups the limiting gap is zero; holding deployment
 eligibility matched across methods, the unshaped direct size ($1.325$) then
 nearly matches the union ($1.321$), so the common threshold incurs little
 observed cost in this symmetric case (the paired difference is about $0.27\%$),
 and only the learned shape enlarges the set ($1.688$), the cost of fitting a
 shape when there is no slack to reclaim
 (Figure~\ref{fig:sim-hamming}; Appendix~\ref{app:reproducibility-direct-hamming}).
 The reported sizes are pooled expected cardinalities, averaged over all test
 points, compared at a common group-conditional risk target $\alpha=0.5$; the
 guarantee is on group-conditional risk, not pooled risk. The pooled risks
 reported in Appendix~\ref{app:reproducibility-direct-hamming} exceed $\alpha$
 without violating the groupwise guarantee, and under feature-defined membership
 about $1/8$ of test points belong to no group and receive the empty set under
 every method, so the pooled sizes partly reflect behavior outside the controlled
 groups. Thus reclaiming reconciliation slack yields smaller sets when a gap is
 present, while in the symmetric zero-gap case the common threshold is not itself
 costly; the coupling cost appears under group heterogeneity, as the DDR study
 below shows. This is an empirical pattern, not a universal dominance theorem.

\subsection{Diabetic-retinopathy grading}
\label{subsec:sim-realdata}

\begin{figure}[t]
\centering
\includegraphics[width=\textwidth]{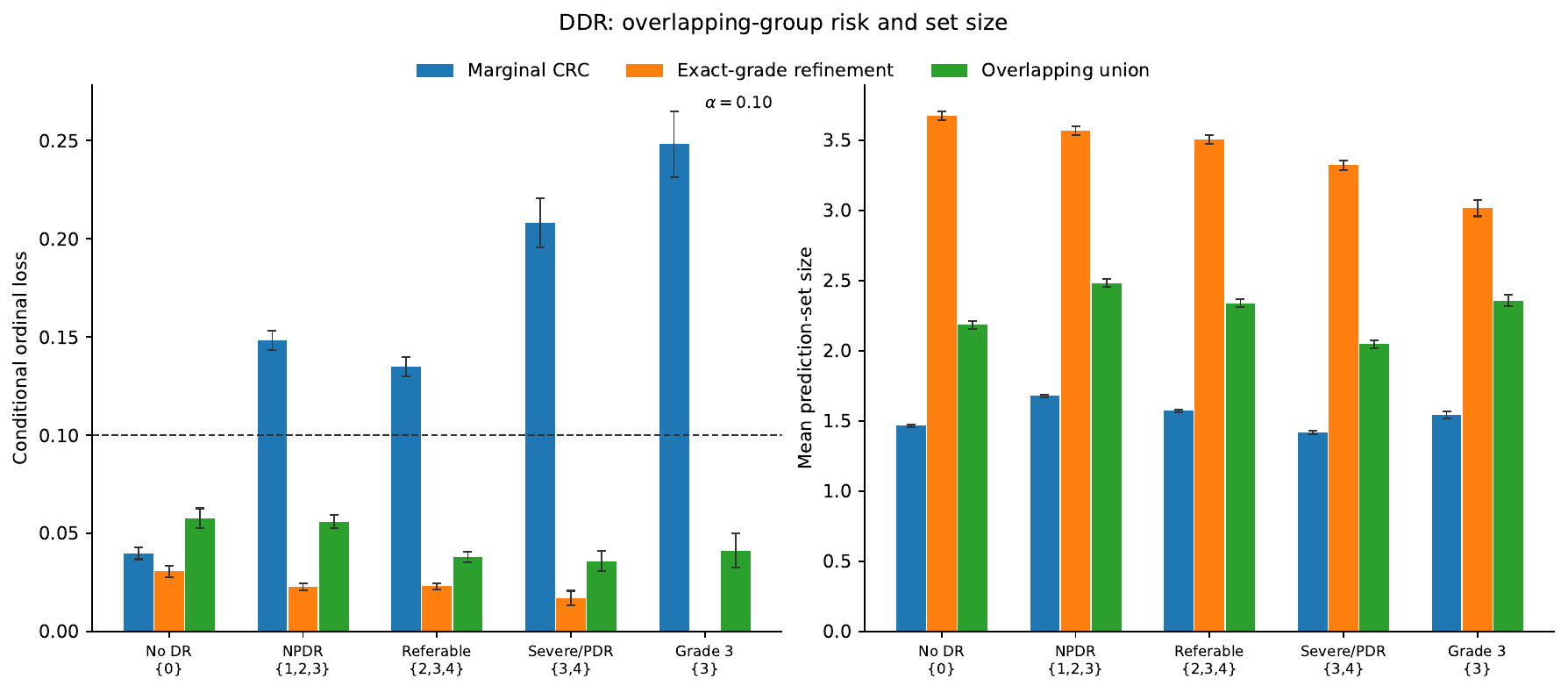}
\caption{DDR: conditional ordinal risk (left) and mean set size (right)
for marginal CRC, exact-grade refinement, and overlapping union.
The dashed line is $\alpha=0.1$; error bars are twice the Monte Carlo
standard error over $100$ split-and-score redraws of the fixed pool.
Detailed numerical results appear in Tables~\ref{tab:ddr-disjoint}
and~\ref{tab:ddr-overlap} of the appendix.}
\label{fig:sim-realdata}
\end{figure}

We analyze a historical pool of $2{,}428$ DDR images~\citep{li2019ddr} scored by
 an EfficientNet-B0 classifier, using randomized APS and ordinal
 distance-to-set loss ($B=4$) at $\alpha=0.1$. The classifier-training and
 deduplication provenance remains unverified, so the findings are descriptive
 fixed-pool comparisons. The original experiment uses $100$ calibration/test
 redraws.
 Marginal CRC has severe-group risk $0.208$ and grade-3 risk $0.248$.
 Disjoint Mondrian CRC reduces severe-group risk to $0.036$; the overlapping
 union controls risks at $0.058,0.056,0.038,0.036$ for
 $\{0\}$, $\{1,2,3\}$, $\{2,3,4\}$, and $\{3,4\}$, respectively.
 Its contribution is broader groupwise guarantees, not smaller sets than
 the three-stratum partition. Exact-grade refinement defaults for grade 3
 in every redraw (mean local count $23.5$); the union does not.
 Grade-3 risk $0.041$ is descriptive, not a separate union guarantee
 (Figure~\ref{fig:sim-realdata}; Appendix~\ref{app:reproducibility-ddr}).

\begin{figure}[t]
\centering
\includegraphics[width=\textwidth]{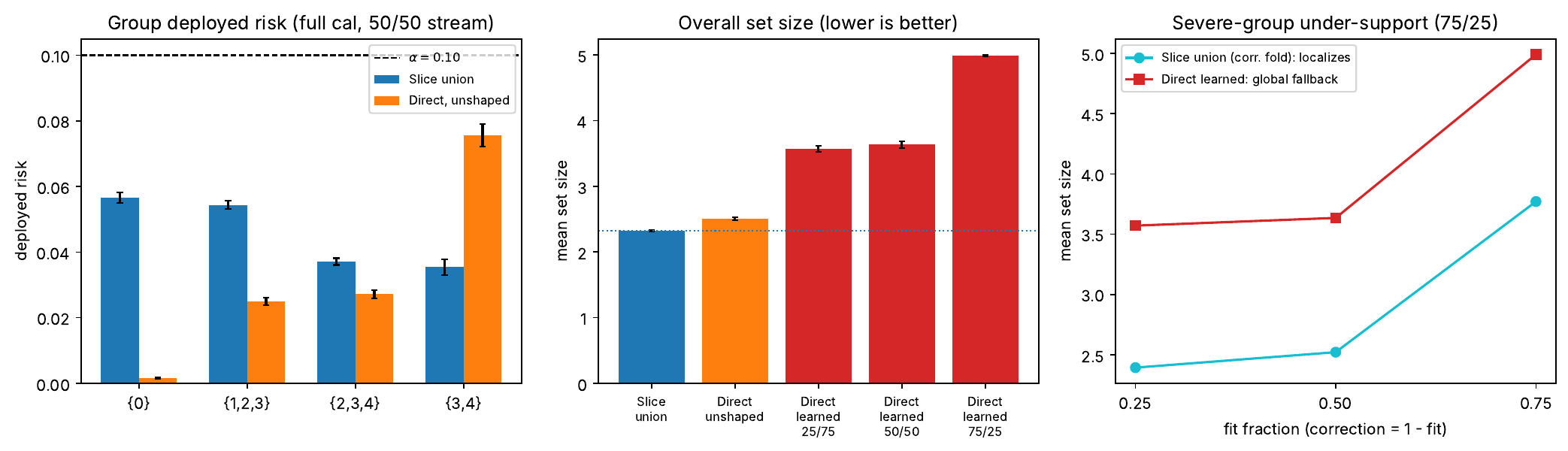}
\caption{Deployment-aware calibration on DDR (groups
$\{0\},\{1,2,3\},\{2,3,4\},\{3,4\}$, $\alpha=0.1$, $200$ redraws). Left: group
deployed risk; the direct cutoff is set by the severe group and over-covers
grade $0$. Center: mean set size, smallest for the group-restricted union. Right: at the
$75/25$ split the severe group's correction support falls below the barrier,
forcing the direct set toward the full universe while the union localizes the
default. Variation is over redraws of the fixed pool, not for independent future
patients.}
\label{fig:sim-realdata-direct}
\end{figure}

\paragraph{Deployment-aware calibration on DDR.} This real ordinal task is where
 the global-threshold coupling cost appears, because the groups have strongly
 heterogeneous threshold demands. On the same DDR pool and groups, $200$ paired
 redraws show that direct calibration is less efficient despite restricted-to-deployed
 gaps of $0.04$--$0.06$. Mean sizes are $2.322$ (union), $2.518$ (direct
 unshaped), and $3.57$--$4.99$ (direct learned); the unshaped direct method is
 $8.4\%$ larger. Its global cutoff is pinned near $0.999$ by the severe group,
 reducing grade-0 risk to $0.002$ through excess coverage. With a $75/25$
 fit/correction split, the severe group defaults in $98\%$ of redraws: the direct
 method nearly returns all five grades (mean size $4.99$), while the union
 localizes fallback (size $3.77$). The latter is therefore preferable on DDR
 (Figure~\ref{fig:sim-realdata-direct}; Appendix~\ref{app:reproducibility-ddr-direct}).
 Empirical risk on the fixed pool is not itself a proof of distribution-free
 validity, and patient-level independence is unverified.

\section{Conclusion}
\label{sec:conclusion}

This work develops a framework for group-conditional conformal
risk control under bounded monotone losses, encompassing
both disjoint and overlapping groups. Beyond establishing exact
finite-sample validity, we identify fundamental limitations
arising from local calibration support and the reconciliation
of overlapping prediction sets. In particular, the distinction
between calibrating group-restricted slices and directly
controlling deployed risk reveals a tradeoff between local
calibration flexibility and global threshold coupling, not
between validity and efficiency, since both strategies preserve
exact groupwise validity. Neither strategy is
uniformly preferable, and the effectiveness of each depends
on the structure of the loss, the groups, and the available
calibration data. These findings motivate the development of
more flexible calibration procedures that preserve exact
groupwise guarantees while reducing unnecessary conservativeness.
\clearpage
\appendix

\noindent\textbf{Appendix organization.}
Appendices~\ref{app:proofs-section2}--\ref{app:joint-direct-proofs}
contain mathematical proofs and deferred results.
Appendix~\ref{app:implementation-algorithms} collects implementable
algorithms and calibration conventions.
Appendices~\ref{app:reproducibility}--\ref{app:diagnostic-simulations}
document the synthetic studies, DDR application, and diagnostic simulations.

\section{Mondrian CRC: validity, tightness, and exact-risk refinements}
\label{app:proofs-section2}

Throughout, fix \(k\) with \(\mathbb P(G_{n+1}=k)>0\) and work conditionally on
\(\{G_{n+1}=k\}\). Write \(N_k:=n_k+1\) and
\(\mathcal I_k^+=\mathcal I_k\cup\{n+1\}\), let \(\mathcal E_k\) be the unordered
multiset of augmented within-group loss functions of
Section~\ref{subsec:mondrian-sandwich}, and let
\(\mathcal G_k:=\sigma(G_1,\ldots,G_{n+1})\). Define the augmented (leave-one-in)
empirical risk
\[
    R^0_k(\mu):=\frac{1}{N_k}\sum_{i\in\mathcal I_k^+}L_i^{(k)}(\mu),
    \qquad \mu\in\Lambda,
\]
a non-increasing, right-continuous function of \(\mu\), and the auxiliary thresholds
\begin{equation}
    \widehat\lambda'_k:=\inf\bigl\{\mu\in\Lambda:R^0_k(\mu)\le\alpha_k\bigr\},
    \qquad
    \widehat\lambda^+_k:=\inf\bigl\{\mu\in\Lambda:
        R^0_k(\mu)\le\alpha_k-\tfrac{B_k}{N_k}\bigr\},
    \label{eq:loo-thresholds}
\end{equation}
each \(\lambda_{\max}\) when the feasible set is empty. The first drives the upper
bound via \(\widehat\lambda'_k\le\widehat\lambda_k\), the second the lower bound via
\(\widehat\lambda_k\le\widehat\lambda^+_k\); both are symmetric functionals of
\(\mathcal E_k\), whereas the deployed threshold \(\widehat\lambda_k\) is not.

\subsection{Proof of Lemma~\ref{lem:within-group-exchangeability}}
\label{app:proof-within-group-exchangeability}

\begin{proof}
Let \(\pi\) be any permutation of \([n+1]\) that permutes the indices in
\(\mathcal I_k^+\) and fixes every other index. By exchangeability of
\((Z_i)_{i=1}^{n+1}\), \((Z_{\pi(i)})_{i=1}^{n+1}\) has the same law. The conditioning
event, that the membership vector equals a fixed realisation with \(G_{n+1}=k\), is
invariant under \(\pi\): every index in \(\mathcal I_k^+\) carries label \(k\), so
permuting group-\(k\) observations changes neither the membership vector nor
\(\mathcal I_k^+\). Hence the conditional law of \((Z_i)_{i\in\mathcal I_k^+}\) given
\(\mathcal G_k\) and \(\{G_{n+1}=k\}\) is invariant under every such \(\pi\).
\end{proof}

\subsection{Proof of Theorem~\ref{thm:mondrian-crc}}
\label{app:proof-validity}

\begin{proof}
Suppose the feasible set in \eqref{eq:mondrian-threshold} is nonempty and let \(\mu\)
belong to it, so \(\tfrac{n_k}{N_k}\widehat R^0_{n,k}(\mu)+\tfrac{B_k}{N_k}\le\alpha_k\).
Since \(L_{n+1}^{(k)}(\mu)\le B_k\),
\[
    R^0_k(\mu)
    =\frac{n_k}{N_k}\widehat R^0_{n,k}(\mu)+\frac{L_{n+1}^{(k)}(\mu)}{N_k}
    \le\frac{n_k}{N_k}\widehat R^0_{n,k}(\mu)+\frac{B_k}{N_k}\le\alpha_k,
\]
so \(\mu\) is feasible for \(\widehat\lambda'_k\) and \(\widehat\lambda'_k\le\widehat\lambda_k\).
If the feasible set is empty then \(\widehat\lambda_k=\lambda_{\max}\) and the
inequality is trivial; the endpoint condition \eqref{eq:endpoint-condition} gives
\(R^0_k(\lambda_{\max})\le\alpha_k\), so the feasible set defining \(\widehat\lambda'_k\)
is never empty. Monotonicity of \(L_{n+1}^{(k)}\) yields
\(L_{n+1}^{(k)}(\widehat\lambda_k)\le L_{n+1}^{(k)}(\widehat\lambda'_k)\).

Since \(\Lambda\) is closed, \(\widehat\lambda'_k\in\Lambda\), and by right-continuity
of \(R^0_k\), \(R^0_k(\widehat\lambda'_k)\le\alpha_k\). Being a symmetric functional of
\(\mathcal E_k\), \(\widehat\lambda'_k\) is measurable with respect to
\((\mathcal G_k,\mathcal E_k)\); by Lemma~\ref{lem:within-group-exchangeability},
conditional on \((\mathcal G_k,\mathcal E_k)\) and \(\{G_{n+1}=k\}\) the test function 
\(L_{n+1}^{(k)}\) is a uniform draw from the \(N_k\) elements of \(\mathcal E_k\). Hence
\begin{equation}
    \mathbb E\bigl[L_{n+1}^{(k)}(\widehat\lambda'_k)\,\big|\,\mathcal G_k,\mathcal E_k,G_{n+1}=k\bigr]
    =\frac{1}{N_k}\sum_{i\in\mathcal I_k^+}L_i^{(k)}(\widehat\lambda'_k)
    =R^0_k(\widehat\lambda'_k)\le\alpha_k.
    \label{eq:loo-averaging}
\end{equation}
Finally, on \(\{G_{n+1}=k\}\),
\(\mathcal C^0_{k,\widehat\lambda_k}(X_{n+1})\subseteq\mathcal C_{\widehat\lambda}(X_{n+1})\).
Set-monotonicity gives
\(\ell(\mathcal C_{\widehat\lambda}(X_{n+1}),Y_{n+1})\le L_{n+1}^{(k)}(\widehat\lambda_k)\).
Combining,
\[
    \mathbb E\bigl[\ell(\mathcal C_{\widehat\lambda}(X_{n+1}),Y_{n+1})\,\big|\,G_{n+1}=k\bigr]
    \le
    \mathbb E\bigl[L_{n+1}^{(k)}(\widehat\lambda'_k)\,\big|\,G_{n+1}=k\bigr]
    \le\alpha_k. \qedhere
\]
\end{proof}

\subsection{Proof of Theorem~\ref{thm:mondrian-crc-sandwich}}
\label{app:proof-sandwich}

\begin{proof}
The upper bound is the chain above. For the lower bound, let
\(\mu\) be feasible for \(\widehat\lambda_k^+\), so
\(R_k^0(\mu)\le\alpha_k-B_k/N_k\). Since \(L_{n+1}^{(k)}(\mu)\ge0\),
\(\tfrac{n_k}{N_k}\widehat R_{n,k}^0(\mu)+\tfrac{B_k}{N_k}
=R_k^0(\mu)-\tfrac{L_{n+1}^{(k)}(\mu)}{N_k}+\tfrac{B_k}{N_k}\le\alpha_k\),
so every point feasible for \(\widehat\lambda_k^+\) is feasible for
\eqref{eq:mondrian-threshold}, whence \(\widehat\lambda_k\le\widehat\lambda_k^+\) and
\(L_{n+1}^{(k)}(\widehat\lambda_k)\ge L_{n+1}^{(k)}(\widehat\lambda_k^+)\).

The threshold \(\widehat\lambda_k^+\) is a symmetric functional of \(\mathcal E_k\), so
\eqref{eq:loo-averaging} gives
\begin{equation}
    \mathbb E\!\left[L_{n+1}^{(k)}(\widehat\lambda_k^+)\,\middle|\,\mathcal G_k,\mathcal E_k,G_{n+1}=k\right]
    =R_k^0(\widehat\lambda_k^+).
    \label{eq:plus-averaging}
\end{equation}
If the feasible set is empty, \(R_k^0(\widehat\lambda_k^+)=R_k^0(\lambda_{\max})>\alpha_k-B_k/N_k\);
if \(\widehat\lambda_k^+=\lambda_{\min}\), \eqref{eq:lower-endpoint-condition} gives the
same with \(\ge\). Otherwise every \(\mu<\widehat\lambda_k^+\) is infeasible, so
\(R_k^0((\widehat\lambda_k^+)^-)\ge\alpha_k-B_k/N_k\), while the drop at
\(\widehat\lambda_k^+\) is
\(R_k^0((\widehat\lambda_k^+)^-)-R_k^0(\widehat\lambda_k^+)
=\tfrac1{N_k}\sum_{i\in\mathcal I_k^+}J(L_i^{(k)},\widehat\lambda_k^+)
\le\mathfrak J_k/N_k\),
with \(J(L,\mu):=L(\mu^-)-L(\mu)\) and \(\mathfrak J_k\) as in
\eqref{eq:total-jump-mass}. Hence
\(R_k^0(\widehat\lambda_k^+)\ge\alpha_k-(B_k+\mathfrak J_k)/N_k\)
in all cases. The tower property with \eqref{eq:plus-averaging} gives
the sharper lower bound; \(\mathfrak J_k\le M_kB_k\) gives the count-based
bound. The lower bound transfers to the deployed loss only when the
losses coincide (Remark~\ref{rem:slice-vs-deployed}).
\end{proof}

\subsection{Exact risk under two-valued functions}
\label{app:exact-two-valued}
\label{sec:exact}

The sandwich becomes exact when the one-coordinate functions are common-height steps.
Fix \(k\) and, conditional on \(\mathcal G_k\) and \(\{G_{n+1}=k\}\), assume:
\begin{enumerate}[label=(P\arabic*),leftmargin=*]
\item there exist \(S_i\), \(i\in\mathcal I_k^+\), with
\(L_i^{(k)}(\mu)=B_k\mathbf 1\{S_i>\mu\}\) for \(\mu\in\Lambda\);
\item \(\Lambda\) is an interval whose interior contains the support of the \(S_i\),
with \(\lambda_{\min}<S_i<\lambda_{\max}\) almost surely;
\item the augmented values \(\{S_i:i\in\mathcal I_k^+\}\) are almost surely distinct.
\end{enumerate}
Under {\rm(P1)}--{\rm(P3)} the functions jump at distinct locations, so \(M_k=1\) almost
surely.

\begin{lemma}
\label{lem:two-valued-routes}
Condition {\rm(P1)} holds in either case:
\begin{enumerate}[label=(\alph*),leftmargin=*]
\item \(\ell(\mathcal C,y)=c(y)\mathbf 1\{y\notin\mathcal C\}\) with \(c(y)=c_k>0\)
whenever \(g(x,y)=k\), and \(B_k=c_k\);
\item for every \(x,k\) the candidate set \(\mathcal Y_k(x)\) is empty or a singleton,
and \(\ell(\{y\},y)=0\), \(\ell(\varnothing,y)=B_k\) whenever \(g(x,y)=k\).
\end{enumerate}
\end{lemma}

Route (b) applies to label-determining groups, such as \(g(x,y)=y\). For \(m\ge0\) and \(B>\alpha>0\), define the rank-grid slack
\begin{equation}
    \Delta_m(\alpha,B):=\alpha-\frac{B}{m+1}\left\lfloor\frac{(m+1)\alpha}{B}\right\rfloor
    \in\left[0,\frac{B}{m+1}\right).
    \label{eq:rank-grid-slack}
\end{equation}
Write \(q_k:=\alpha_k/B_k\), \(\kappa_k:=\lfloor N_kq_k\rfloor\); for \(\kappa_k\ge1\)
let \(S_{(1)}>\cdots>S_{(n_k)}\) be the decreasing calibration order statistics.

\begin{prop}
\label{prop:exact-two-valued}
Under the assumptions of Theorem~\ref{thm:mondrian-crc} and {\rm(P1)}--{\rm(P3)},
\(\widehat\lambda_k=S_{(\kappa_k)}\) when \(\kappa_k\ge1\) and
\(\widehat\lambda_k=\lambda_{\max}\) when \(\kappa_k=0\), and
\begin{equation}
    \mathbb E\!\left[L_{n+1}^{(k)}(\widehat\lambda_k)\,\middle|\,\mathcal G_k,G_{n+1}=k\right]
    =\frac{B_k}{N_k}\left\lfloor\frac{N_k\alpha_k}{B_k}\right\rfloor
    =\alpha_k-\Delta_{n_k}(\alpha_k,B_k).
    \label{eq:exact-risk}
\end{equation}
\end{prop}

\begin{proof}
If \(m:=n_k=0\) then \(N_k=1\), \(\kappa_k=0\), the calibrated threshold is
\(\lambda_{\max}\), and {\rm(P1)}--{\rm(P2)} give \(L_{n+1}^{(k)}(\lambda_{\max})=0\), so
\eqref{eq:exact-risk} holds. Assume \(m\ge1\). By {\rm(P1)},
\(\widehat R_{n,k}^0(\mu)=(B_k/m)\#\{i\in\mathcal I_k:S_i>\mu\}\), so
\eqref{eq:mondrian-threshold} is equivalent to
\(\#\{i\in\mathcal I_k:S_i>\mu\}\le N_kq_k-1\); the left side being integer-valued and
\(\lfloor N_kq_k-1\rfloor=\kappa_k-1\), the feasible set is empty when \(\kappa_k=0\)
(then \(\widehat\lambda_k=\lambda_{\max}\) and the loss is zero). For \(\kappa_k\ge1\),
since \(q_k<1\) we have \(\kappa_k\le m\), and by {\rm(P2)}--{\rm(P3)},
\(\#\{i\in\mathcal I_k:S_i>\mu\}\le\kappa_k-1\) iff \(\mu\ge S_{(\kappa_k)}\), so
\(\widehat\lambda_k=S_{(\kappa_k)}\). Let \(R\) be the rank of \(S_{n+1}\) among the
\(N_k\) augmented values; \(\{S_{n+1}>S_{(\kappa_k)}\}=\{R\le\kappa_k\}\), and
conditional exchangeability with {\rm(P3)} makes \(R\) uniform on \(\{1,\ldots,N_k\}\),
so
\(\mathbb E[L_{n+1}^{(k)}(\widehat\lambda_k)\mid\mathcal G_k,G_{n+1}=k]
=B_k\mathbb P(R\le\kappa_k\mid\cdot)=B_k\kappa_k/N_k\), which is \eqref{eq:exact-risk}.
When the deployed and one-coordinate losses coincide the same calculation gives the
deployed risk; otherwise monotonicity gives only that it is no larger.
\end{proof}

\begin{remark}
\label{rem:singleton-slice}
The identity concerns restricted risk; it transfers to deployed risk only when the
two losses coincide.
\end{remark}

\begin{corollary}
\label{cor:minimax}
Fix \(m\ge1\) and \(\alpha_k\in(0,B_k)\), and let \(\mathcal M(m,B_k)\) be the class of
exchangeable laws of \(N=m+1\) augmented functions satisfying {\rm(P1)}--{\rm(P3)} with
common height \(B_k\). Then
\[
    \inf_A\sup_{P\in\mathcal M(m,B_k)}
    \left\{\alpha_k-\mathbb E_P[L_N(A(L_1,\ldots,L_m))]\right\}
    =\Delta_m(\alpha_k,B_k),
\]
the infimum over deterministic symmetric threshold procedures valid at level
\(\alpha_k\) over \(\mathcal M(m,B_k)\); the Mondrian threshold attains it.
\end{corollary}

\begin{proof}
By Proposition~\ref{prop:exact-two-valued} the Mondrian threshold has slack exactly
\(\Delta_m(\alpha_k,B_k)\) at every \(P\in\mathcal M(m,B_k)\), bounding the minimax
value above. For the reverse, Lemma~\ref{lem:witness-realizability} realizes the
one-jump witness \(P_t\) of Proposition~\ref{thm:minimax-lower-bound} by a bounded
monotone loss with distinct jump locations, placing it in \(\mathcal M(m,B_k)\);
since that witness uses validity only at \(P_t\), every deterministic symmetric
procedure valid over \(\mathcal M(m,B_k)\) has worst-case slack at least
\(\Delta_m(\alpha_k,B_k)\). The restriction to symmetric rules is necessary because an
order-dependent rule can use the sample ordering as an implicit randomization device.
\end{proof}

\subsection{Endpoint and jump conventions}
\label{app:sandwich-conditions}

Condition~\eqref{eq:lower-endpoint-condition} holds if every augmented
loss equals $B_k$ at $\lambda_{\min}$, as when this threshold gives
the empty set and $\ell(\varnothing,Y)=B_k$. For $n_k>0$, a sufficient calibration-only condition is
\[
 \overline L_k\ge\frac{(n_k+1)\alpha_k-B_k}{n_k},
\]
where $\overline L_k$ is the calibration mean at $\lambda_{\min}$;
the test loss is nonnegative.

The jump multiplicity $M_k$ counts the augmented loss functions jumping at the
same threshold, not the number of jumps per function. For one-jump restricted losses each $L_i^{(k)}$ jumps once, at its true-label
threshold $S_i$, so $M_k=1$ holds precisely when the augmented jump locations
$\{S_i:i\in\mathcal I_k^+\}$ are almost surely distinct, the no-ties
condition {\rm(P3)} of Appendix~\ref{app:exact-two-valued}. Continuity of the
marginal score does not imply this under arbitrary exchangeability; a sufficient
condition is that the augmented scores be conditionally i.i.d.\ from an atomless
law, while discrete grids or repeated values can induce ties. The relevant
coincidences are among the jump locations $S_i$, not among the candidate scores
$s(X_i,y)$ at non-true labels.
For iid observations $n_k\sim\mathrm{Binomial}(n,p_k)$, and when $M_k=1$,
\[
  2B_k\,\mathbb E[(n_k+1)^{-1}]
  =\frac{2B_k\{1-(1-p_k)^{n+1}\}}{(n+1)p_k}.
\]
For random $M_k$, the expectation of $(M_k+1)B_k/(n_k+1)$ must be retained
without generally factoring it.

\section{Local-support and rank-granularity results}\label{app:bounds-and-ranks}

\subsection{Rank-granularity analysis}
\label{app:rank-granularity}

We fix $m=n_k$, $N=m+1$ and the exchangeable loss-function class
$\mathcal P(m,M,B)$ of Section~\ref{sec:lower-bounds}. All rules below
are deterministic and permutation-invariant unless stated otherwise.
Witnesses require zero loss at $\lambda_{\max}$.
For a finite ordered grid, a strict downward jump at $\mu>\lambda_{\min}$
means $L(\mu^-) > L(\mu)$, where $L(\mu^-)$ is the value at the
immediately preceding grid point. At a left-accumulation point (including $+\infty$ when
$\Lambda$ has arbitrarily large finite thresholds) it is the
usual left limit along $\Lambda$. The same convention is used in the main-text
loss-curve class $\mathcal P(m,M,B)$ and in the finite-grid
counterexample below.

Recall $\Delta_m(\alpha,B)$ from \eqref{eq:rank-grid-slack}, the distance
to the preceding rank-grid risk.

\begin{prop}
\label{thm:minimax-lower-bound}
Let \(m\ge1\), \(1\le M\le N\), and suppose \(\Lambda\) contains
\(N\) distinct interior thresholds
\(\lambda_{\min}<t_1<\cdots<t_N<\lambda_{\max}\).
Let \(P_t\) be the exchangeable law obtained by uniformly permuting
the \(N\) loss paths
\(L_i(\mu)=B\mathbf1\{\mu<t_{\pi(i)}\}\).
Then \(P_t\in\mathcal P(m,1,B)\subseteq\mathcal P(m,M,B)\).
Every deterministic symmetric rule \(A\) valid at level \(\alpha\)
over \(\mathcal P(m,M,B)\) satisfies
\[
 \mathbb E_{P_t}\left[L_N\{A(L_1,\ldots,L_m)\}\right]
 \le\frac{B}{N}\left\lfloor\frac{N\alpha}{B}\right\rfloor
 =\alpha-\Delta_m(\alpha,B).
\]
In fact, the expected loss under \(P_t\) is always an integer multiple
of \(B/N\), even for a rule required to be valid only under \(P_t\).
\end{prop}

\begin{remark}
\label{rem:asymmetric-randomization}
The rank-grid restriction is nonrandomized and symmetric. With an
independent random coin, interpolation between adjacent calibration
rank thresholds attains expected loss \(\alpha\) in the common-height,
distinct one-jump submodel. A deterministic rule depending on the
arrival order of distinct calibration paths may similarly exploit that
order as an implicit randomization device. Such rules are outside the
class in Proposition~\ref{thm:minimax-lower-bound} and
Corollary~\ref{cor:minimax}. The statement is a minimax lower bound
\emph{within the indicated procedure class}, not a lower bound on
slack for every possible conformal method.
\end{remark}

\begin{lemma}
\label{lem:witness-realizability}
The one-jump witness \(P_t\), the tied-jump witness in
Proposition~\ref{thm:minimax-lower-bound-general-M}, and the
single-spike witnesses in Section~\ref{sec:lower-bounds} are
realizable as within-group restricted-loss paths of a finite-label
prediction-set problem with a fixed real-valued score and bounded
monotone cost-sensitive miscoverage loss, provided the threshold
domain $\Lambda$ has finite endpoints, so that every jump location used
by the witnesses, including $\lambda_{\max}$ for the support-barrier
spike, is finite. The abstract lower bounds of
Proposition~\ref{thm:small-group-nontriviality},
Proposition~\ref{thm:minimax-lower-bound}, and
Proposition~\ref{thm:minimax-lower-bound-general-M} hold on their
loss-curve class for any $\Lambda$; only this realization as a concrete
prediction-set problem uses finite endpoints. For integer \(B\ge1\), the same
witnesses can be realized using ordinal absolute-error loss on
\(\{0,\ldots,B\}\).
\end{lemma}

\begin{prop}
\label{thm:minimax-lower-bound-general-M}
Fix \(m\ge0\), \(1\le M\le N=m+1\), and assume that \(\Lambda\) contains
\(G=\lfloor N/M\rfloor\) distinct interior points
\(\tau_1<\cdots<\tau_G\).
Assign the \(N\) indices uniformly to \(G\) blocks of size \(M\)
and \(r=N-GM\) inactive positions. In each active block \(g\), let
\(L_i(\mu)=B\mathbf1\{\mu<\tau_g\}\); in an inactive position,
let \(L_i\equiv0\). Denote the resulting exchangeable law by
\(P_\tau^{(M)}\). Then \(P_\tau^{(M)}\in\mathcal P(m,M,B)\), and every
deterministic symmetric procedure \(A\) valid at level \(\alpha\)
over \(\mathcal P(m,M,B)\) satisfies
\begin{equation}
 \mathbb E_{P_\tau^{(M)}}\left[L_N\{A(L_1,\ldots,L_m)\}\right]
 \le\frac{BM}{N}\left\lfloor\frac{N\alpha}{BM}\right\rfloor
 =\alpha-\Delta_{m,M}(\alpha,B),
 \label{eq:tied-rank-bound}
\end{equation}
where
\[
 \Delta_{m,M}(\alpha,B)
 :=\alpha-\frac{BM}{N}\left\lfloor\frac{N\alpha}{BM}\right\rfloor
 \in[0,BM/N).
\]
For \(M=1\) this reduces to the one-jump witness.
\end{prop}

The attainable witness risks are multiples of $BM/N$, including when
$M\nmid N$. The gap may vanish at rank-grid coincidences, so there is no
uniform positive $\Omega(BM/N)$ lower bound without an arithmetic condition.

As a positive counterpart, whenever the fractional part
$\{N\alpha/(BM)\}\ge\eta$ for some $0<\eta<1$, every deterministic
symmetric rule valid at level $\alpha$ over $\mathcal P(m,M,B)$ admits a
witness law $P^\star$ with shortfall
$\alpha-\mathbb E_{P^\star}[L_N\{A(L_1,\ldots,L_m)\}]\ge\eta\,BM/N$:
writing $N\alpha/(BM)=\lfloor N\alpha/(BM)\rfloor+\theta$, the witness
shortfall of \eqref{eq:tied-rank-bound} is at least
$BM\theta/N\ge\eta BM/N$.

\paragraph{Comparison with Section~\ref{sec:mondrian-crc}.}
The sandwich gives a restricted-risk shortfall at most $(M+1)B/N$ under its
additional endpoint assumptions, while the witness attains a gap in
$[0,BM/N)$. These are respectively a uniform upper bound and an
arithmetic lower-bound witness, not matching realized risks.
\subsection{Proofs of rank-granularity and local-support results}
\label{app:lower-bounds-proofs}

Write \(m=n_k\), \(N=m+1\), and \(q=\alpha/B\in(0,1)\).
The loss-function  classes and their zero-endpoint and jump conventions are
specified in Section~\ref{sec:lower-bounds}. Symmetry is assumed only
when it appears in the corresponding proposition.

\subsubsection{Proof of Proposition~\ref{thm:minimax-lower-bound}}
\label{app:proof-minimax}

\begin{proof}
Let \(\kappa=\lfloor Nq\rfloor\), choose the distinct thresholds
\(t_1<\cdots<t_N\), and let \(\pi\) be uniform over permutations.
Put \(T_i=t_{\pi(i)}\) and
\(L_i(\mu)=B\mathbf1\{\mu<T_i\}\).
Every function is bounded, monotone and right-continuous, and all are zero
at \(\lambda_{\max}\). No two jump locations coincide, so
\(P_t\in\mathcal P(m,1,B)\subseteq\mathcal P(m,M,B)\).

For each \(j\in\{1,\ldots,N\}\), set
\[
 \widehat\lambda^{(-j)}
 :=A(L_1,\ldots,L_{j-1},L_{j+1},\ldots,L_N).
\]
Since \(A\) is symmetric, holding out the curve with threshold
\(t_r\) produces the same deterministic action \(a_r\) for every
permutation \(\pi\), regardless of the order of the remaining
curves. Consequently
\[
 S:=\sum_{j=1}^N
  \mathbf1\{\widehat\lambda^{(-j)}<T_j\}
  =\sum_{r=1}^N\mathbf1\{a_r<t_r\}
\]
is a deterministic integer in \(\{0,\ldots,N\}\).
Exchangeability and symmetry yield
\[
 \begin{aligned}
 \mathbb E_{P_t}\left[L_N\{A(L_1,\ldots,L_m)\}\right]
 &=\frac1N\sum_{j=1}^N
   \mathbb E_{P_t}\left[L_j(\widehat\lambda^{(-j)})\right]\\
 &=\frac{B}{N}S.
 \end{aligned}
\]
Even validity at the single law \(P_t\) forces
\(BS/N\le\alpha\), hence
\(S\le\lfloor N\alpha/B\rfloor=\kappa\).
The claimed upper bound follows. This quantization is specific to the witness law.
\end{proof}

\subsubsection{Proof of Lemma~\ref{lem:witness-realizability}}
\label{app:proof-realizability}

\begin{proof}
The abstract witnesses are defined on the loss-curve class for any
$\Lambda$; here we construct a concrete prediction-set problem realizing
them, which uses finite endpoints $\lambda_{\min},\lambda_{\max}\in\mathbb R$
so that the score below is real-valued. Choose a finite threshold-support
set \(\mathcal T\) containing all the jump locations used by the witness,
together with \(\lambda_{\min}\). Let
\(\mathcal X=\mathcal T\), \(\mathcal Y=\{0,1\}\), and use the fixed
identifier \(g\equiv1\), the fixed score
\(s(x,0)=x\), \(s(x,1)=\lambda_{\max}\), and the bounded
monotone loss
\(\ell(C,y)=B\mathbf1\{y\notin C\}\).
Give each observation true label \(Y_i=0\), and let its feature
\(X_i\) equal its assigned jump location (\(\lambda_{\min}\)
for an identically zero path). Its restricted loss satisfies
\[
 \ell(\mathcal C^0_{1,\mu}(X_i),Y_i)
 =B\mathbf1\{X_i>\mu\}.
\]
The distinct-jump witness is obtained by uniformly permuting
\(t_1,\ldots,t_N\); the tied-jump witness by uniformly assigning the
specified blocks and inactive slots. The support-barrier witnesses
arise by including either no spike or one spike, uniformly allocated
across indices. All resulting observation arrays are exchangeable,
because their joint feature arrays are invariant under permutations.

For the ordinal realization, let
\(\mathcal Y=\{0,1,\ldots,B\}\) for integer \(B\ge1\), keep
\(Y_i=0\), take \(s(x,0)=x\) and \(s(x,y)=\lambda_{\max}\) for
\(y\ne0\), and use
\(\ell(C,y)=\min_{c\in C}|c-y|\) with
\(\ell(\varnothing,y)=B\). Below the true-label threshold the restricted set
is empty and has loss \(B\); above it, the restricted set includes label zero
and has loss zero. This again realizes the designated paths.
\end{proof}

\subsubsection{Proof of Proposition~\ref{thm:minimax-lower-bound-general-M}}
\label{app:proof-minimax-general-M}

\begin{proof}
Let \(G=\lfloor N/M\rfloor\) and \(r=N-GM\). The law
\(P_\tau^{(M)}\) uniformly assigns the \(N\) indices to \(G\)
labelled active blocks of size \(M\), plus \(r\) inactive slots.
Each function is monotone, right-continuous and zero at the maximal
endpoint. At each \(\tau_g\), exactly \(M\) paths jump and no other
function jumps, so the law lies in \(\mathcal P(m,M,B)\).

For an index \(j\), write
\(\widehat\lambda^{(-j)}=A(L_1,\ldots,L_{j-1},L_{j+1},\ldots,L_N)\).
When an index from block \(g\) is omitted, the remaining
\emph{multiset} of paths is the same for every index in that block.
By deterministic symmetry, the returned threshold is a fixed value
\(a_g\) depending on \(g\), independently of the random allocation.
Omitting an inactive function produces some fixed action \(a_0\)
(if \(r>0\)) but contributes zero test loss. Therefore
\[
 \sum_{j=1}^N L_j(\widehat\lambda^{(-j)})
 =BM\sum_{g=1}^G\mathbf1\{a_g<\tau_g\}
 =BM S',
\]
where \(S'\in\{0,\ldots,G\}\) is a deterministic integer.
By exchangeability and symmetry,
\[
 \mathbb E_{P_\tau^{(M)}}
 \left[L_N\{A(L_1,\ldots,L_m)\}\right]
 =\frac{BM}{N}S'.
\]
Validity at \(P_\tau^{(M)}\) gives
\(S'\le\lfloor N\alpha/(BM)\rfloor\). This proves
\eqref{eq:tied-rank-bound}. Inactive slots do not affect the rank increments.
\end{proof}

\subsubsection{Proof of Proposition~\ref{thm:small-group-nontriviality}}
\label{app:proof-small-cell}

\begin{proof}
\emph{(i)} Let \(J\) be uniform on \(\{1,\ldots,N\}\).
Place the single nonzero path
\(L_J(\mu)=B\mathbf1\{\mu<\lambda_{\max}\}\)
in position \(J\), leaving all other paths identically zero.
The resulting law is exchangeable, belongs to
\(\mathcal P(m,1,B)\), and has test loss zero unless
\(J=N\). When \(J=N\), the calibration input is
\((0,\ldots,0)\), so
\[
 \mathbb E[L_N\{A(L_1,\ldots,L_m)\}]
 =\frac{B}{N}
 \mathbf1\{A(0,\ldots,0)<\lambda_{\max}\}.
\]
Since \(B/N>\alpha\), validity implies
\(A(0,\ldots,0)=\lambda_{\max}\). This also covers
\(m=0\).

\emph{(ii)} Fix any \(\varepsilon\in(0,1]\) and take a mixture which
with probability \(1-\varepsilon\) has all paths zero, and with
probability \(\varepsilon\) has exactly one path
\(B\mathbf1\{\mu<\lambda_{\max}\}\), placed uniformly
among the \(N\) indices. This is an exchangeable law in
\(\mathcal P(m,1,B)\); the probability of a nonzero augmented path
is exactly \(\varepsilon\).
The event that the calibration paths are all zero has probability
\[
 (1-\varepsilon)+\varepsilon\frac1N
 =1-\frac{m}{N}\varepsilon.
\]
On this event, part~(i) forces the action \(\lambda_{\max}\).
This establishes \eqref{eq:barrier-mixture}. Since
\(N\alpha/B<1\), substituting
\(\varepsilon=N\alpha/B\) gives
\eqref{eq:barrier-original-probability}. Taking
\(\varepsilon=1\) proves the final assertion with a nonzero
augmented function almost surely.
\end{proof}

\subsubsection{Proof and grid counterexample for Proposition~\ref{prop:symmetric-full-support-barrier}}
\label{app:proof-symmetric-full-barrier}

\begin{proof}
The case \(m=0\) follows from
Proposition~\ref{thm:small-group-nontriviality}(i).
For \(m\ge1\), fix any admissible tuple of loss paths
\((\ell_1,\ldots,\ell_m)\) with at most \(M\) coincident
downward jumps at any threshold.
Suppose a deterministic symmetric valid rule selects
\(a=A(\ell_1,\ldots,\ell_m)<\lambda_{\max}\).
Each bounded non-increasing real function on an interval has at most
countably many jump locations. Consequently, there exists a threshold
\(t\in(a,\lambda_{\max})\) at which none of the \(m\) curves jumps.
Adjoin the function \(h_t(\mu)=B\mathbf1\{\mu<t\}\), which has a
single downward jump at \(t\). The \(N\)-tuple consisting of the
\(m\) fixed curves and \(h_t\), after a uniform random permutation,
is exchangeable and lies in \(\mathcal P(m,M,B)\): by construction
its paths are right-continuous, zero at \(\lambda_{\max}\), and
the newly appended function has a jump at a location not shared with
any original path, so no threshold has more than \(M\) jumps. When the appended function is in the test
position, an event of probability \(1/N\), the \(m\) calibration
paths form a permutation of the original tuple, so symmetry forces
the action \(a<t\). The test loss on that event is exactly \(B\).
All losses are nonnegative; therefore
\[
 \mathbb E\bigl[L_N\{A(L_1,\ldots,L_m)\}\bigr]
 \ge\frac{B}{N}>\alpha,
\]
contradicting distribution-free validity. Hence \(A\) always selects
\(\lambda_{\max}\) on admissible inputs.
\end{proof}

\paragraph{Why a finite-grid extension is false.}
Let \(m=1\), \(\Lambda=\{0,1\}\), and \(0<\alpha<B/2\).
Every admissible function in \(\mathcal P(1,1,B)\) has
\(L(1)=0\), and at most one of the two augmented paths can
have positive loss at \(0\); if both did, both would jump at
\(1\), violating the \(M=1\) restriction. Define
\[
 A(L_1)=
 \begin{cases}
 0,&L_1(0)>0,\\
 1,&L_1(0)=0.
 \end{cases}
\]
If the test function has positive loss at \(0\), the calibration path
must be identically zero and the rule returns \(1\), where the test
loss is zero. If the test function is zero, its loss is zero regardless
of the returned threshold. Thus this deterministic, symmetric local
rule has \emph{zero} test risk under every admissible exchangeable
law, despite returning the informative threshold \(0\) on some
admissible nonzero calibration inputs. A fresh jump location above
that threshold is unavailable on this grid.

\section{Overlapping-group reconciliation and alignment}
\label{app:two-group}

The results in Section~\ref{sec:split-multigroup-estimator} concern arbitrary
bounded monotone losses. For inclusion-type losses, the reconciliation
shortfall has a simple exact form. We also give a two-group construction
in which the population restricted cutoffs can be aligned without refining
calibration into intersection cells.

\subsection{Inclusion-type reconciliation}
\label{app:overlap-decomposition}

Let $\ell(C,y)=c(y)\mathbf1\{y\notin C\}$, $0\le c(y)\le B$,
and write $U=s(X,Y)-\widehat b(X,Y)$. On $\{Z\in G_j\}$, the true
label is eligible for restricted set $j$, so
\begin{equation}\label{eq:reconciliation-identity}
 R_j^{\mathrm{slice}}-R_j^{\mathrm{dep}}
 =\mathbb E\!\left[c(Y)\mathbf1\{
 \widehat t_j<U\le T_{\max}(X,Y)\}
 \middle|Z\in G_j,\mathcal D_{\mathrm{fit}}\right].
\end{equation}
Thus the additional risk reduction occurs only where another applicable
cutoff exceeds $\widehat t_j$. If these cutoffs agree, the inclusion-type
reconciliation gap is zero. For a general structured loss, candidate labels
imported from another group's restricted set may lower loss even at equal cutoffs; the
example in Section~\ref{sec:split-multigroup-estimator} demonstrates this
separate membership effect.

\subsection{Feasibility of a common cutoff}
\label{app:overlap-feasibility}

For a membership pattern $S\subseteq[m]$, set
$S(Z)=\{j:Z\in G_j\}$, $\pi_S=\mathbb P\{S(Z)=S\}$,
and $p_j=\sum_{S\ni j}\pi_S$. Restrict attention to patterns
with positive probability and take a patternwise constant shape $b_S$.
Define the weighted conditional score tail
$\rho_S(u)=\mathbb E[c(Y)\mathbf1\{s(X,Y)>u\}\mid S(Z)=S]$.
At a common threshold $K$, each group-restricted set has risk
\begin{equation}\label{eq:feasibility-system}
 p_j r_j(K;b)=\sum_{S\ni j}\pi_S\rho_S(K+b_S),
 \qquad j=1,\ldots,m.
\end{equation}
Consequently, simultaneous tightness at prescribed levels $\alpha_j$
is feasible within this shape class precisely when there are attainable
region cutoffs $u_S$ satisfying
$\sum_{S\ni j}\pi_S\rho_S(u_S)=p_j\alpha_j$ for every $j$;
then take $b_S=u_S-K$. This is a population feasibility condition,
not a distribution-free existence or estimation guarantee.

For two groups with three positive-probability regions $10$, $01$ and $11$,
the system has two equations and three region cutoffs. A useful
construction fixes any attainable overlap tail level $v=\rho_{11}(u_{11})$
and then solves
\[
\rho_{10}(u_{10})=\frac{p_1\alpha_1-\pi_{11}v}{\pi_{10}},
\qquad
\rho_{01}(u_{01})=\frac{p_2\alpha_2-\pi_{11}v}{\pi_{01}},
\]
provided the right-hand sides belong to the respective attainable ranges.
The three regionwise shifts are representable by
$b_\beta(x,y)=\beta_0+\beta_1\mathbf1\{(x,y)\in G_1\}
+\beta_2\mathbf1\{(x,y)\in G_2\}$, with
$\beta_0=b_{10}+b_{01}-b_{11}$,
$\beta_1=b_{11}-b_{01}$ and
$\beta_2=b_{11}-b_{10}$. For larger overlapping families an additive
shape need not represent every patternwise solution.

\subsection{Bounded-score illustration}
\label{app:overlap-illustration}

Let $\mathcal Y=\{10,01,11\}$, where each label is a two-bit string
$y_1y_2\in\{0,1\}^2$ (not a decimal number), and
$G_1=\{Y:y_1=1\}=\{10,11\}$, $G_2=\{Y:y_2=1\}=\{01,11\}$.
Give these three labels respective probabilities $(0.4,0.4,0.2)$.
Conditional on $Y=r$, let $X$ be uniform on $[\mu_r,\mu_r+1]$,
where $(\mu_{10},\mu_{01},\mu_{11})=(0,1,1/2)$, and take
$s(x,y)=x$, $c\equiv1$, and $\alpha_1=\alpha_2=0.2$.
The unshaped population cutoffs are
\[
 \lambda_1=\frac{29}{30},\qquad \lambda_2=\frac{17}{10}.
\]
Consequently, the union's group-$1$ risk is
$(2/3)(1-29/30)=1/45$, rather than $0.2$:
its looser second cutoff removes the overlap-region contribution entirely. In contrast, the region cutoffs
\[
(u_{10},u_{01},u_{11})=(0.8,1.8,1.3)
\]
yield weighted tail risk $0.2$ within every membership region.
With $K=0$ and $b_r^\star=u_r$, both shaped population restricted cutoffs
are $K$, and the deployed union also has both group risks equal to $0.2$.
These scores and shapes are bounded; choosing $\Lambda=[-2,3]$
ensures that the maximal restricted set includes all eligible candidates.
This illustrates an achievable population alignment, not an efficiency
optimality result or a finite-sample claim about estimating $b^\star$.
\subsection{Proof of Corollary~\ref{prop:pinball-alignment}}
\label{app:pinball-alignment}
\begin{proof}
\emph{Population moment equations.}
Because $\Phi$ and $c$ are bounded, $Q$ is finite near its
unconstrained minimizer, and the shaped true-label residual has no atom
at zero conditional on each $G_j$, the hinge is differentiable in every
coefficient direction at $\beta^\star$. Indeed, the only potential
nondifferentiability in coordinate $j$ occurs on
$\{Z\in G_j,\,s(X,Y)=q_{\beta^\star}(X,Y)\}$, an event of
probability zero. Dominated differentiation therefore gives
\[
 \frac{\partial Q(\beta^\star)}{\partial\beta_j}
 =\alpha_jp_j-
 \mathbb E\!\left[c(Y)\Phi_j(X,Y)
 \mathbf1\{s(X,Y)>q_{\beta^\star}(X,Y)\}\right].
\]
Optimality implies that these derivatives vanish. Dividing by
$p_j$ gives
\[
 r_j(0;b^\star)=
 \mathbb E\!\left[c(Y)\mathbf1\{s(X,Y)-b^\star(X,Y)>0\}
 \middle|Z\in G_j\right]=\alpha_j.
\]
In particular, the limiting \emph{true-label} shaped score
$s(X,Y)-b^\star(X,Y)$ has no atom at zero conditional on $G_j$.
This does not imply the all-candidate boundary-stability
condition~\eqref{eq:boundary-stability}.

\emph{Finite-sample validity.}
Conditional on $\mathcal D_{\mathrm{fit}}$, the fitted shape
$\widehat b$ is fixed. The groupwise restricted losses are
non-increasing and right-continuous in their thresholds;
under the maximal-endpoint condition, Theorem~\ref{thm:joint-fit-crc}
gives the assertion in part~(a).

\emph{Convergence of the corrected thresholds.}
Hypothesis~(i) gives the required shape convergence directly:
\[
 \lVert\widehat b-b^\star\rVert_\infty
 =\lVert q_{\widehat\beta}-q_{\beta^\star}\rVert_\infty
 \xrightarrow{\mathbb P}0.
\]
No uniqueness of $\beta^\star$ is needed: all arguments depend
only on the limiting threshold shape.
Let $\widehat R_j(t;\widehat b)$ denote the empirical
mean restricted loss in the correction fold. Its actual
CRC feasibility criterion is
\[
 \frac{n_j}{n_j+1}\widehat R_j(t;\widehat b)
 +\frac{B_j}{n_j+1}\le\alpha_j.
\]
For each sufficiently small $\varepsilon>0$,
hypotheses~(ii)--(iii) and $n_j\to\infty$ imply that, with
probability tending to one, the criterion is false at
$t=-\varepsilon$ and true at $t=\varepsilon$.
Monotonicity, the smallest-feasible-threshold convention,
and $0\in\operatorname{int}\Lambda$ then yield
$-\varepsilon<\widehat t_j\le\varepsilon$ with probability
approaching one. Hence $\widehat t_j\to0$ in probability.

\emph{Deployed risk.}
On $Z\in G_j$, the true label belongs to at least one
calibrated group, so its deployed admission event is
\[
 \left\{s(X,Y)-\widehat b(X,Y)
 \le \max_{k:Z\in G_k}\widehat t_k\right\}.
\]
The finite number of group thresholds and the convergence
above imply that the right-hand cutoff converges in
probability to zero. The true-label atomlessness assumed above therefore gives convergence in probability of the
miscoverage indicator to
$\mathbf1\{s(X,Y)-b^\star(X,Y)>0\}$ on $G_j$.
Multiplication by bounded $c(Y)$ yields $L^1$ convergence;
conditional-expectation contraction, averaging over the
correction fold conditional on the fit, gives
$R_j^{\mathrm{dep}}\xrightarrow{\mathbb P}r_j(0;b^\star)=\alpha_j$.
This argument avoids the stronger all-candidate boundary
assumption of Theorem~\ref{thm:residual-gap}.
\end{proof}

\subsection{Proof and scope of Theorem~\ref{thm:product-reconciliation-obstruction}}
\label{app:proof-product-obstruction}

\begin{proof}
Throughout, write $A_v=\mathbf1\{U_v\le K+b(v)\}$, so the
$A_v$ are independent Bernoulli variables with means $a_v$.
As $Y$ is uniform and independent of $(A_v)$, all risks below follow
by conditioning on $Y$ and enumerating candidate-admission patterns.

\emph{Part (a).} Put $(a,b,c)=(a_0,a_1,a_2)$ and $B=2^r$.
For the two group-restricted sets and their deployed union,
\begin{align*}
 R_1^{\mathrm{slice}}
 &=B\left(1-\frac{a+c}{2}\right),\\
 R_2^{\mathrm{slice}}
 &=B-\frac{2B-1}{2}(b+c)+(B-1)bc,\\
 \Delta_1^\star&=\frac{B-1}{2}b(2-a-c),\\
 \Delta_2^\star&=\frac{B-1}{2}a(1-b)(1-c).
\end{align*}
Tightness at $(B/2,(B+1)/4)$ forces $c=1-a$ and
\[
 b=\frac{2(2B-1)a-(B-1)}{2+4(B-1)a},
 \qquad a_0:=\frac{B-1}{2(2B-1)}\le a\le1.
\]
The resulting gaps are
\[
 \Delta_1^\star=\frac{B-1}{2}b,
 \qquad
 \Delta_2^\star=\frac{B-1}{2}\,
   \frac{a^2(B+1-2a)}{2+4(B-1)a}.
\]
The derivative of $\Delta_2^\star$ has the sign of
$P_B(a)=B+1+(B^2-4)a-4(B-1)a^2$.
For $B\ge2$, this concave quadratic is positive at $a=0$
and has exactly one positive root, so $\Delta_2^\star$
is unimodal on $a\ge0$; its minimum on $[a_0,1]$
is at an endpoint. At these endpoints,
\[
 \Delta_2^\star(a_0)=\frac{(B-1)^3}{8(2B-1)^2},
 \qquad
 \Delta_2^\star(1)=\frac{(B-1)^2}{4(2B-1)}
 >\Delta_2^\star(a_0).
\]
Thus $\max_j\Delta_j^\star\ge\Delta_2^\star(a_0)$.
At $a=a_0$, $b=0$ and $c=1-a_0$, giving
$\Delta_1^\star=0$ and equality in the claimed bound.
The cases $r=1,2,3$ give $1/72$, $27/392$, and $343/1800$.

\emph{Part (b).} Write $(a_0,a_1,a_2,h)$ for the four
candidate-admission probabilities, with $h=a_3$. Direct
conditioning on $Y$ gives
\[
 R_{j+1}^{\mathrm{slice}}
 =3-\frac{3+j}{2}(a_j+h)+j a_jh,
 \qquad j=0,1,2.
\]
The restricted-risk tightness equations yield
\[
 a_0=1-h,\qquad
 a_1=\frac{7/4-2h}{2-h},\qquad
 a_2=\frac{2-(5/2)h}{(5/2)-2h},
 \qquad 0\le h\le\frac45.
\]
For the first group,
\begin{align*}
 \Delta_1^\star
 &=\frac12\left[h\{2a_1+(1-a_1)a_2\}
 +(1-h)\{2a_2+(1-a_2)a_1\}\right]\\
 &=\frac{12h^3+20h^2-110h+71}{8(h-2)(4h-5)}.
\end{align*}
Its derivative has the sign of
$P(h)=16h^4-104h^3+180h^2-56h-59$.
To verify $P(h)<0$ throughout $[0,4/5]$, write
\[
 P(h)=h^3(16h-104)+(180h^2-56h-59).
\]
For $h\le1/2$ the first term is nonpositive, since $16h-104<0$; the second is a
convex quadratic with vertex at $h=56/360\approx0.16$, so on $[0,1/2]$ it is
largest at an endpoint, with value $-42<0$ at $h=1/2$. For $1/2\le h\le4/5$,
using $16h-104<0$ and $h^3\ge\tfrac18$ gives
$h^3(16h-104)\le\tfrac18(16h-104)\le\tfrac18\bigl(-\tfrac{456}{5}\bigr)=-\tfrac{57}{5}$,
the last step by $16h-104\le16\cdot\tfrac45-104=-\tfrac{456}{5}$; the same convex
quadratic is increasing on this interval, its vertex lying to the left, so it is
at most its value $\tfrac{57}{5}$ at $h=4/5$. The two bounds are attained at
different points, so their sum is strictly negative. Consequently,
$\Delta_1^\star$ is strictly decreasing, and
\[
 \max_j\Delta_j^\star\ge\Delta_1^\star
 \ge\Delta_1^\star(4/5)=\frac9{80}.
\]
At $h=4/5$ the admission vector is
$(a_0,a_1,a_2,a_3)=(1/5,1/8,0,4/5)$;
its restricted risks meet all three targets and its gaps are
$(9/80,21/200,11/200)$. Equality follows.

In both designs, a candidate-wise cutoff vector $a_v\in[0,1]$
can be realized with $K=0$ and $b(v)=a_v$ because
$U_v\sim\mathrm{Unif}[0,1]$. It also lies in the span of
the intercept and group-indicator functions: the $m+1$
membership vectors in the two displayed examples give a
full-rank $(m+1)\times(m+1)$ design matrix.
\end{proof}

\begin{proof}[Proof of Corollary~\ref{cor:small-alpha-obstruction}]
From the part-(a) identities,
$R_1^{\mathrm{slice}}=B\bigl(1-\tfrac{a_0+a_2}{2}\bigr)$,
$R_2^{\mathrm{slice}}=B-\tfrac{2B-1}{2}(a_1+a_2)+(B-1)a_1a_2$, and
$\Delta_1^\star=\tfrac{B-1}{2}a_1(2-a_0-a_2)$. Tightness of the first restricted set gives
$2-a_0-a_2=2\alpha/B$, so $\Delta_1^\star=\tfrac{B-1}{B}\alpha a_1$. Since
$\partial R_2^{\mathrm{slice}}/\partial a_2=(B-1)a_1-\tfrac{2B-1}{2}\le-\tfrac12<0$
for every $a_1\in[0,1]$, the map $a_2\mapsto R_2^{\mathrm{slice}}$ is strictly
decreasing; with $a_2\le1$,
$\alpha=R_2^{\mathrm{slice}}\ge R_2^{\mathrm{slice}}\big|_{a_2=1}=\tfrac{1-a_1}{2}$,
hence $a_1\ge1-2\alpha$. Therefore
$\max_j\Delta_j^\star\ge\Delta_1^\star=\tfrac{B-1}{B}\alpha a_1
\ge\tfrac{B-1}{B}\alpha(1-2\alpha)$. At $(1-2\alpha/B,\,1-2\alpha,\,1)$ both restricted
risks equal $\alpha$, $\Delta_2^\star=\tfrac{B-1}{2}a_0(1-a_1)(1-a_2)=0$, and
$\Delta_1^\star=\tfrac{B-1}{B}\alpha(1-2\alpha)$, so the bound is attained.
For the fixed candidate-wise shape $b(v)=a_v$ and $K=0$, the population
restricted risk has a strict local crossing at $0$ in both groups: candidate $0$
has interior admission probability $1-2\alpha/B$ and is the true label
with positive conditional probability in $G_1$, while candidate $1$ has
interior admission probability $1-2\alpha$ and is the true label with
positive conditional probability in $G_2$. The shared candidate $2$ has
admission probability one at the reference cutoff, but its saturation
does not remove these crossings. Under iid sampling and an independent fixed
shape, the corrected restricted thresholds therefore converge to zero, and the
small-target residual gap is also realized as an asymptotic CRC limit.
\end{proof}

\paragraph{The attaining shapes as CRC limits.}
Fix the attaining cutoff vector of either part and the corresponding shape
$b(v)=a_v$ with $K=0$ on the threshold domain $\Lambda=[-2,3]$, under the
independent uniform-score model of the theorem and i.i.d.\ sampling. Since
$U_v\sim\mathrm{Unif}[0,1]$ and $a_v\in[0,1]$, the interior point $0$ admits
each candidate $v$ with probability $a_v$, and $\lambda_{\max}=3$ admits every
candidate, so the corrected thresholds range over a real interval with an
interior crossing. Each group $G_j$ contains a candidate with strictly interior
admission probability: in part~(a) the shared candidate $2$ has admission
$c=1-a_0\in(0,1)$ and lies in both $G_1$ and $G_2$, and in part~(b) the shared
candidate $3$ has admission $4/5$ in all three groups. Writing
$R_j^{\mathrm{slice}}(t)$ for the population group-$j$ restricted risk under the
shifted threshold $K+t$, the shared candidate has an interior cutoff and
positive conditional probability of being the true label in every group;
admitting it with strictly higher probability strictly reduces the structured
loss. Hence $t\mapsto R_j^{\mathrm{slice}}(t)$ is continuous and strictly
decreasing at $t=0$, where it equals $\alpha_j$ by construction, establishing a
strict local risk crossing irrespective of candidates whose admission
probabilities equal zero at the reference cutoff. With $b$ fixed, so
$\widehat b=b^\star=b$, the corrected split-CRC threshold $\widehat t_j$ of
\eqref{eq:overlap-pw-threshold} then satisfies
$\widehat t_j\xrightarrow{\mathbb P}0$ by the level-crossing argument of
Appendix~\ref{app:pinball-alignment}, using uniform convergence of the
empirical restricted risk and $n_j\to\infty$. Continuous scores give the
boundary-stability condition \eqref{eq:boundary-stability}, so
Theorem~\ref{thm:residual-gap} applies and
$R_j^{\mathrm{slice}}\to\alpha_j$ while
$R_j^{\mathrm{dep}}\xrightarrow{\mathbb P}\alpha_j-\Delta_j^\star$. Since $m$ is
fixed, the continuous mapping theorem gives
\[
 \max_{j=1,\ldots,m}\bigl(\alpha_j-R_j^{\mathrm{dep}}\bigr)
 \ \xrightarrow{\mathbb P}\
 \max_{j=1,\ldots,m}\Delta_j^\star\ =\ c_\star\ >\ 0 ,
\]
where $c_\star$ is the sharp constant of
Theorem~\ref{thm:product-reconciliation-obstruction}; equivalently, for every
$0<\varepsilon<c_\star$,
$\mathbb P\{\max_j(\alpha_j-R_j^{\mathrm{dep}})\ge c_\star-\varepsilon\}\to1$.
The lower bound is therefore realized as the asymptotic limit of the separately
corrected CRC procedure applied with these fixed shapes, not merely as a
population optimization. This is a statement for the stated models and shapes,
not a universal lower bound over arbitrarily fitted procedures.

\paragraph{Why coupled admissions can eliminate the gap.}
Return to part~(a) with $r=1$, and keep $Y$ uniform.
Consider instead a coupled random set, independent of $Y$,
with
\[
 \mathbb P(C=\varnothing)=\tfrac14,\qquad
 \mathbb P(C=\{2\})=\tfrac12,\qquad
 \mathbb P(C=\{0,1,2\})=\tfrac14.
\]
Define the group-$j$ restricted set as $C\cap\{v:(x,v)\in G_j\}$.
Direct enumeration gives
$R_1^{\mathrm{slice}}=R_1^{\mathrm{dep}}=1$ and
$R_2^{\mathrm{slice}}=R_2^{\mathrm{dep}}=3/4$.
The three candidate admissions are dependent, so this
construction does not belong to the product-form class.
It can be represented by an appropriately feature-dependent
threshold shape, but need not satisfy the unique-crossing
hypotheses used for consistent CRC correction.

\paragraph{Why label prevalences matter.}
Keep the part~(a) groups, $r=1$, and targets $(1,3/4)$,
but change the true-label probabilities to $(1/4,1/4,1/2)$.
For admission probabilities $(a_0,a_1,a_2)=(0,0,3/4)$,
\[
 R_1^{\mathrm{slice}}=R_1^{\mathrm{dep}}=1,
 \qquad R_2^{\mathrm{slice}}=R_2^{\mathrm{dep}}=3/4.
\]
Hence the product-form lower bound is not uniform over
label distributions, even with the same loss and group memberships.

\section{Deployment-aware CRC: efficiency and finite-library guarantees}
\label{app:joint-direct-proofs}

\subsection{Proof of Proposition~\ref{prop:joint-direct-size-separation}}
\label{app:joint-direct-size-proof}

\begin{proof}
Let $Y$ be uniform on $\{0,1,2\}$ and independent of the uniform score
coordinates. A union of separately calibrated restricted sets can use distinct candidate-wise
cutoffs for the two groups. Write $a$ for the admission probability of
candidate $0$ in the $G_1$ restricted set, $b$ for that of candidate $1$ in the
$G_2$ restricted set, and $c_1,c_2$ for the probabilities of admitting candidate
$2$ in the $G_1$ and $G_2$ restricted sets, respectively. Because both restricted sets
threshold the \emph{same} uniform score $U_2$, the union admits candidate
$2$ with probability $c=\max(c_1,c_2)$; the expected union size is
$a+b+c$.

From the part-(a) risk identities in the proof of
Theorem~\ref{thm:product-reconciliation-obstruction},
\[
 R_1^{\mathrm{slice}}=B\left(1-\frac{a+c_1}{2}\right),
 \qquad
 R_2^{\mathrm{slice}}
 =B-\frac{2B-1}{2}(b+c_2)+(B-1)bc_2.
\]
The first constraint $R_1^{\mathrm{slice}}\le\alpha$ forces
$a+c_1\ge2-2\alpha/B$. The second restricted risk is decreasing in $c_2$,
and its minimum for fixed $b$ occurs at $c_2=1$, where it equals
$(1-b)/2$. Hence $R_2^{\mathrm{slice}}\le\alpha$ implies
$b\ge1-2\alpha$. Since $c\ge c_1$,
\[
 a+b+c\ge\left(2-\frac{2\alpha}{B}\right)+(1-2\alpha)
 =3-2\alpha-\frac{2\alpha}{B}.
\]
This value is attained at $(a,b,c_1,c_2)=
(1-2\alpha/B,1-2\alpha,1,1)$: both restricted risks equal $\alpha$.
This proves~\eqref{eq:joint-direct-slice-opt} even allowing different
thresholds on the shared candidate in the two restricted sets.

For the direct construction, put
$A=1-2\alpha/\{1+2\alpha(B-1)\}$,
$D=1-2\alpha$, and admit the three candidates independently with
probabilities $(A,D,1)$. The deployed set always contains candidate
$2$. Conditional on $Y=0$, the loss is zero if candidate $0$ is
admitted, one if $0$ is excluded but $1$ is admitted, and $B$ if both
$0$ and $1$ are excluded. Conditional on $Y=2$, the loss is zero.
Consequently,
\[
 R_1^{\mathrm{dep}}=
 \frac12(1-A)\{D+B(1-D)\}
 =\frac12\frac{2\alpha}{1+2\alpha(B-1)}
      \{1+2\alpha(B-1)\}=\alpha.
\]
Conditional on $Y=1$, the loss is one precisely when candidate $1$
is excluded, because candidate $2$ is always admitted; conditional on
$Y=2$, it is zero. Thus
$R_2^{\mathrm{dep}}=(1-D)/2=\alpha$.
The expected cardinality is $A+D+1$, giving
\eqref{eq:joint-direct-size}; subtraction and simplification give
\eqref{eq:joint-direct-strict-gain}.

Finally, choose the fixed candidate-only offset $h(v)=a_v$ from
\eqref{eq:joint-direct-attainer}, with $s(X,v)=U_v$, $t=0$ and
$\mathcal U=\{0,1,2\}$. Then $D_0^h$ admits $v$ with probability
$a_v$. The two population direct risk curves are continuous and
strictly decreasing in a neighborhood of $t=0$: for group $1$, the
admission probability of candidate $0$ varies in the interior of
$(0,1)$, and for group $2$ the same holds for candidate $1$.
Both curves cross their target $\alpha$ at zero. Under i.i.d. data,
positive group probabilities and uniform laws of large numbers for
these bounded nested step-loss families, the corrected thresholds of
\eqref{eq:joint-direct-thresholds} converge in probability to zero.
Atomless candidate scores then imply convergence of conditional risks
and expected cardinality to the displayed population values.
\end{proof}

\paragraph{Offset-shift invariance.}
For any constant $c$, $D_t^{h+c}=D_{t+c}^{h}$ and
$\widehat t_j^{\mathrm{dir}}(h+c)=\widehat t_j^{\mathrm{dir}}(h)-c$.
Hence the corrected deployed set is invariant to intercept shifts,
although the nominal fitting objective at $t=0$ is not. One may fix
an offset normalization and vary the nominal cutoff, or keep an intercept
for fitting and deduplicate equivalent corrected shapes.

\subsection{A finite-library size oracle inequality}
\label{app:joint-direct-oracle}

The fitting step can be made quantitatively conservative without
assuming consistency of a nonconvex optimizer. The result below is a
finite-library statement: it does not yield an oracle bound for an
unrestricted neural-network shape class. We assume independent i.i.d.\
observations in the fitting, correction, and test folds for this
performance bound; Theorem~\ref{thm:joint-direct-validity} requires only
the stated conditional exchangeability.

For a fixed offset $h$, abbreviate
\[
 r_j(h)=\mathbb E[\ell(D_0^h(X),Y)\mid Z\in G_j],
 \qquad S(h)=\mathbb E[|D_0^h(X)|].
\]
Fix $\delta\in(0,1)$ and prespecified lower support values
$k_j\ge1$ for the correction counts. Let $H=|\mathcal H|$,
$L=|\mathcal Y|$, $n_{\mathrm{fit}}=|\mathcal D_{\mathrm{fit}}|$, and,
when $n_{\mathrm{fit},j}\ge1$, set
\begin{align*}
 u_j&=B_j\sqrt{\frac{\log(6mH/\delta)}{2n_{\mathrm{fit},j}}},
 &v_j&=B_j\sqrt{\frac{\log(6m/\delta)}{2k_j}},\\
 w&=L\sqrt{\frac{\log(6H/\delta)}{2n_{\mathrm{fit}}}}.
\end{align*}
Use the margins
$\eta_j=u_j+v_j+B_j/k_j$ in
\eqref{eq:joint-direct-fitting} and define the (count-dependent)
strict-feasibility comparison class
\begin{equation}
\label{eq:joint-direct-oracle-class}
 \mathcal H_{\mathrm{strict}}
 =\left\{h\in\mathcal H:
  r_j(h)\le\alpha_j-2u_j-v_j-\frac{B_j}{k_j}
  \ \text{for all }j\right\}.
\end{equation}

\begin{prop}
\label{prop:joint-direct-oracle}
Assume the setting above and that every fitting-group count is
positive. With probability at least $1-\delta$, whenever
$n_j\ge k_j$ for all correction groups and
$\mathcal H_{\mathrm{strict}}\ne\varnothing$, the fitted problem is
feasible, each direct correction threshold obeys
$\widehat t_j^{\mathrm{dir}}\le0$, and
\begin{equation}
\label{eq:joint-direct-oracle-bound}
 \mathbb E\!\left[|\widehat{\mathcal C}_{\mathrm{dir}}(X)|
 \middle|\mathcal D_{\mathrm{fit}},\mathcal D_{\mathrm{cal}}\right]
 \le\min_{h\in\mathcal H_{\mathrm{strict}}} S(h)+2w.
\end{equation}
The probability statement means that the indicated implication holds
outside an event of probability at most $\delta$; it does not assert
that the support or strict-feasibility events occur with high
probability. Exact groupwise validity holds even when these events
fail.
\end{prop}

\begin{proof}
By Hoeffding's inequality and a union bound over $H$ candidates and
$m$ groups, with failure probability at most $\delta/3$,
\[
 |\widehat r_{\mathrm{fit},j}(h)-r_j(h)|\le u_j
 \quad\text{simultaneously for all }h\in\mathcal H,\ j\in[m]
\]
on the event that all fitting-group counts are positive. This also
holds with overlapping groups: for each fixed $j$, conditional on its
membership count, the observations selected by $G_j$ are i.i.d.\ from
the corresponding conditional distribution. With an additional failure
probability at most $\delta/3$, uniformly over $h\in\mathcal H$,
$|\widehat S_{\mathrm{fit}}(h)-S(h)|\le w$.

On these events every $h\in\mathcal H_{\mathrm{strict}}$ satisfies the
fitting constraints, so the fitted problem is feasible, and the selected
$\widehat h$ obeys
\[
 r_j(\widehat h)\le\alpha_j-v_j-B_j/k_j,
 \qquad
 S(\widehat h)\le
 \min_{h\in\mathcal H_{\mathrm{strict}}}S(h)+2w.
\]
Because the correction fold is independent of the fitted $\widehat h$,
Hoeffding's inequality gives, with additional failure probability at
most $\delta/3$, the simultaneous bounds
$\widehat r_{\mathrm{cal},j}(\widehat h;0)
\le r_j(\widehat h)+v_j$ on $\{n_j\ge k_j\}$, where
$\widehat r_{\mathrm{cal},j}$ is the correction-fold groupwise
empirical \emph{direct} risk. It follows that
$\widehat r_{\mathrm{cal},j}(\widehat h;0)
\le\alpha_j-B_j/k_j$. Since $n_j\ge k_j$,
\[
 \frac{n_j\widehat r_{\mathrm{cal},j}(\widehat h;0)+B_j}{n_j+1}
 \le\alpha_j+\frac{B_j-\alpha_j-n_jB_j/k_j}{n_j+1}
 \le\alpha_j.
\]
Thus $t=0$ is feasible for every group, giving
$\widehat t_{\mathrm{dir}}\le0$. Monotonicity yields
$\widehat{\mathcal C}_{\mathrm{dir}}(x)
\subseteq D_0^{\widehat h}(x)$, so its conditional expected
cardinality is at most $S(\widehat h)$, proving the claim.
\end{proof}

The bound compares against \emph{strictly} feasible offsets in a fixed
finite library; the margin accounts for both fitting uncertainty and
correction-fold feasibility. It is not an excess-size bound relative
to the globally optimal jointly valid set. If a rare correction group
defaults to $+\infty$, the global maximum in
\eqref{eq:joint-direct-deployment} makes the prediction set
$\mathcal U$; the positive part of the comparison is therefore most
relevant when all group correction counts are adequate. In particular,
the Hoeffding margins can exceed a small target level, making the
strict-feasibility class empty; they should not be read as a practical
recommendation for all data sizes.

\section{Computational algorithms and implementation conventions}
\label{app:implementation-algorithms}

The algorithms below supplement the main-text calibration procedures. They
specify exact threshold selection, independent shape fitting, comparison
methods, and the three-label fitting variant. All fitting operations precede
and are independent of their stated CRC correction folds.

\subsection{Exact correction and breakpoint search}
\label{app:computation}

For a finite label space the group-\(k\) threshold \eqref{eq:mondrian-threshold} is computed
by sweeping the loss-drop events of the within-group functions. For each calibration point
\(i\in\mathcal I_k\), the function \(\mu\mapsto L_i^{(k)}(\mu)\) is then a non-increasing step
function whose breakpoints are the scores \(s(X_i,y)\) of the group-\(k\) candidate
labels \(y\); sorting those scores identifies the breakpoints. The loss at each
breakpoint is evaluated or updated as labels are admitted.
Pooling the breakpoints of all \(n_k\) functions and sweeping them in increasing order
maintains the running empirical risk \(\widehat R_{n,k}^0(\mu)\), and the threshold is
the first allowed threshold at which the corrected criterion is at most \(\alpha_k\).
For a group with \(c_k\) candidate labels per point, sorting the scores costs
\(O(n_kc_k\log(n_kc_k))\), in addition to evaluating or updating the structured
loss at each breakpoint. For miscoverage-type losses only the true-label score
matters, giving \(O(n_k\log n_k)\) per group and \(O(n\log n)\) in total,
with \(O(K)\) threshold storage. On a discrete threshold grid, score events
are processed at the next allowed threshold. Any candidate-list truncation must
be fixed independently of the calibration fold and preserve the endpoint
condition for Theorem~\ref{thm:mondrian-crc} to apply.

\begin{algorithm}[t]
\caption{\textsc{CRC-Cutoff}: exact event-sweep calibration}
\label{alg:crc-event-sweep}
\begin{algorithmic}[1]
\Require Calibration index set $I$ of size $n$; nested finite-label sets
$F_t(x)=\{v\in E(x):a(x,v)\le t\}$; bound $B_0$, level $0<\alpha<B_0$;
$\Lambda$ either a closed interval or a finite ordered grid, with
maximum $\lambda_{\max}$.
\If{$n=0$}
  \State \Return $\lambda_{\max}$.
\EndIf
\If{$\Lambda$ is an interval}
  \State $\mathcal T\gets
    \bigl(\{\lambda_{\min},\lambda_{\max}\}
    \cup\{a(X_i,v):i\in I,\ v\in E(X_i)\}\bigr)\cap\Lambda$.
\Else
  \State $\mathcal T\gets\Lambda$.
\EndIf
\State Sort the distinct thresholds $\mathcal T$ increasingly.
\For{each $t\in\mathcal T$, in increasing order}
  \State $S(t)\gets\sum_{i\in I}\ell(F_t(X_i),Y_i)$
  \Comment{Admit all candidates tied at $t$}
  \If{$\bigl(S(t)+B_0\bigr)/(n+1)\le\alpha$}
    \State \Return $t$.
  \EndIf
\EndFor
\State \Return $\lambda_{\max}$ \Comment{No feasible corrected threshold}
\end{algorithmic}
\end{algorithm}

\subsection{Shape fitting}

The two routines below select a deployment-aware offset on an independent
fitting fold, prior to any correction-fold calibration.
Algorithm~\ref{alg:direct-library-fit} searches a prespecified finite
library for the minimal-size offset meeting the groupwise fitting
constraints, while Algorithm~\ref{alg:weighted-hinge-fit} fits an additive
group-indicator shape for inclusion-type losses through a convex
weighted-hinge program. In both cases an independent CRC correction is
applied afterward and is what restores exact finite-sample validity.

\begin{algorithm}[t]
\caption{\textsc{Finite-Library-Fit}: deployment-aware offset selection}
\label{alg:direct-library-fit}
\begin{algorithmic}[1]
\Require Independent fitting pairs $\mathcal D_{\mathrm{fit}}$;
prespecified ordered finite library $\mathcal H$; candidate universe
$\mathcal U$, groups $G_j$, score $s$, loss $\ell$, targets $\alpha_j$, margins $\eta_j\ge0$.
\For{$j=1,\ldots,m$}
  \State $I_j^{\mathrm{fit}}\gets
     \{i\in\mathcal D_{\mathrm{fit}}:(X_i,Y_i)\in G_j\}$.
  \If{$|I_j^{\mathrm{fit}}|=0$}
    \State \Return $\mathrm{FAIL}$.
  \EndIf
\EndFor
\For{each $h\in\mathcal H$}
  \State $D_0^h(x)\gets\{v\in\mathcal U(x):s(x,v)\le h(x,v)\}$.
  \State $\widehat S(h)\gets
    |\mathcal D_{\mathrm{fit}}|^{-1}\sum_{i\in\mathcal D_{\mathrm{fit}}}|D_0^h(X_i)|$.
  \For{$j=1,\ldots,m$}
    \State $\widehat r_j(h)\gets
    |I_j^{\mathrm{fit}}|^{-1}\sum_{i\in I_j^{\mathrm{fit}}}\ell(D_0^h(X_i),Y_i)$.
  \EndFor
\EndFor
\State $\mathcal H_{\mathrm{feas}}\gets
 \{h\in\mathcal H:\widehat r_j(h)\le\alpha_j-\eta_j\ \forall j\}$.
\If{$\mathcal H_{\mathrm{feas}}=\varnothing$}
  \State \Return $\mathrm{FAIL}$.
\EndIf
\State \Return the first $h\in\mathcal H_{\mathrm{feas}}$ in the
prespecified order minimizing $\widehat S(h)$.
\end{algorithmic}
\end{algorithm}

\begin{algorithm}[t]
\caption{Weighted-hinge shape fit for inclusion-type losses}
\label{alg:weighted-hinge-fit}
\begin{algorithmic}[1]
\Require Independent fitting pairs $Z_i=(X_i,Y_i)$, $i=1,\ldots,N$;
score $s$; groups $G_j$; levels $\alpha_j$; weights $c(Y_i)\ge0$.
\State $u_i\gets s(X_i,Y_i)$ and
$\Phi_{ij}\gets\mathbf1\{Z_i\in G_j\}$ for all $i,j$.
\State Solve the following linear program (or an equivalent convex hinge problem):
\Statex $\displaystyle\quad\min_{\beta\in\mathbb R^m,\,\xi\in\mathbb R^N}
 \frac1N\sum_{i=1}^N\left(c(Y_i)\xi_i+
             \sum_{j=1}^m\alpha_j\beta_j\Phi_{ij}\right)$
\Statex $\displaystyle\quad\text{subject to}\quad
 \xi_i\ge u_i-\sum_{j=1}^m\Phi_{ij}\beta_j,\qquad
 \xi_i\ge0\quad (i=1,\ldots,N).$
\If{the program is unbounded, has no finite minimizer, or the solver fails}
  \State \Return $\mathrm{FAIL}$ \Comment{Do not deploy unverified coefficients}
\EndIf
\State Select a solution $\widehat\beta$ using a fixed, documented
rule if multiple minimizers exist.
\State \Return $\widehat b(x,v)=\sum_{j=1}^m
 \widehat\beta_j\mathbf1\{(x,v)\in G_j\}$.
\Statex \emph{Fitted union:} use $\widehat b$ in Algorithm~\ref{alg:overlap-union-crc}
with an independent correction fold.
\Statex \emph{Uncorrected joint-quantile comparison:} deploy
$\{v:s(x,v)\le\widehat b(x,v)\}$ without CRC correction.
\end{algorithmic}
\end{algorithm}

If the weighted-hinge fit returns $\mathrm{FAIL}$, use a prespecified,
independently chosen endpoint-safe shape and recalibrate it on the
independent correction fold, or deploy a candidate universe whose
loss satisfies the endpoint bound for every group. Fit failure alone
never licenses an uncorrected prediction set.

\subsection{Alternative calibration and comparison procedures}

The remaining algorithms specify the comparators used in the numerical
studies. Algorithm~\ref{alg:atomwise-crc} is the atom-wise Mondrian
refinement over disjoint membership patterns;
Algorithm~\ref{alg:randomized-classwise} is the randomized
rank-interpolation rule that attains the nominal level in the
common-height one-jump submodel; and
Algorithm~\ref{alg:three-label-gridfit} is the grid-fit variant used in the
three-label deployment-aware experiment. Any comparator that fits a shape
does so on an independent fold and is corrected separately.

\begin{algorithm}[t]
\caption{Atom-wise Mondrian CRC by disjoint group refinement}
\label{alg:atomwise-crc}
\begin{algorithmic}[1]
\Require Fixed original groups $G_1,\ldots,G_m$; score $s$;
calibration pairs; threshold domain $\Lambda$; targets $\alpha_j$;
valid atom loss bounds $B_S$ with $0<\alpha_S<B_S$ and a safe maximal endpoint.
\State For each candidate $v$ define its membership pattern
$M(x,v)\gets\{j:(x,v)\in G_j\}$.
\State Form nonempty disjoint atoms
$A_S\gets\{(x,v):M(x,v)=S\}$ for $\varnothing\ne S\subseteq[m]$.
\For{each atom $A_S$ with nonempty candidate domain}
  \State $\alpha_S\gets\min_{j\in S}\alpha_j$.
  \State $I_S\gets\{i:(X_i,Y_i)\in A_S\}$.
  \State $F_{S,t}(x)\gets\{v:(x,v)\in A_S,\ s(x,v)\le t\}$.
  \State $\widehat\lambda_S\gets
     \Call{CRC-Cutoff}{I_S,F_{S,\cdot},B_S,\alpha_S,\Lambda}$.
\EndFor
\State \Return $\widehat{\mathcal C}_{\mathrm{atom}}(x)
  =\bigcup_{S\ne\varnothing}F_{S,\widehat\lambda_S}(x)$.
\Statex \emph{Patterns $S=\varnothing$ are excluded from the common eligible universe.}
\end{algorithmic}
\end{algorithm}

\begin{algorithm}[t]
\caption{Randomized interpolation between classwise CRC ranks}
\label{alg:randomized-classwise}
\begin{algorithmic}[1]
\Require $m$ calibration scores with distinct values;
common-height one-jump loss $B\mathbf1\{S>t\}$;
level $0<\alpha<B$; threshold endpoints.
\State $N\gets m+1$,
$\kappa\gets\lfloor N\alpha/B\rfloor$,
$\vartheta\gets N\alpha/B-\kappa$.
\State Sort calibration scores in descending order
$S_{(1)}>\cdots>S_{(m)}$.
\State Set $S_{(0)}\gets\lambda_{\max}$ and
$S_{(m+1)}\gets\lambda_{\min}$.
\State Draw $U\sim\mathrm{Unif}[0,1]$ independently of the sample.
\If{$U<\vartheta$}
  \State \Return $S_{(\kappa+1)}$.
\Else
  \State \Return $S_{(\kappa)}$.
\EndIf
\end{algorithmic}
\end{algorithm}

\begin{algorithm}[t]
\caption{Three-label grid-fit variant using expected-size surrogate}
\label{alg:three-label-gridfit}
\begin{algorithmic}[1]
\Require Independent fitting pairs $(U_{i0},U_{i1},U_{i2},Y_i)$;
$Y_i\in\{0,1,2\}$; groups $G_1=\{0,2\}$, $G_2=\{1,2\}$;
ordinal loss $\ell$, target $\alpha=0.10$.
\State $\mathcal H_3\gets\{0,0.1,\ldots,1\}^3$.
\For{$j=1,2$}
  \State $I_j\gets\{i:Y_i\in G_j\}$.
  \If{$|I_j|=0$}
    \State \Return $\mathrm{FAIL}$.
  \EndIf
\EndFor
\For{$h=(h_0,h_1,h_2)\in\mathcal H_3$}
  \State $D_0^h(U_i)\gets\{v\in\{0,1,2\}:U_{iv}\le h_v\}$.
  \For{$j=1,2$}
    \State $\widehat r_j(h)\gets |I_j|^{-1}
       \sum_{i\in I_j}\ell(D_0^h(U_i),Y_i)$.
  \EndFor
\EndFor
\State $\mathcal F\gets\{h\in\mathcal H_3:
    \widehat r_1(h)\le\alpha,\ \widehat r_2(h)\le\alpha\}$.
\If{$\mathcal F=\varnothing$}
  \State \Return $\mathrm{FAIL}$.
\EndIf
\State \Return a prespecified-tie-broken minimizer of
 $h_0+h_1+h_2$ over $\mathcal F$.
\Statex \emph{Then independently correct using Algorithm~\ref{alg:joint-direct-crc}.}
\end{algorithmic}
\end{algorithm}

\FloatBarrier

\section{Synthetic experiments: generation and calibration protocols}
\label{app:reproducibility}

The companion reproducibility package is intended for public release at
\url{https://github.com/TareqAldirawi94/Group-Conditional-Conformal-Risk-Control}.
The original synthetic experiments use \textsf{R} 4.6.0; the direct-method
simulation implementations are in Python, and the image classifier uses
\textsf{PyTorch} and \textsf{torchvision}. The prepared package is
intended to pin Python dependencies in \texttt{requirements.txt} (with
\texttt{requirements-ddr.txt} for the classifier-training pipeline),
list R packages in \texttt{R\_PACKAGES.md}, and provide file hashes in
\texttt{MANIFEST.csv}. A public immutable version tag and verification of
those files are still required before claiming that the materials are
publicly reproducible.

The DDR evaluation protocols appear in Appendix~\ref{app:ddr-protocols};
additional tightness and barrier simulations appear in
Appendix~\ref{app:diagnostic-simulations}. Exact replication requires publicly accessible scripts, inputs,
experiment-specific seeds, and a pinned release identifier.

\subsection{Heterogeneous-group experiment}
\label{app:sim-heterogeneity}

The label space is $\mathcal Y=\{0,1,2,3,4\}$. Independently sample
$G\sim\mathrm{Unif}\{1,2,3\}$, $Y\sim\mathrm{Unif}(\mathcal Y)$, and
$U\sim\mathrm{Unif}(-\sqrt{3},\sqrt{3})$, and let
$\widehat Y=Y+\sigma_G U$ with
$(\sigma_1,\sigma_2,\sigma_3)=(0.7,1.3,2.0)$. The feature vector is
$X=(G,\widehat Y)$, the candidate score is $s(X,j)=|j-\widehat Y|$, and
$\mathcal C_t(X)=\{j:s(X,j)\le t\}$. We use
\[
 \ell(C,y)=\min_{j\in C}|j-y|\quad(C\ne\varnothing),
 \qquad \ell(\varnothing,y)=4.
\]
Each of $500$ independent replications uses $1{,}500$ calibration and
$3{,}000$ test observations, with thresholds searched on $[0,12]$;
the upper endpoint includes every grade. For feature-defined, disjoint
groups, the restricted and deployed losses coincide. Groupwise RCPS uses the
Hoeffding--Bentkus upper bound for $\ell/B$ with $\delta=0.10$ per group;
this is not a simultaneous $\delta$ guarantee. The Mondrian tightness
band shown in Figure~\ref{fig:sim-group-conditional} uses $M_g=1$
and random local sample counts, and is not a confidence interval.

\subsection{Overlapping-label miscoverage experiment}
\label{app:sim-overlap-coverage}

Let $Y\in\{0,1\}^3$ and define
$G_j=\{(x,y):y_j=1\}$, $j=1,2,3$, together with
$G_0=\{(x,y):y=(0,0,0)\}$. Generate
\[
 Z_j=\sqrt\rho\,F+\sqrt{1-\rho}\,E_j,\qquad
 Y_j=\mathbf1\{Z_j>\tau_j\},\qquad
 \eta_j=\gamma_j(Z_j-\tau_j)+\xi_j,
\]
where $F,E_j,\xi_j$ are independent standard normals,
$\tau_j=\Phi^{-1}(1-p_j)$,
$(p_1,p_2,p_3)=(0.55,0.40,0.18)$, and
$(\gamma_1,\gamma_2,\gamma_3)=(1.70,1.00,0.40)$.
For $q_j=\operatorname{logit}^{-1}(\eta_j)$ clipped to
$[10^{-6},1-10^{-6}]$, define
\[
 s(X,v)=-\sum_{j=1}^3\{v_j\log q_j+(1-v_j)\log(1-q_j)\},
 \qquad v\in\{0,1\}^3.
\]
We take miscoverage loss ($B=1$), $\alpha=0.1$, and $\rho\in\{0,0.5\}$.
At $n\in\{100,400,1600,6400\}$ we generate $500$ independent
calibration samples and fresh test samples of size $6{,}000$.

The pointwise union calibrates the four original groups: candidate
$v\ne(0,0,0)$ receives cutoff $\max_{j:v_j=1}\widehat t_j$, while
$(0,0,0)$ receives $\widehat t_0$. The atom-wise comparator calibrates
each of the eight response-vector cells separately. These three methods
(marginal CRC, atom-wise Mondrian, and pointwise union) share the same
$500$ calibration/test replications. For independent labels and $n=100$,
the union and atom-wise set sizes are $4.59$ and $6.30$, respectively;
$47.8\%$ of atom cells and $0.1\%$ of original groups return the maximal
threshold (test-weighted default frequencies $0.174$ and below $0.001$).
At $n=6400$, the union's pooled miscoverage is $0.096$ for independent
labels and $0.102$ for correlated labels. The joint-pinball Kandinsky
comparator is evaluated in the separate fitted-shape experiment below,
not in these $500$ replications or Figure~\ref{fig:sim-overlap-coverage}.

\subsection{Fitted-shape overlapping-group experiment}
\label{app:reproducibility-fitted}

This experiment reuses the overlapping-label model of
Appendix~\ref{app:sim-overlap-coverage} (focal groups $G_j=\{y_j=1\}$, background
$G_0=\{y=(0,0,0)\}$, log-likelihood score, miscoverage loss, $\alpha=0.10$).
Under a fixed total budget $n$ we compare five procedures: the unshaped union
($\widehat b\equiv0$, full budget); the same union restricted to the correction
fold, to isolate the cost of splitting; the fitted-hinge union, whose shape
minimizes the weighted-hinge objective \eqref{eq:pinball-shape-objective} on an
independent fitting fold and is then CRC-corrected per group on the remainder;
Kandinsky's fitted $(1-\alpha)$ quantile over the four group indicators, deployed
without correction; and the eight-cell atom-wise refinement (full budget). The
joint hinge and Kandinsky objectives coincide up to an
additive constant, so they share a fitted shape; their deployed procedures
differ by the independently calibrated CRC correction. The correction-only
unshaped-union control isolates sample splitting and is not displayed in
the four-method fitted-shape figure. This study has its own random draws
and replication counts and must not be pooled with the $500$-replication
three-method experiment above. We sweep
$n\in\{100,400,1600,6400\}$, $\rho\in\{0,0.5\}$, and fit/correction splits
$\{25/75,50/50,75/25\}$, with $200$ replications per cell ($120$ at $n=6400$) and
$6{,}000$ test points, and report per group the restricted and deployed risk, the gap
$R^{\rm slice}-R^{\rm dep}$, mean set size, maximal-threshold default frequency,
and the dispersion of the corrected thresholds. At $n=1600$ the fitted union's
pooled set size is $4.6$ to $4.9$ across the three splits, against $3.7$ to $4.2$
for the unshaped union, so a larger fitting fraction does not reduce the fitted
sets. Under correlated labels ($\rho=0.5$), fitted and Kandinsky deployments
can have smaller sets than the union at higher pooled risk; these are
within-study comparisons, not estimates from the preceding three-method
experiment.

\begin{figure}[t]
\centering
\includegraphics[width=\textwidth]{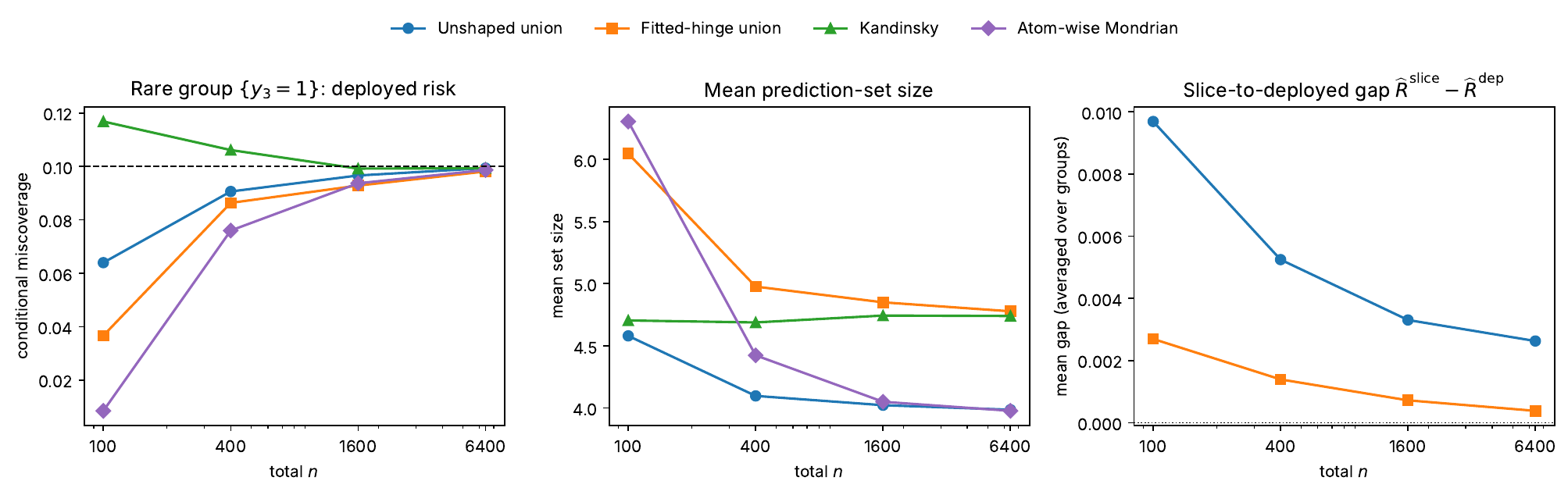}
\caption{Fitted-shape comparison on the overlapping-label model ($\alpha=0.10$,
independent labels, $50/50$ fit/correction split). Left: deployed miscoverage on
the rare focal group $\{y_3=1\}$, dashed line at $\alpha$; Kandinsky's uncorrected
quantile exceeds $\alpha$ at small $n$, while the CRC-corrected methods stay at or
below it. Middle: mean prediction-set size, smallest for the unshaped union.
Right: restricted-to-deployed gap averaged over groups, smaller for the fitted shape.
Averages over $200$ replications per configuration ($120$ at $n=6400$).}
\label{fig:sim-fitted-shape}
\end{figure}

\subsection{Structured-loss reconciliation experiment}
\label{app:sim-overlap-reconciliation}

On $\mathcal Y=\{0,1\}^3$ we use Hamming-distance-to-set loss
$\ell(C,y)=\min_{v\in C}d_{\mathrm H}(v,y)$ for nonempty $C$ and
$\ell(\varnothing,y)=3$, with $B=3$ and $\alpha=0.5$.
We take $Z_j=\sqrt{0.4}\,F+\sqrt{0.6}\,E_j$,
$Y_j=\mathbf1\{Z_j>0\}$, and $\eta_j=1.3Z_j+\xi_j$, with
independent standard normal factors and the candidate score from
Appendix~\ref{app:sim-overlap-coverage}. We compare feature-defined
groups $G_j=\{(x,y):M_j(x)=1\}$ with independent
$M_j\sim\mathrm{Bernoulli}(1/2)$ to label-defined groups
$G_j=\{(x,y):y_j=1\}$. In the latter case, the all-zero candidate
belongs to none of the focal groups and is excluded from the union.
The shape is $b\equiv0$ in both settings.

We estimate population risk crossings from $400{,}000$ draws.
We then run $100$ replications at $n=400,1600,6400$ and $25{,}600$,
with $15{,}000$ fresh test observations per replication. Coordinate symmetry gives
a common limiting restricted threshold within each scenario. For label-defined
groups, the estimated population restricted risk is $0.500$ and deployed risk
$0.362$; at $n=25{,}600$, the empirical counterparts are $0.500$ and
$0.360$. For feature-defined groups the limiting reconciliation gap is
zero. Throughout, the computed gap satisfies
\[
0\le R_j^{\mathrm{slice}}-R_j^{\mathrm{dep}}\le B\Delta S_j.
\]
The estimated residual $0.138$ is specific to this Hamming-loss model
and is not the sharp constant for the distinct product-form lower bound.

\begin{figure}[t]
\centering
\includegraphics[width=\textwidth]{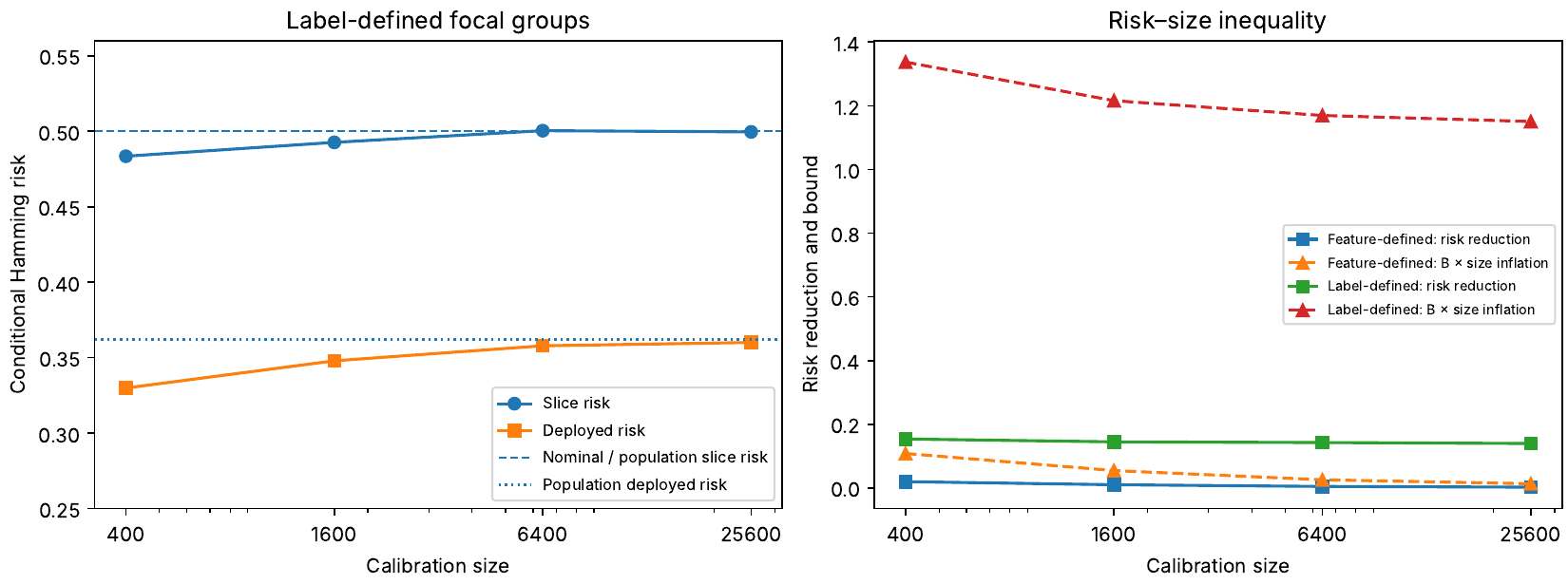}
\caption{Hamming-loss reconciliation: restricted versus deployed risk for
label-defined groups (left), and risk reduction versus the bound
$B\Delta S_j$ for label- and feature-defined groups (right). Population
limits are estimated by simulation; results average $100$ replications.}
\label{fig:sim-overlap-gap}
\end{figure}

\subsection{Reproducibility of the deployment-aware experiment}
\label{app:reproducibility-deployment}

This experiment uses the three-label ordinal model of
Proposition~\ref{prop:joint-direct-size-separation}: $Y\sim\mathrm{Unif}\{0,1,2\}$
independent of candidate scores $U_0,U_1,U_2\stackrel{\mathrm{iid}}\sim
\mathrm{Unif}[0,1]$, nested family $D_t^h=\{v:U_v-h_v\le t\}$, squared ordinal
loss $\ell(C,y)=\min_{v\in C}|v-y|^2$ with $\ell(\varnothing,y)=B=4$, overlapping
groups $G_1=\{0,2\}$ and $G_2=\{1,2\}$, and $\alpha=0.10$. The thresholds solve
the nested-family CRC equation by the ascending event sweep of
Appendix~\ref{app:computation}: for each group the per-point deployed loss is
initialized at $B$ and lowered as candidates enter, and the smallest cutoff whose
corrected group risk does not exceed $\alpha$ is returned, with the full-set
default triggered when $(n_j+1)\alpha<B$.

We compare the following procedures. Methods that learn no shape, namely the group-restricted
union ($h\equiv0$, per-group restricted CRC deployed as the pointwise union of
Section~\ref{subsec:overlap-validity}), the deployment-aware procedure of
Section~\ref{subsec:deployment-aware-crc} with $h\equiv0$ and with the oracle shape
$h^\star=(1-2\alpha/(1+2\alpha(B-1)),\,1-2\alpha,\,1)=(0.875,0.800,1.000)$, the
atom-wise Mondrian refinement over the three singleton cells, and the
group-indicator quantile deployed without correction, calibrate on the full budget
$n$. The learned deployment-aware variant fits its shape on an independent fit fold
and runs the CRC correction on the remaining correction fold. A group-restricted union
restricted to the correction fold is recorded as a split-cost diagnostic. The
oracle-labeled shape is the explicit population feasible construction of
Proposition~\ref{prop:joint-direct-size-separation}; that proposition does not
prove optimality among all directly valid sets. The learned shape minimizes
$\sum_v h_v$ over the grid $\{0,0.1,\dots,1\}^3$ ($1{,}331$ offsets) subject to both
fitting-fold group risks not exceeding $\alpha$ with a zero fitting margin, after
which the independent CRC correction restores exact finite-sample validity; for
large fit folds the search subsamples the fit fold to at most $2{,}000$ rows with a
deterministic seed, leaving the correction fold untouched. The learned shape is
therefore a practical variant and is not claimed to satisfy the finite-library
oracle inequality of Proposition~\ref{prop:joint-direct-oracle}. In this model the
intercept-augmented group-indicator quantile model (an intercept plus the two
group membership indicators) is saturated over the three observed membership
patterns $(1,0)$, $(0,1)$, $(1,1)$, so its pinball minimizer equals the per-label
empirical quantile, taken as the largest pinball minimizer (the
$(\lfloor(1-\alpha)n_v\rfloor+1)$-th smallest score, matching the specified implementation);
the resulting comparator coincides with the
Kandinsky construction for this design and targets coverage rather than structured
risk. Being intercept-augmented, it is distinct from the intercept-free objective
$Q$ of Corollary~\ref{prop:pinball-alignment}, which uses only the two group
indicators and is not saturated in this three-class model. We sweep $n\in\{100,400,1600,6400\}$ and
fit/correction splits $\{25/75,50/50,75/25\}$, with $600$ replications per cell and
$6{,}000$ fresh test points, and record for each group the deployed risk, mean set
size and maximal-threshold default frequency.

At $n=6400$ and the $50/50$ split the mean set sizes are $2.870$ (group-restricted union),
$2.760$ (direct, unshaped), $2.740$ (direct, learned), $2.687$ (direct, oracle),
$2.926$ (atom-wise) and $2.699$ (group-indicator quantile), with worst-group
deployed risks (maximum over groups of the group-mean risk) $0.055$, $0.099$,
$0.088$, $0.097$, $0.027$ and $0.130$; Monte Carlo standard errors on mean size are
at most $0.005$. The oracle variant converges toward $S_{\mathrm{direct}}=2.675$
(reaching $2.687$ at $n=6400$), and the realized reduction exceeds the population
separation $0.075$ because the empirical group-restricted union sits above its own optimum
$S^\star_{\mathrm{slice}}=2.750$. At $n=100$ with $50/50$ splitting, the full-budget
group-restricted union and the unshaped and oracle direct methods do not default (mean sizes
$2.93$, $2.89$, $2.89$); the learned variant, whose split correction fold has
per-group counts below $B/\alpha=40$, defaults on every replication, and the
correction-fold union defaults on about $88\%$. The planned companion archive is intended to provide the
simulation script, per-cell outputs, and random seeds.

\subsection{Direct-calibration Hamming comparison}
\label{app:reproducibility-direct-hamming}

This comparison extends the Hamming-distance-to-set experiment of
Appendix~\ref{app:sim-overlap-reconciliation} using the same generative model,
$B=3$, $\alpha=0.5$, and $\mathcal Y=\{0,1\}^3$. For label-defined groups
$G_j=\{y:y_j=1\}$ the all-zero vector is ineligible for every group, so the group-restricted
union and the direct variants share the common seven-candidate universe. For
feature-defined groups, a zero-gap benchmark, a test point belonging to no feature
group receives the empty set under the per-row union; the global-threshold direct
methods use the same ``no group implies empty'' deployment eligibility, so the
size comparison isolates the effect of the common threshold from candidate
eligibility. The learned-shape fit-fold size objective uses the same eligibility.
No candidate-universe expansion is used to favor direct correction.

A common total budget $n\in\{400,1600,6400,25600\}$ is used for the group-restricted union,
direct unshaped, direct fitted, and atom-wise comparators; the direct fitted shape
uses an independent fitting fold, followed by groupwise CRC on the correction fold
and global-maximum deployment. A group-indicator quantile union is recorded as a
coverage reference; with three focal indicators over seven nonzero label patterns
this model is not saturated, so these are three independent conditional quantiles
and are not the joint Kandinsky optimizer, and their numbers are not attributed to
Kandinsky. The study reports $100$ replications across the fitting/correction
splits of the preceding direct experiment. At $n=25600$ for label-defined groups,
the respective mean set sizes for the group-restricted union, direct unshaped, direct fitted and
atom-wise are $1.579$, $1.313$, $1.157$, and $2.387$, with worst-group risks
$0.362$, $0.497$, $0.493$, and $0.184$. For feature-defined groups, with
deployment eligibility matched across methods, the group-restricted-union, unshaped-direct and
learned-direct sizes are $1.321$, $1.325$, and $1.688$: the unshaped direct size
nearly matches the union (a paired difference of about $0.27\%$), so the common
threshold incurs little observed cost in this symmetric case, and only the learned
shape enlarges the set. Figure~\ref{fig:sim-hamming} displays the comparisons.

The reported sizes are pooled expected cardinalities, averaged over all test
points, compared at a common group-conditional risk target $\alpha=0.5$; they
are not themselves group-conditional, and no pooled-risk claim is made. For reference, at $n=25{,}600$ and the
$50/50$ split the pooled (marginal) risk is $0.626$ for the group-restricted union and $0.790$
for unshaped direct under label-defined groups, and $0.809$ and $0.807$
respectively under feature-defined groups; all exceed $\alpha$ without
contradicting the groupwise guarantee, which controls conditional risk only. Under
feature-defined membership the three independent $\mathrm{Bernoulli}(1/2)$ masks
leave about $1/8$ of test points in no group, where every method returns the empty
set, so unconditional average set size partly reflects behavior outside the
controlled groups. The exact fitted-shape library, random seeds, and
replication-level outcomes are intended for the companion archive.

\FloatBarrier

\section{Diabetic-retinopathy study: data and paired comparisons}
\label{app:ddr-protocols}

We first document the original fixed-pool DDR evaluation and then the
independent paired comparison with deployment-aware calibration. The two
studies use different redraw counts and must not be combined when computing
Monte Carlo standard errors.

\subsection{Reproducibility of the DDR experiment}
\label{app:reproducibility-ddr}

\textbf{Historical score table and classifier provenance.}
The reported analysis uses an existing $2{,}428$-image DDR grading score
table~\citep{li2019ddr}, with observed grade counts
$(1186,125,889,47,181)$ and softmax columns
$(\texttt{id},\texttt{grade},p_0,\dots,p_4)$. The scores are attributed
to an EfficientNet-B0 classifier; evaluation of this table yields
quadratic-weighted kappa $0.8507$ and balanced accuracy $0.6678$.
These numbers describe the supplied scored observations, but the
available artifacts do not establish that the classifier was trained
after deduplication, or that its fitting and scored partitions were
independent at the image or patient level. Accordingly, we treat the
$2{,}428$ observations as a \emph{historical fixed scored pool}, not as a
provenance-verified held-out test sample.

A separate reproducibility pipeline implements perceptual-hash screening,
exact-byte deduplication before splitting, grade-stratified model fitting,
and provenance checks. Its proposed protocol and training recipe must not
be interpreted as having generated the historical scores. A definitive
real-data analysis requires the actual retrained checkpoint, scored-pool
manifest, image/content audit, near-duplicate review or conservative
exclusions, and matching file hashes; the corresponding results must then
be recomputed rather than inferred from the historical table. The
conformal redraws below do not retrain the classifier.

\textbf{CRC evaluation and randomization.}
Each of $100$ redraws independently partitions the fixed scored pool
uniformly into $1{,}214$ calibration and $1{,}214$ test images, without
grade stratification. The randomized adaptive prediction-set score is
\[
  s(x,k)=\sum_{j:p_j(x)>p_k(x)}p_j(x)+U_xp_k(x),
  \qquad U_x\sim\mathrm{Unif}(0,1),
\]
with a new independent $U_x$ for every image and redraw, shared across
candidate grades and evaluated methods. The loss is the bounded ordinal
distance to the prediction set, with $B=4$ and $\alpha=0.10$. Group-specific
correction thresholds use~\eqref{eq:overlap-pw-threshold} and $\Lambda=[0,1]$;
if there is no feasible threshold, the procedure uses $\lambda_{\max}$.
This default is forced when $n_j+1<B/\alpha=40$. In the completed redraws,
the exact-grade calibration for grade~$3$ defaults in every replication;
none of the four original overlapping groups defaults. The evaluator reports
zero pointwise-nesting failures. The result tables are generated from the
raw redraw-level CSVs, with reconciliation differences computed before
rounding.

\textbf{Scope of the uncertainty statements.}
Reported standard errors describe variation over the $100$ random
calibration/test partitions and APS-score redraws \emph{conditional on
the historical fixed pool}. They neither quantify uncertainty for
independent future patients nor establish that the original model-fitting
and scoring partitions were independent. Random partitioning supports
a within-pool evaluation of the calibration rules, but does not repair
unverified upstream image or patient duplication. All DDR numerical
results below are therefore descriptive fixed-pool findings pending a
provenance-verified rerun; the distribution-free theorem requires its
stated exchangeability conditions.

\paragraph{Detailed DDR results.}
The following tables report means over the same $100$ historical-pool redraws. The exact-grade
refinement is the five-grade partition, which is the finest disjoint
refinement of the four groups considered in the main text. For structured
losses, restricted risk may exceed the deployed risk because candidates admitted
from other groups lower the distance-to-set loss. The reported
$\widehat\Delta_j=\widehat R_j^{\rm slice}-\widehat R_j^{\rm dep}$ is descriptive,
not necessarily the population residual at aligned cutoffs.

\begin{table}[t]
\centering\small
\caption{Historical DDR score pool: disjoint severity groups ($\alpha=0.10$). Values average
100 uniform split-and-score redraws. Risk and restricted-risk Monte Carlo SEs
are at most $0.007$, and mean set-size SEs are at most $0.016$, conditional
on the fixed pool.}
\label{tab:ddr-disjoint}
\resizebox{\textwidth}{!}{%
\begin{tabular}{lrrrrrrr}
\toprule
Severity group & $\bar n_{\rm cal}$ & Marg. risk & Mond. risk & Restricted risk & Marg. size & Mond. size & Fallback\\
\midrule
No DR $\{0\}$ & 594.4 & 0.040 & 0.057 & 0.098 & 1.47 & 2.19 & 0.00\\
Mild/mod. $\{1,2\}$ & 507.3 & 0.143 & 0.055 & 0.098 & 1.68 & 2.51 & 0.00\\
Severe/PDR $\{3,4\}$ & 112.3 & 0.208 & 0.036 & 0.079 & 1.42 & 2.05 & 0.00\\
\bottomrule
\end{tabular}}
\end{table}

\begin{table}[t]
\centering\small
\caption{Historical DDR score pool: overlapping groups. $\widehat\Delta$ is the restricted-to-deployed
risk difference, computed before rounding. Risk and restricted-risk Monte Carlo
SEs are at most $0.009$ and set-size SEs are at most $0.030$. The grade-$3$
row is an evaluation subgroup, not a separately guaranteed union group.}
\label{tab:ddr-overlap}
\resizebox{\textwidth}{!}{%
\begin{tabular}{lrrrrrrr}
\toprule
Clinical group & $\bar n_{\rm cal}$ & Marg. risk & Grade risk & Union risk & Restricted risk & $\widehat\Delta$ & Union size\\
\midrule
No DR $\{0\}$ & 594.4 & 0.040 & 0.031 & 0.058 & 0.098 & 0.041 & 2.18\\
NPDR $\{1,2,3\}$ & 530.8 & 0.148 & 0.023 & 0.056 & 0.100 & 0.044 & 2.48\\
Referable $\{2,3,4\}$ & 557.2 & 0.135 & 0.023 & 0.038 & 0.098 & 0.060 & 2.34\\
Severe/PDR $\{3,4\}$ & 112.3 & 0.208 & 0.017 & 0.036 & 0.079 & 0.044 & 2.05\\
Grade $\{3\}$ & 23.5 & 0.248 & 0.000 & 0.041 & -- & -- & 2.36\\
\bottomrule
\end{tabular}}
\end{table}

\FloatBarrier

\subsection{Paired deployment-aware DDR evaluation}
\label{app:reproducibility-ddr-direct}

The direct-method evaluation reuses the same historical fixed $2{,}428$-image
DDR score pool, randomized APS score, five-grade universe, ordinal loss
($B=4$), target $\alpha=0.10$, and four overlapping severity groups as
Appendix~\ref{app:reproducibility-ddr}. It is a separate $200$-redraw
paired comparison and must not be conflated with the $100$ redraws used
for the original DDR tables above. Each redraw allocates $1,214$ images
to conformal calibration and $1,214$ to evaluation. For the fitted direct
variant, fitting and correction are disjoint subsets of the calibration
allocation, examined at $25/75$, $50/50$, and $75/25$ splits; the sub-split
is randomized without sorting the underlying grade-ordered scored pool.
The original implementation's sorting-induced grade confounding was
corrected before the reported results were rerun.

The full-budget union has mean size $2.322$ and group risks
$(0.056,0.054,0.037,0.036)$. Unshaped direct CRC has mean size
$2.518$ and group risks $(0.002,0.025,0.027,0.074)$.
The fitted direct sizes are $3.57$, $3.64$, and $4.99$ across the
three fitting/correction allocations. In the $75/25$ allocation, the severe-group correction
count averages about $28$, below the corrected feasibility scale
$B/\alpha=40$, leading to a groupwise maximal-threshold default in
about $98\%$ of redraws. The direct method then deploys the full
universe, whereas the union confines its maximal threshold to the
eligible grades; its corresponding mean size is about $3.77$.
Figure~\ref{fig:sim-realdata-direct} displays the summary. All standard errors are conditional on this historical scored pool and do
not support inference to independent patients. The direct-method evaluator
accepts a score table but does not establish its training provenance.
A definitive repeat must regenerate the scores with the duplicate-aware
pipeline and pass the separate integrity checks for the scoring recipe,
content audit, split manifest, near-duplicate review, and file hashes.
The companion direct-method scripts, fitting library, seeds, and saved
historical replication-level outcomes are intended to document this fixed-pool comparison;
they are not a substitute for that definitive rerun.

\FloatBarrier

\section{Additional tightness and local-support simulations}
\label{app:diagnostic-simulations}

This section supplies the rank-granularity and support-barrier experiments
used to check the finite-sample conclusions of Sections~\ref{sec:mondrian-crc}
and~\ref{sec:lower-bounds}.

\subsection{Sandwich envelope and rank-grid witnesses}
\label{app:reproducibility-sandwich}

We use ordinal absolute-error loss on five grades,
$\ell(C,y)=\min_{c\in C}|c-y|$ with $\ell(\varnothing,y)=B=4$,
and target $\alpha=0.4$. The minimum threshold returns the empty set
and the maximum returns the full set, so both endpoint conditions hold.
In the benign experiment, $n_k\in\{100,200,500,1000,2000\}$ and
$M\in\{1,2,4\}$. The $(n_k+1)K$ label scores are sampled without
replacement from distinct threshold values replicated at most $M$
times; group labels are uniform. This ensures at most $M$ augmented
loss paths can jump at a common threshold. Each configuration uses
$50{,}000$ replications and checks the realized jump multiplicity.
The observed risks approach $\alpha$ as support increases; thirteen of the
fifteen point estimates lie inside the sandwich interval, and the remaining two
(both at $n_k=2000$) exceed $\alpha$ by at most $0.2$ Monte Carlo standard
errors, so all fifteen are compatible with the interval within two standard
errors.

For the rank-grid witness, $N=n_k+1$ loss paths are randomly assigned
to $\lfloor N/M\rfloor$ blocks of $M$ simultaneous jumps of height $B$,
with any remainder paths zero. At $n_k\in\{10,12,\ldots,120\}$ (the $56$ even
support sizes) and $M\in\{1,2,4\}$, using $20{,}000$ replications per
configuration, the exact witness risk is
\[
 \frac{BM}{N}\left\lfloor\frac{N\alpha}{BM}\right\rfloor
 =\alpha-\Delta_{n_k,M}(\alpha,B).
\]
The simulated risks follow this sawtooth across all $168$ configurations: the
$20$ zero-variance cells match the prediction exactly, and each of the remaining
$148$ cells agrees with it within three Monte Carlo standard errors ($139$ within
two). Zero-variance cells are compared by equality rather than standardized
deviation.
These experiments distinguish the sandwich's uniform tightness bound
from the arithmetic rank-grid lower bound.

\begin{figure}[t]
\centering
\includegraphics[width=0.85\textwidth]{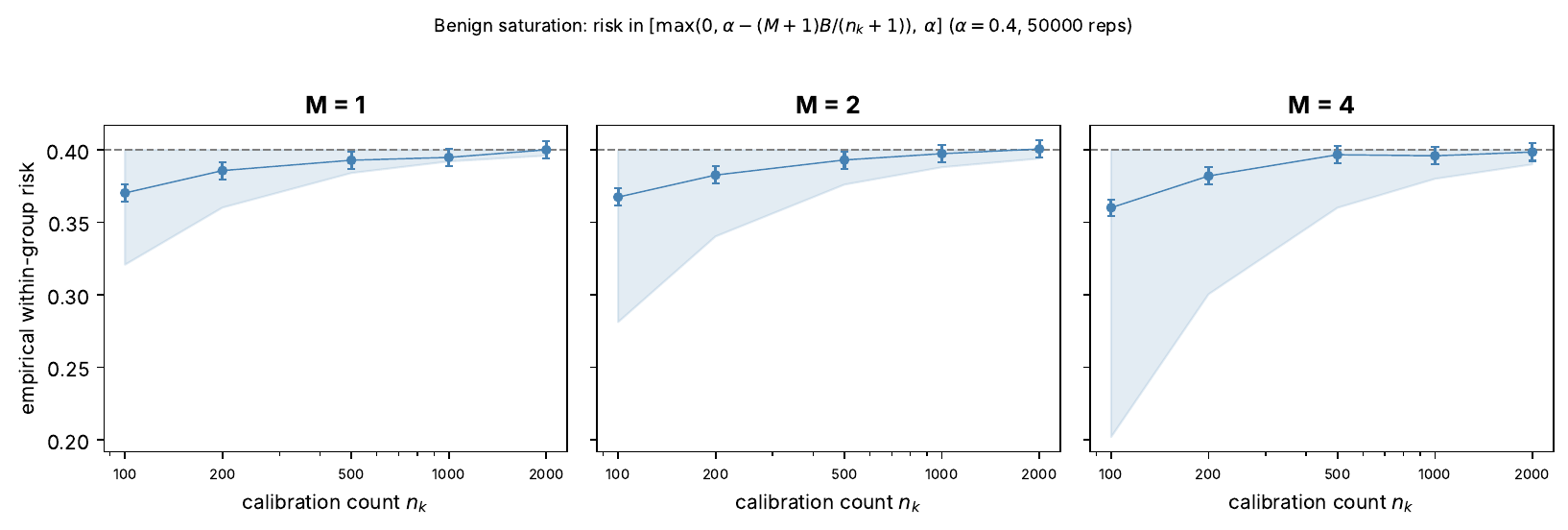}
\caption{Ordinal loss ($K=5$, $B=4$, $\alpha=0.4$): observed restricted risk
under a benign jump budget $M$ and the theoretical sandwich interval.
$50{,}000$ replications per configuration.}
\label{fig:sim-sandwich}
\end{figure}

\begin{figure}[t]
\centering
\includegraphics[width=0.85\textwidth]{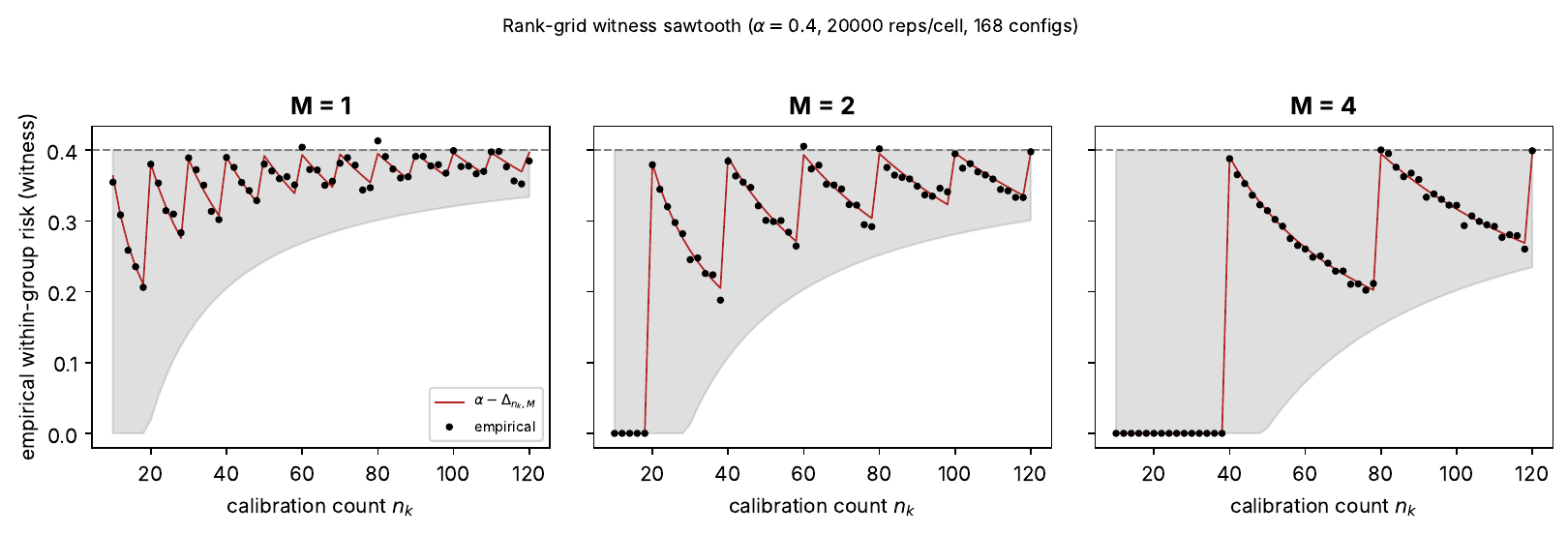}
\caption{Rank-grid witness: observed restricted risk and
$\alpha-\Delta_{n_k,M}(\alpha,B)$, alongside the sandwich interval.
$20{,}000$ replications per configuration.}
\label{fig:sim-sandwich-witness}
\end{figure}

\subsection{Small-group barrier}
\label{app:reproducibility-barrier}
\label{subsec:sim-small-group}

We use five labels, iid continuous $\mathrm{Unif}(0,1)$ scores, and
$\alpha\in\{0.05,0.10,0.20\}$. The designs are: ordinal
absolute error ($B=4$) with $g\equiv1$; multilabel false-negative
rate ($B=1$) for true label sets of size 1--3 with $g\equiv1$;
and ordinal error with $g(x,y)=y$.
The first two are general bounded monotone loss-curve experiments;
the last has distinct common-height one-jump restricted losses.

Below $n+1<B/\alpha$, the corrected criterion is infeasible at every
threshold. At or above that boundary, a nonmaximal feasible threshold
exists because the maximum observed candidate score lies strictly
below $\lambda_{\max}$. Consequently the first two designs have the
deterministic default indicator
$\mathbf1\{n+1<B/\alpha\}$, assuming the full-set endpoint loss is
zero. The classwise design checks conditions (P1)--(P3) of
Appendix~\ref{app:exact-two-valued}.

The classwise randomized comparator interpolates between adjacent
rank thresholds: with $\kappa=\lfloor N\alpha/B\rfloor$ and
$\theta=N\alpha/B-\kappa$, it selects $S_{(\kappa+1)}$ with
probability $\theta$ and $S_{(\kappa)}$ otherwise, using
$S_{(0)}=\lambda_{\max}$ and $S_{(n+1)}=\lambda_{\min}$.
In the sub-barrier regime it defaults
with probability $1-N\alpha/B$. The simulation uses 500
replications for each deterministic panel and 20,000 for the
randomized classwise comparison. The separate exchangeable
spike-mixture witness of Proposition~\ref{thm:small-group-nontriviality}
is covered analytically, not simulated.

\begin{figure}[t]
\centering
\includegraphics[width=0.86\textwidth]{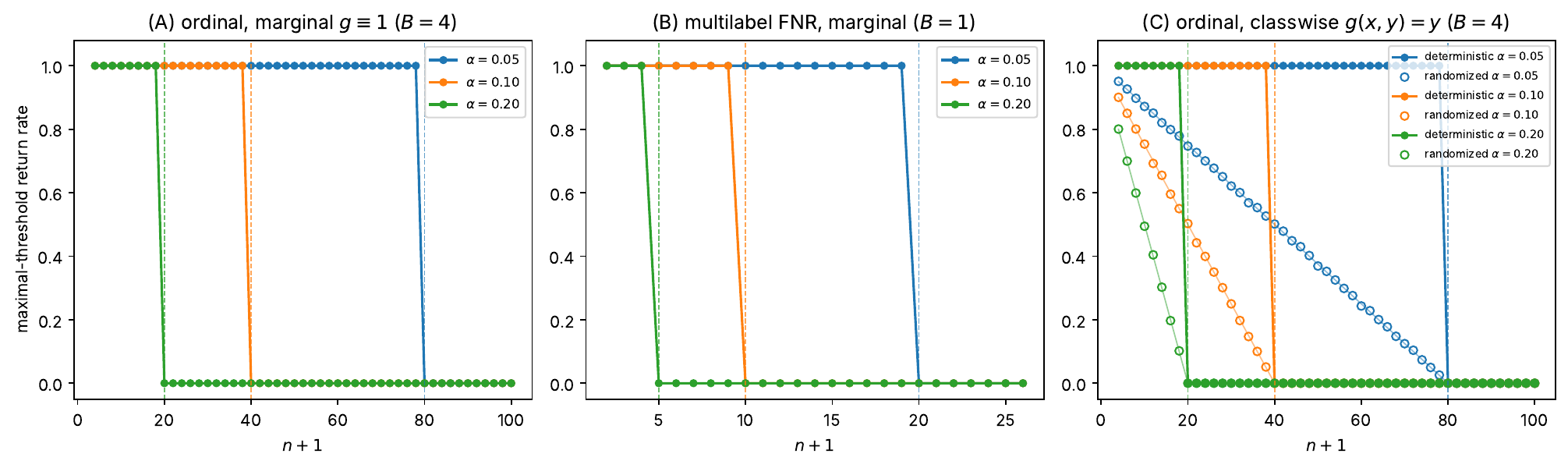}
\caption{Empirical maximal-threshold return (default) rates against $n+1$ for
three designs: ordinal absolute-error loss with a marginal group
($g\equiv1$, $B=4$, left); multilabel false-negative rate with a marginal
group ($B=1$, middle); and ordinal absolute-error loss with classwise groups
($g(x,y)=y$, $B=4$, right). Dashed verticals mark the feasibility boundaries
$n+1=B/\alpha$. Filled markers are the deterministic Mondrian rule, whose
default rate equals $\mathbf1\{n+1<B/\alpha\}$ in the first two designs. In the
right panel, open circles are the randomized comparator, whose sub-barrier
default probability matches the predicted $1-N\alpha/B$ (thin lines). The figure
illustrates these specific CRC algorithms, not the procedure-independent witness
of Proposition~\ref{thm:small-group-nontriviality}.}
\label{fig:sim-small-group}
\end{figure}

\bibliographystyle{plainnat}
\bibliography{References}
\end{document}